\documentclass[10pt,twocolumn,letterpaper]{article}

\usepackage[pagenumbers]{wacv} 

\usepackage{colortbl}   
\usepackage{pifont}     

\newcommand{\cmark}{\ding{51}}
\newcommand{\xmark}{\ding{55}}

\definecolor{wacvblue}{rgb}{0.21,0.49,0.74}

\usepackage[breaklinks,colorlinks,allcolors=wacvblue]{hyperref}

\usepackage[accsupp]{axessibility}  

\def\wacvPaperID{2818} 
\def\confName{WACV}
\def\confYear{2027}

\title{StablePrivacy: Zero-Shot Full-Body Video Anonymization Using Diffusion Models For Downstream Data Utility}

\author{
Rohit Bharadwaj$^{1}$\quad
Duolikun Danier$^{1}$\quad
Zhenzhang Ye$^{1}$\\
Jugesh Sundram$^{2}$\quad
Laura Sevilla-Lara$^{1}$\quad
Hakan Bilen$^{1}$\\[0.5em]
$^{1}$University of Edinburgh\qquad
$^{2}$Toyota Motor Europe\\[0.3em]
{\tt\small \{rohit.bharadwaj, duolikun.danier, zye2, l.sevilla, h.bilen\}@ed.ac.uk}\\
{\tt\small jugesh.sundram@toyota-europe.com}
}

\begin{document}
\maketitle
\begin{abstract}
Existing privacy research mainly targets facial anonymization in images, overlooking identifying cues beyond the face (\eg, clothing and hair) or the challenges of video data. We propose \textbf{StablePrivacy}, a diffusion-based framework for full-body video anonymization that improves the privacy--utility trade-off. Built on Stable Diffusion, our method operates in a zero-shot manner without dataset-specific training. The framework preserves pose and scene context via ControlNet conditioning and latent-space inpainting, while a memory bank enforces temporal consistency across frames. We further introduce an injection module that retains task-relevant structure while removing identity cues. Experiments on video instance segmentation, action recognition, and person re-identification show that StablePrivacy attains a stronger privacy--utility trade-off than existing full-body anonymization methods, with competitive downstream performance across tasks and the clearest gains on person re-identification. Code is available at \href{https://github.com/VICO-UoE/StablePrivacy}{https://github.com/VICO-UoE/StablePrivacy}
\end{abstract}

\section{Introduction}

\begin{figure}[t]
  \centering
  \includegraphics[width=\linewidth]{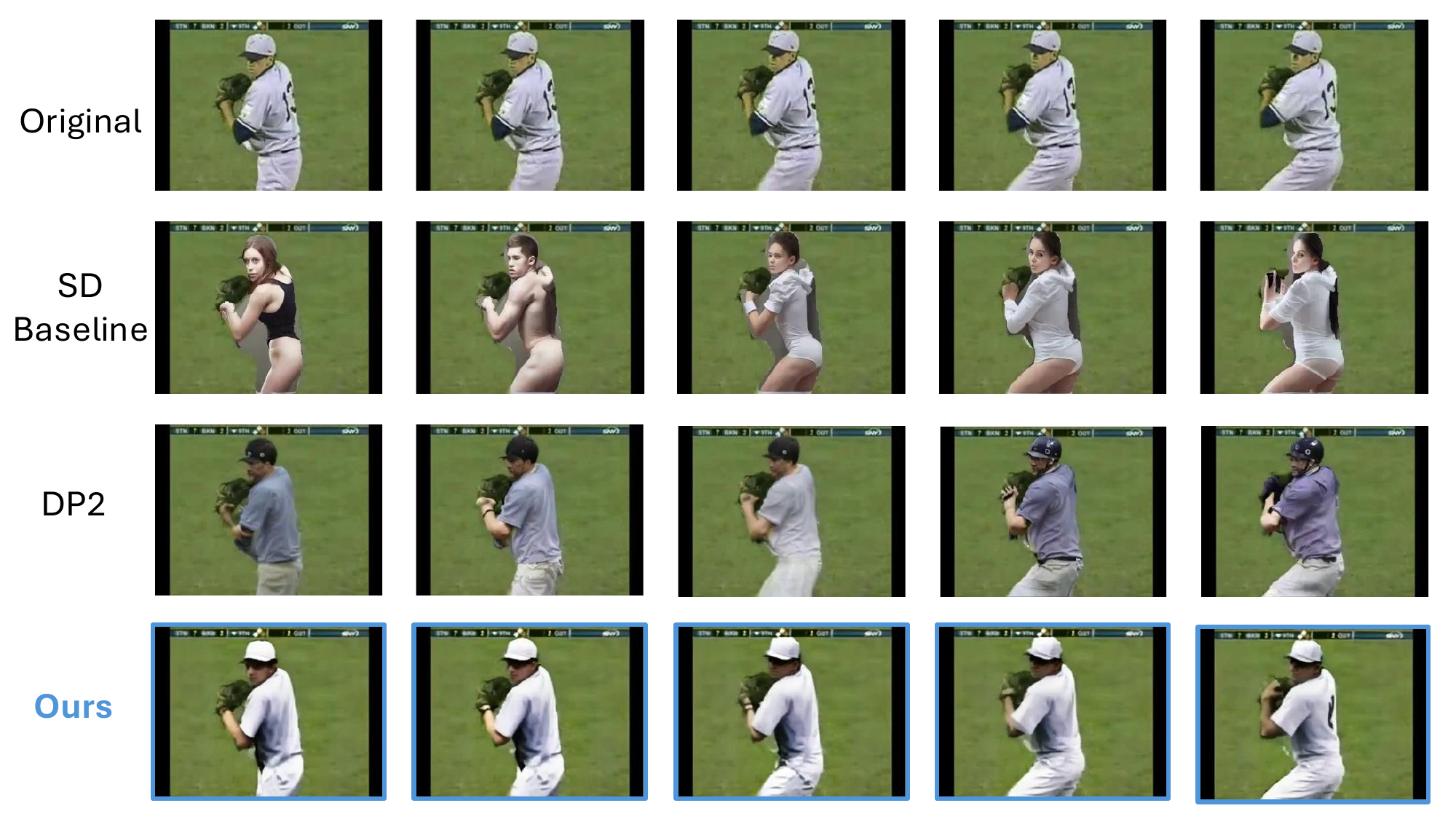}
  \caption{
  Qualitative comparison on UCF101: SD (Stable Diffusion) Baseline leads to unrealistic generations; DP2~\cite{hukkelaas2023deepprivacy2} (DeepPrivacy2) suffers from temporal inconsistencies; our method maintains a consistent, realistic appearance throughout the sequence.}
  \label{fig:teaser}
\end{figure}

Large-scale visual datasets containing people are a key driver of progress in modern computer vision. However, such data frequently contains personally identifiable information (PII), raising serious privacy concerns and regulatory challenges. Legal frameworks such as the General Data Protection Regulation (GDPR) impose strict requirements on the collection, storage, and use of personal data. As a result,  there is increasing interest in methods that enable the use of visual data while preventing the recovery of sensitive identity information.

A common strategy for protecting privacy in visual data is \emph{anonymization}, where identifying attributes of individuals are removed or replaced while preserving the information required for
downstream tasks. Traditional techniques such as blurring, pixelation, or masking are widely used in practice, but they often degrade visual content and can be reversed or bypassed by modern machine learning systems \cite{barattin2023attribute}. Consequently, recent work has explored {generative
anonymization} approaches that synthesize new identities while preserving scene context and task-relevant attributes.

Early generative anonymization methods were primarily based on generative adversarial networks  (GANs). Representative approaches such as CIAGAN \cite{maximov2020ciagan} and DeepPrivacy2
\cite{hukkelaas2023deepprivacy2} generate synthetic human appearances conditioned on structural cues such as pose or segmentation masks. While these methods produce plausible anonymized images, GAN-based approaches often struggle with training instability, limited diversity, and difficulties
maintaining high-fidelity details. Moreover, most existing work focuses on \emph{static image anonymization}, whereas many real-world applications involve video data.

Recently, diffusion models have emerged as a powerful alternative for image synthesis due to their improved training stability and high visual quality \cite{dhariwal2021diffusion,rombach2022high}.
Several recent works adapt diffusion models for anonymization, typically by conditioning Stable Diffusion on pose or segmentation cues and replacing the original identity through inpainting or
image-to-image generation \cite{malm2023rad,zwick2024context,ciftci2024my}. Although these approaches achieve promising results on images, they exhibit several important limitations.

First, many anonymization methods~\cite{kung2025face, el2025vera, he2024diff} focus exclusively on \emph{facial identity}, despite the fact that people can often be recognized from other cues such as clothing, body shape, hairstyle, and
accessories. Second, existing diffusion-based anonymization methods~\cite{ciftci2024my, zwick2024context, malm2023rad} typically operate on individual frames, generating each frame independently. This leads to temporal artifacts such as flickering, inconsistent identity appearance across frames, and unrealistic blending between foreground and background when applied to videos (see \cref{fig:teaser}).
Third, several privacy-preserving vision methods are designed for specific downstream tasks such as action recognition and rely on task-specific adversarial objectives \cite{dave2022spact, li2023stprivacy}.
While effective for their intended tasks, such methods often produce unrealistic outputs and do not generalize well across tasks. These limitations highlight the need for anonymization methods that simultaneously (i) remove identity information beyond the face, (ii) maintain temporal consistency in videos, and (iii) preserve utility for multiple downstream tasks without requiring dataset-specific training.

In this work, we introduce \textbf{StablePrivacy} (high-level overview in Fig.~9 of supplementary), a diffusion-based framework for \emph{zero-shot full-body video anonymization}. Our approach builds upon Stable Diffusion
\cite{rombach2022high} and integrates several mechanisms to balance privacy preservation and downstream utility. Specifically, we use pose-conditioned generation via ControlNet~\cite{zhang2023adding} together with latent-space inpainting to preserve the original pose and background context.
To address temporal inconsistencies, we incorporate a memory bank within the
diffusion UNet that captures inter-frame dependencies during generation.
In addition, we propose an \emph{Injection Module} that selectively transfers structural information from the input video to the generation process, enabling the model to preserve task-relevant cues while removing identity-related features.

Unlike many existing anonymization approaches, StablePrivacy operates in a \emph{zero-shot} manner without dataset-specific fine-tuning, enabling generalization across datasets and tasks. We evaluate our method on multiple downstream benchmarks, including video instance segmentation, action recognition, and person re-identification. Our experiments demonstrate that StablePrivacy achieves a stronger privacy--utility trade-off compared to existing full-body anonymization methods. Naive combinations of off-the-shelf components do not work in this setting, as our ablations show. Our main contributions are summarized as follows:
\begin{itemize}
\item We introduce StablePrivacy, a diffusion-based framework for zero-shot full-body video
anonymization that preserves pose, background context, and temporal consistency.
\item We propose an Injection Module and memory-bank-based temporal conditioning that enable
the model to preserve downstream task utility while removing identity information.
\item We conduct extensive experiments across multiple downstream tasks and datasets, showing that
our approach achieves a better privacy--utility trade-off compared to existing anonymization methods.
\end{itemize}

\section{Related Work}
\label{sec:related-work}

\begin{figure*}[t]
    \centering
    \includegraphics[width=0.64\linewidth]{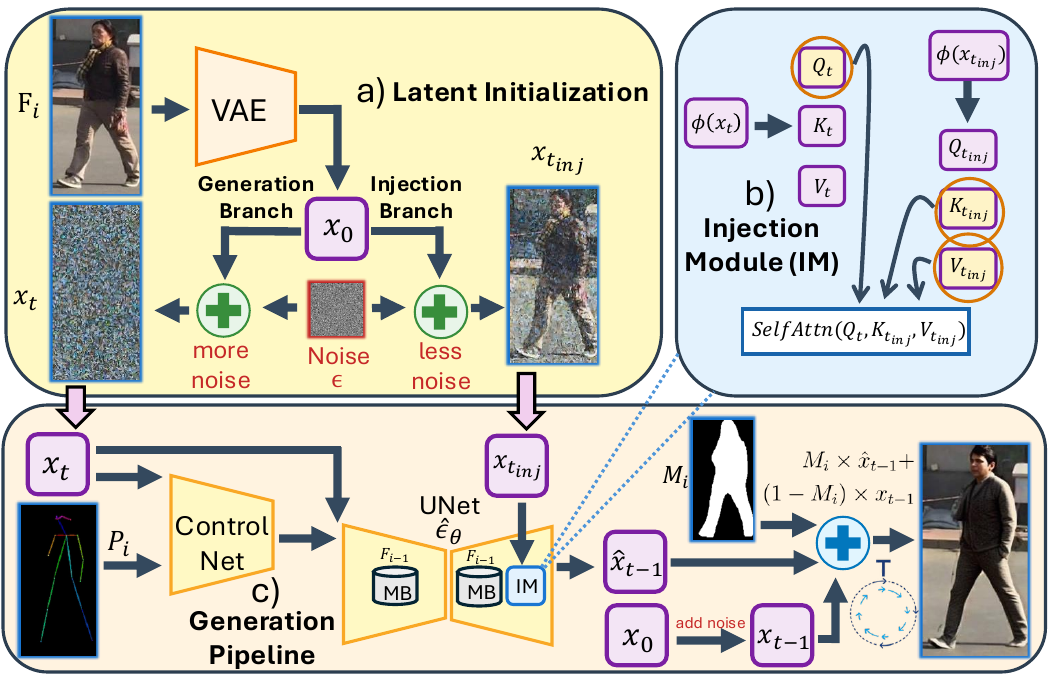}
    \caption{\textbf{StablePrivacy}: Our zero-shot, diffusion-based full-body video anonymization framework has three main components. a) In the latent initialization phase, the input video frames are encoded into the diffusion model's latent space through two parallel branches. b) The Injection Module (IM) preserves features that are important for downstream tasks while keeping the visual domain aligned with the input. c) The main generation pipeline then utilizes ControlNet, memory bank (MB), IM, and latent inpainting to anonymize the input video while preserving the scene context and maintaining temporal consistency.}
    \label{fig:main-method}
\end{figure*}

\noindent \textbf{Generative Anonymization:}
Classical techniques such as blurring or pixelation hide identity information, but they degrade data utility and can be reversed by modern learning-based methods~\cite{maximov2024data, barattin2023attribute}. Generative approaches instead synthesize new identities while preserving non-sensitive attributes such as pose and scene context. Early methods including CIAGAN~\cite{maximov2020ciagan}, FALCO~\cite{barattin2023attribute}, and DeepPrivacy2 (DP2)~\cite{hukkelaas2023deepprivacy2} rely on GANs~\cite{goodfellow2020generative}, while diffusion models have more recently been adopted for their improved training stability, output quality, and diversity~\cite{dhariwal2021diffusion, rombach2022high}.

\noindent \textbf{Full-body Anonymization:}
Many approaches target facial identity alone~\cite{kung2025face, el2025vera, he2024diff}, yet individuals remain recognizable from clothing, body shape, accessories, hair, or skin tone~\cite{hukkelaas2023deepprivacy2, maximov2024data, ban2024faker}. DP2~\cite{hukkelaas2023deepprivacy2} remains a strong GAN-based full-body baseline using dense pose conditioning, and diffusion-based methods such as MBMC~\cite{ciftci2024my}, FADM~\cite{zwick2024context}, and RAD~\cite{malm2023rad} adapt Stable Diffusion with ControlNet conditioning to anonymize humans via image inpainting. All of these operate on individual frames, so when applied to videos they suffer from temporal flickering, inconsistent identity appearance across frames, and unrealistic foreground--background blending (see~\cref{fig:teaser,fig:acd_metric,fig:qualitative_comparison}).

\noindent \textbf{Privacy-Preserving Action Recognition:}
Another line of work suppresses privacy-sensitive attributes while preserving information for a specific downstream task, typically action recognition~\cite{wu2018towards, dave2022spact, li2023stprivacy, hinojosa2022privhar, kumawat2022privacy, ilic2024selective}, by learning an anonymization transformation with adversarial objectives. The resulting representations are task-specific and often visually unrealistic, which limits their applicability to other downstream tasks or to general-purpose anonymization.

\noindent \textbf{Human Animation and Video Translation:}
Pose-conditioned animation methods~\cite{hu2024animate, hu2025animate, tu2025stableanimator} render a subject from a reference image following a target pose sequence, with recent models such as Stable Animator~\cite{tu2025stableanimator} achieving temporally coherent pose-to-video generation. Video-to-video translation methods~\cite{liang2024looking, geyer2023tokenflow, cong2023flatten, liang2024flowvid} instead edit an input video while preserving its temporal structure. Both families are designed to \emph{preserve} identity rather than remove it, making them unsuitable for anonymization, but they provide useful baselines for how well pose and video conditioned generators can replace identity while preserving pose, motion, and scene context.

\section{StablePrivacy}
\label{sec:method}

\subsection{Overview}
\label{sec:method-overview}
Our goal is to anonymize the given input video in a manner that discards most of the private or sensitive information in the data, while still preserving the utility of the data in downstream task training by generating a temporally consistent realistic anonymized video.
Thus, the anonymized data generated by our method can be used to train a deep neural network model with minimal degradation to the downstream task performance (See Supplementary Fig.~10). Formally, we consider an input triplet $(F, M, P)$, where $F$ is the original video with $T$ frames that contains at least one person, $M$ is the corresponding human segmentation mask sequence, and $P$ is the human pose sequence. Each $F$, $M$ and $P$ has the same spatial ($H\times W$) and temporal ($T$) dimensions.
The goal is to develop an anonymization function: $\mathcal{A} : (F, M, P) \rightarrow F',$ that produces an anonymized video $F' \in \mathbb{R}^{T \times C \times H \times W}$, such that downstream models trained on $\mathcal{A}(F)$ maintain the task performance similar to the ones trained on the original video $F$, while ensuring that the identity in $F'$ is no longer recoverable or visually attributable to the original subject.
Thus, to achieve the above goal, we present our \emph{StablePrivacy} video anonymization framework as shown in~\cref{fig:main-method}. Each input video frame is first encoded into a latent space using SD's VAE. We then create two noisy latent versions: a heavily noised one for the generation branch, which provides strong privacy, and a lightly noised one for the injection branch, which preserves key structural and domain-specific cues for downstream tasks. The generation branch latent is processed by the pipeline in~\cref{fig:main-method}(c), which uses ControlNet conditioning and latent inpainting to keep the input pose and background. A memory bank (MB) module enforces temporal consistency, while the injection module (IM) ensures the generated structure and visual domain match the input, which is crucial for downstream performance. We next outline the preliminaries and then describe each component in detail.

\subsection{Preliminaries}
The SD model is based on a UNet architecture, which takes as input the noisy latent, $x_t$, and a diffusion timestep, $t$, and predicts the noise that was added to the clean latent, $x_0$. Formally, UNet can be represented as $\hat{\epsilon}_{\theta}(x_t, t)$, where $\theta$ denotes the parameters of the neural network. The UNet in SD employs two types of attention mechanism: self-attention, and cross-attention. For a given self-attention layer ($l$), the intermediate noisy latents $\phi^{(l)}(x_{t})$ are transformed into three matrices, i.e., $Q^{(l)}$, $K^{(l)}$, and $V^{(l)}$, through learned linear projections, representing the query, key, and value features, respectively. The query and key features are used to compute the self-attention map ($A$) by computing $A = \text{softmax}\left( ({Q \cdot K^{T})/ \sqrt{d}} \right)$
where $d$ denotes the dimensionality of $Q$.
For clarity, the layer index $l$ is omitted. Each element $A_{i,j}$ represents the correlation between the $i^{\text{th}}$ query patch and the $j^{\text{th}}$ key patch, and reflects how relevant each key is to the corresponding query. The attention map is then applied to the value features to produce the output, $A \cdot V$.
This weighted aggregation of value features is followed by a linear projection and residual connection to yield the final hidden states of the self-attention block. Thus, we denote the above self-attention computation as $\text{SelfAttn}(Q, K, V)$.

\subsection{Latent Initialization}
To achieve the objective stated in~\cref{sec:method-overview}, we first start with a given input video frame, $F_i$, which is first encoded by the SD's VAE encoder, to obtain the clean latent tensor, $x_0$. Once we have $x_0$, we pass it through two parallel branches, namely, the injection branch and the generation branch, where the same noise given by $\epsilon$, is added to the clean latent, $x_0$.
Since we want to remove most of the private information in the input data, we initialize our generation branch by adding a higher amount of noise to the clean latent $x_0$, compared to the injection branch, which is then used in our proposed injection module (IM) to preserve crucial context features for downstream utility. Noise addition is governed by the forward diffusion process, i.e., $q(x_t|x_0) \sim \mathcal{N}(x_t; \sqrt{\bar{\alpha_t}}x_0, \bar{\beta_t}I)$. Thus, the noisy latent, $x_t$, at the end of the generation branch is given by:
\begin{equation}
\label{eq:noise-addn}
    x_t = \sqrt{\bar{\alpha_t}}x_0 + \sqrt{\bar{\beta_t}} \epsilon
\end{equation}
where $\alpha_t, \beta_t = 1 - \alpha_t$, are the parameters of the noise schedule in a given denoising timestep $t$, of the diffusion process. In addition, $\bar{\alpha_t} = \prod_{s=1}^{t} \alpha_s$, and $\bar{\beta_t} = 1 - \bar{\alpha_t}$, $\epsilon \sim \mathcal{N}(0, I)$.

Starting the generation from a highly noisy latent state allows the diffusion model to explore a broader region of the data manifold, thereby enabling more diverse and creative outputs that can deviate significantly from the input. In contrast, initialization from a low noise level keeps the latent close to the original data point, causing the model to perform primarily local refinements or reconstructions that preserve the overall content and structure of the input. Since we are motivated to anonymize the data and ensure that the privacy of individuals is maintained, we start the generation branch with a high noise level.
But at the same time preserve the overall structure and the visual domain of the generation by making use of low noise in the injection branch.
The noise addition in the injection branch follows a similar equation as in \cref{eq:noise-addn}, except that we significantly reduce the amount of noise added to the clean latent, $x_0$, as illustrated in \cref{fig:main-method}. We denote the noise strength of the generation and injection branches by $\sigma_g$ and $\sigma_i$.

The obtained injection latent, $x_{t_{inj}}$, thus captures the overall structure and fine-grained texture details of the input. However, simply reducing the noise level and starting the generation from the lower noise level will not work as it will fail to preserve privacy (refer~\cref{fig:privacy_utility}). Thus, to solve for the above privacy-utility dilemma, we make use of our proposed method of injecting features from the injection branch, i.e. $x_{t_{inj}}$, directly into the SD's UNet layers, so as to retain the input structure, but at the same time anonymizing the inputs due to our generation starting from high noise latents.

\subsection{Injection Module (IM)}
Previous works like Cross Image Attention~\cite{alaluf2024cross}, and SD-VICL~\cite{oorloff2025stable}, have shown the importance of query, key, and value features, in the self-attention layers of SD, to capture the semantic meaning of each location, offering context to each query, and representing the content that needs to be generated, respectively. Based on this insight, we inject the keys and values, i.e. $K_{t_{inj}}$, and $V_{t_{inj}}$, of the injection latent, to attend to the query $Q_{t}$ of the generation.
Formally, we can then rewrite the output as:
\begin{equation}
    \text{Output} = \text{softmax}\left( \frac{Q_{t} \cdot K_{t_{inj}}^{T}}{\sqrt{d}} \right) \cdot V_{t_{inj}}
\end{equation}
This injection of features allows the query to pay attention and focus on the semantically corresponding high quality input structure and textures that are captured by $x_{t_{inj}}$. For example, if a query token is located in the head region, it can then attend to the corresponding head region in the injection latent to refine and improve the fidelity of the head in the generation. In addition, we also inject the value features $V_{t_{inj}}$, which allows the generation to maintain the input structure and respect the input visual domain. The injection module (IM) in our approach is not the same as starting the generation from a lower noise level, as we only perform this injection process in the UNet decoder, specifically in the block with an output resolution of $16 \times 16$ for a $512 \times 512$ input tensor. Thus, our proposed IM helps generate higher-quality features that are important for both downstream data utility and visual domain alignment while maintaining respectable privacy. Moreover, unlike existing methods~\cite{alaluf2024cross, oorloff2025stable}, our keys and values are derived from the same image. As a result, we naturally avoid domain gap issues and achieve domain alignment without any additional adaptation steps. Since the injected keys and values are derived from a lightly noised latent of the input, the IM necessarily carries over some coarse appearance information, such as clothing type and colour, which are precisely the cues that Re-ID models exploit. This is a measured trade-off: \cref{fig:privacy_utility} shows that the IM does not simply move along the privacy--utility curve obtained by lowering the generation noise, and at a comparable privacy level our full method attains a $+9.5$ mAP utility gain over the variant without the IM, so this gain cannot be reproduced by noise tuning alone.

\subsection{Generation Pipeline}

After latent initialization (\cref{fig:main-method}(a)), the noisy latent $x_t$ is passed through the diffusion generation pipeline shown in \cref{fig:main-method}(c). The goal of this stage is to generate anonymized frames that preserve the original pose, maintain temporal consistency across frames, and seamlessly integrate the anonymized foreground with the original background.

\paragraph{Pose Conditioning.}
To ensure that the generated person follows the motion of the input video, we incorporate ControlNet~\cite{zhang2023adding} into the SD framework~\cite{rombach2022high}. ControlNet introduces spatial conditioning signals that guide the diffusion process beyond what can be achieved with text prompts alone. While ControlNet supports multiple conditioning modalities, we use only human pose conditioning in our method. This preserves the pose of the original subject and enables a fair comparison with the state-of-the-art anonymization method DP2~\cite{hukkelaas2023deepprivacy2}, which also relies on pose-based generation. Specifically, the pose image $P_i$ is provided as input to the ControlNet module, which produces intermediate feature maps that are injected into the diffusion UNet $\hat{\epsilon}_{\theta}$ together with the noisy latent $x_t$ and text prompt conditioning.

%
\paragraph{Temporal Consistency.}
Since our method operates on video inputs, maintaining temporal consistency across frames is essential. To address this, we incorporate a memory bank (MB)~\cite{liang2024looking} into the UNet layers. The memory bank stores the keys, values, and attention outputs computed for the previously generated frame $F_{i-1}$. When generating the current frame $F_i$, these stored features are concatenated with the current frame's attention features along the sequence-length dimension (refer Sec.~9 of the supplementary for further details). This allows the model to reference information from previous frames, encouraging consistent appearance and reducing temporal flickering (see~\cref{fig:acd_metric}).

\begin{figure*}[tb]
\centering
\setlength{\tabcolsep}{1pt}
\renewcommand{\arraystretch}{1}

\small
\begin{tabular}{@{}c@{\hspace{2pt}}c@{\hspace{2pt}}c@{\hspace{2pt}}c@{\hspace{2pt}}c@{}}
& \textbf{Frame 20} & \textbf{Frame 40} & \textbf{Frame 60} & \textbf{Frame 80} \\[2pt]

\rotatebox{90}{\hspace{1em}\small Original} &
\includegraphics[width=0.14\linewidth]{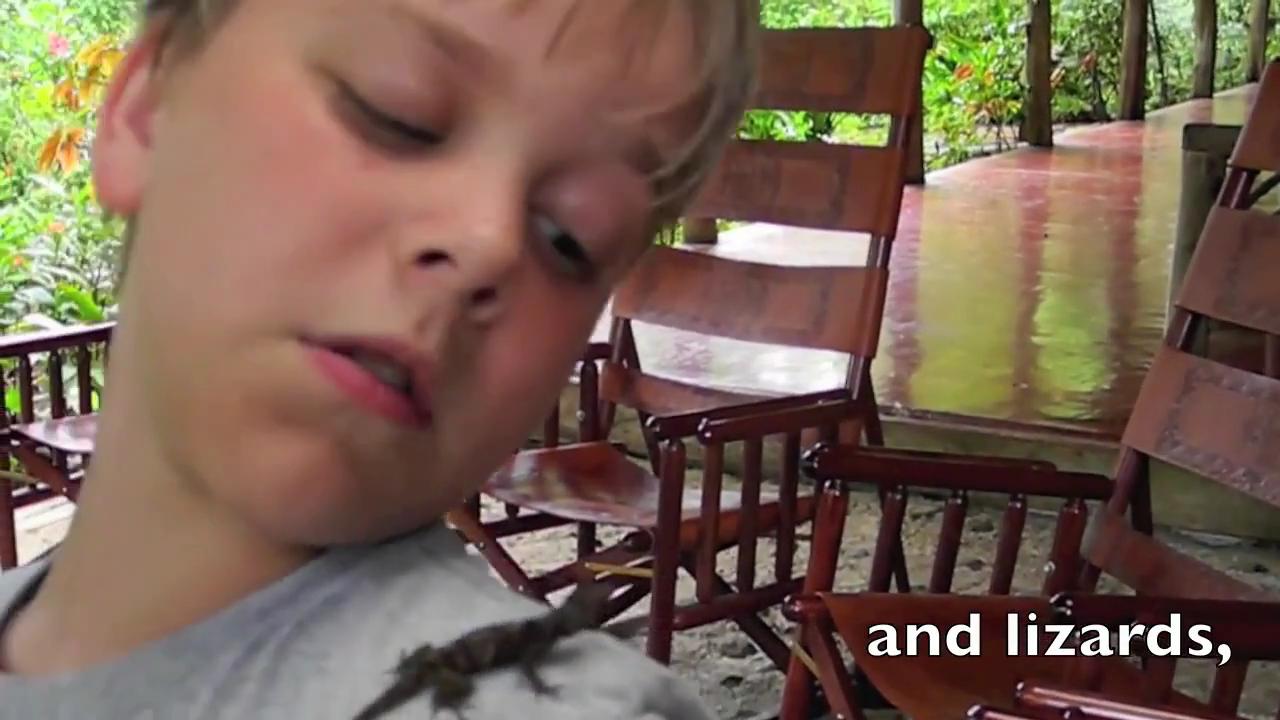} &
\includegraphics[width=0.14\linewidth]{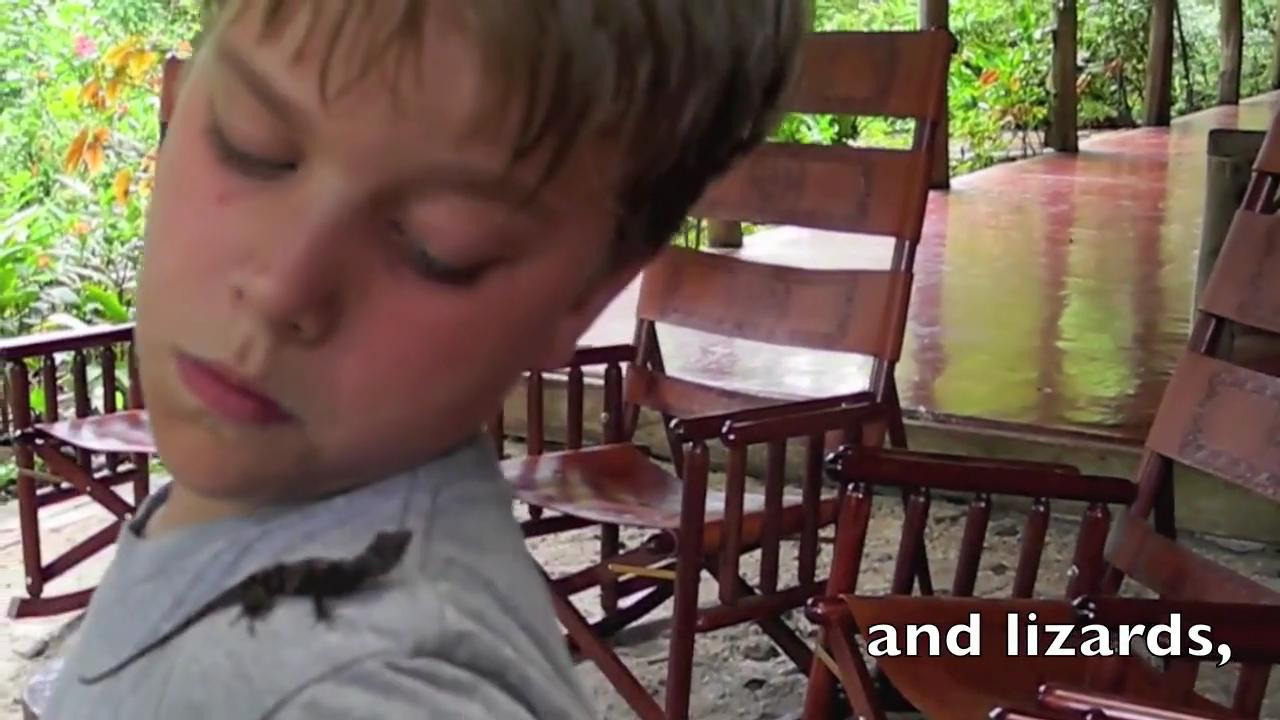} &
\includegraphics[width=0.14\linewidth]{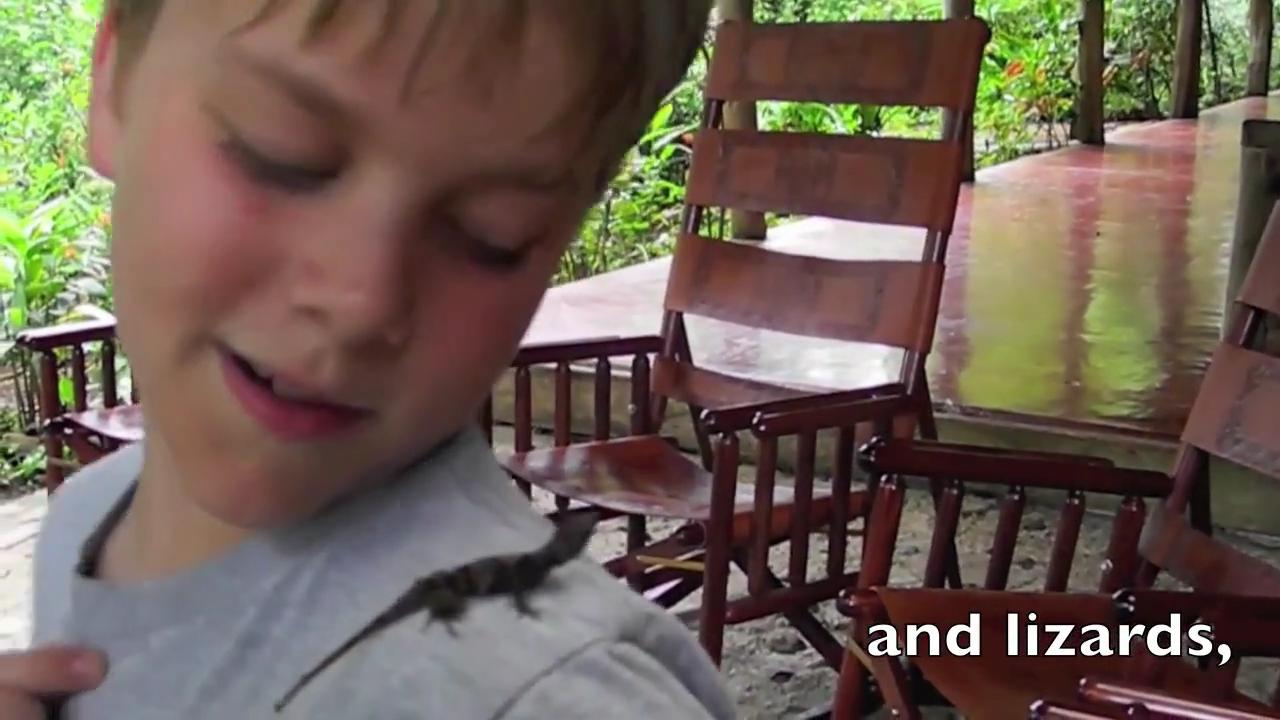} &
\includegraphics[width=0.14\linewidth]{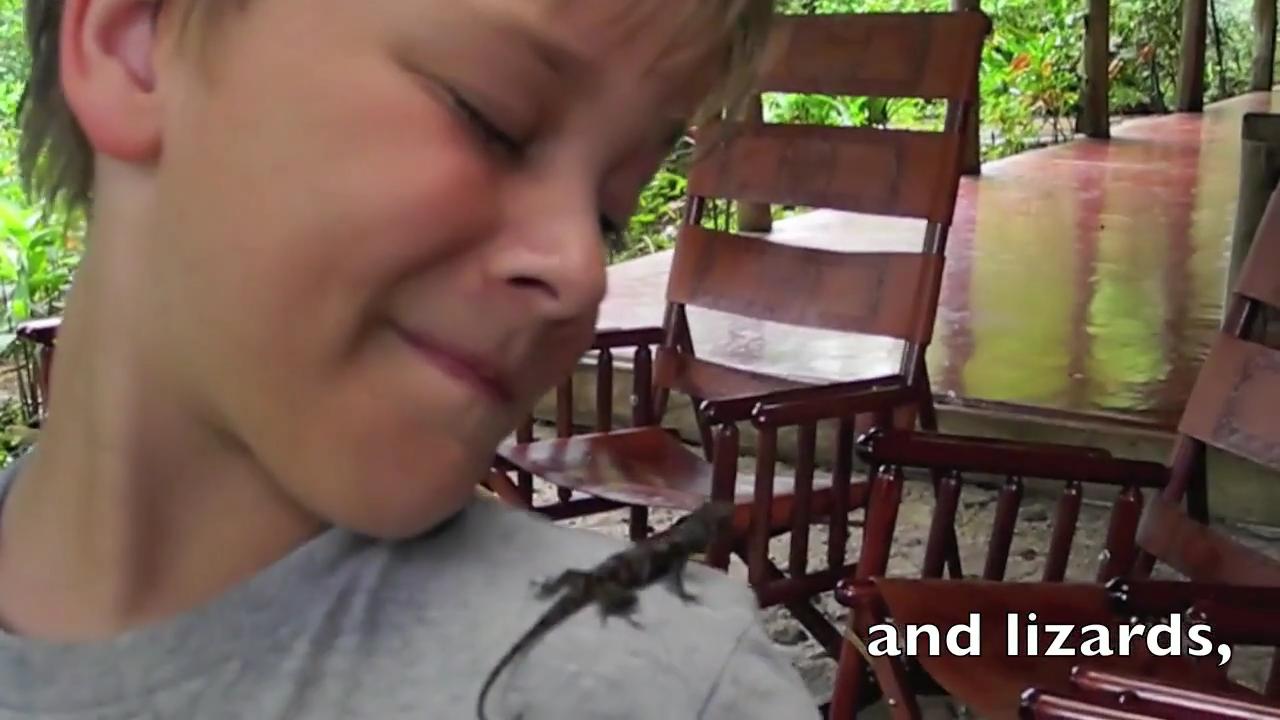} \\[2pt]

\rotatebox{90}{\hspace{0.5em}\small SD Baseline} &
\includegraphics[width=0.14\linewidth]{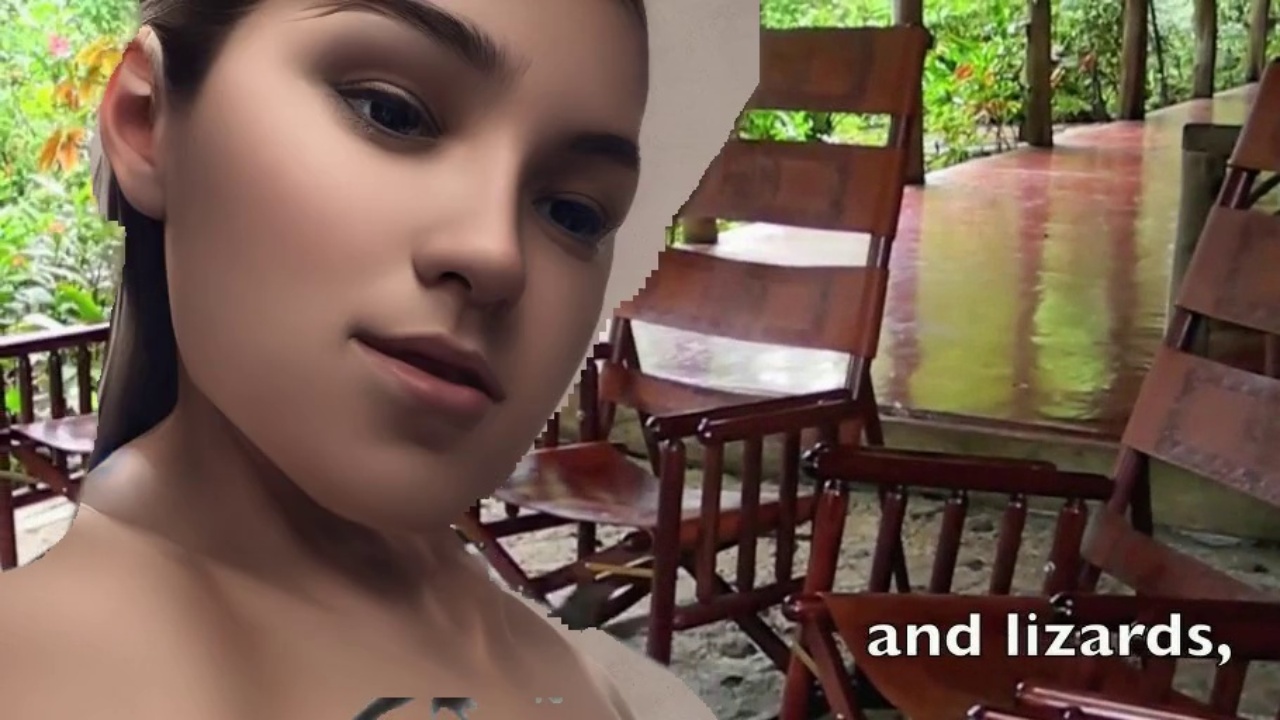} &
\includegraphics[width=0.14\linewidth]{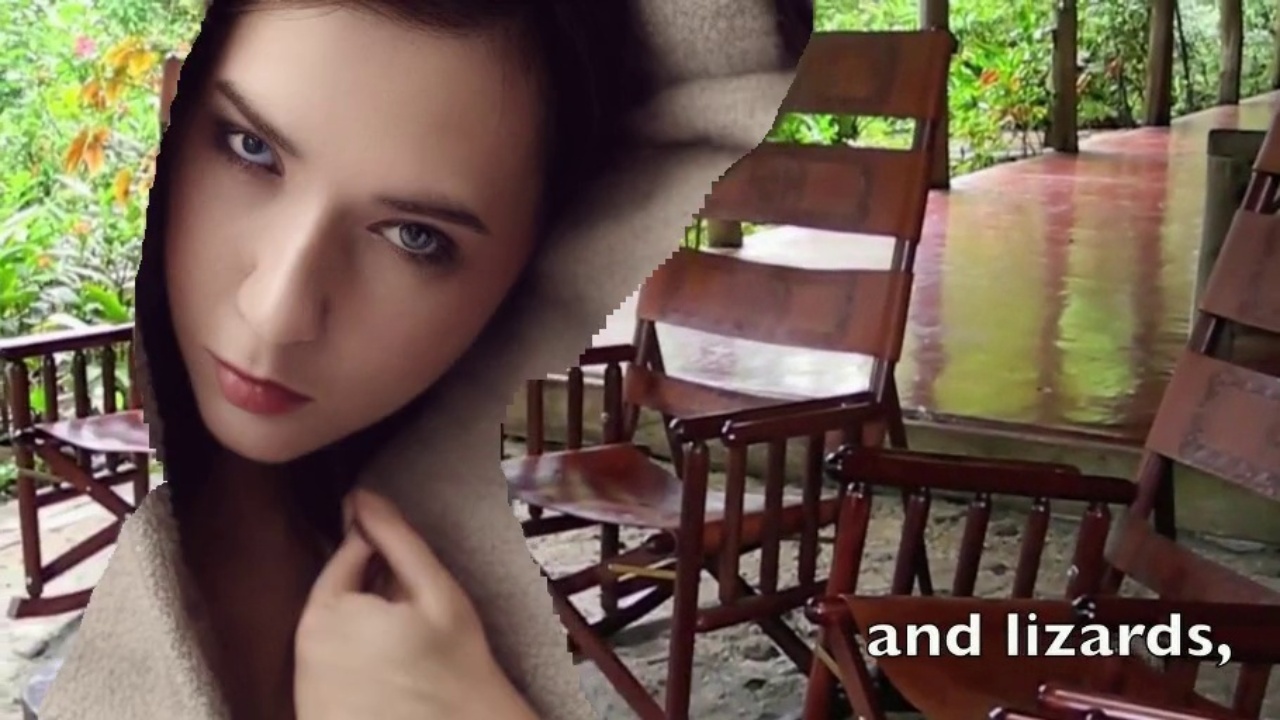} &
\includegraphics[width=0.14\linewidth]{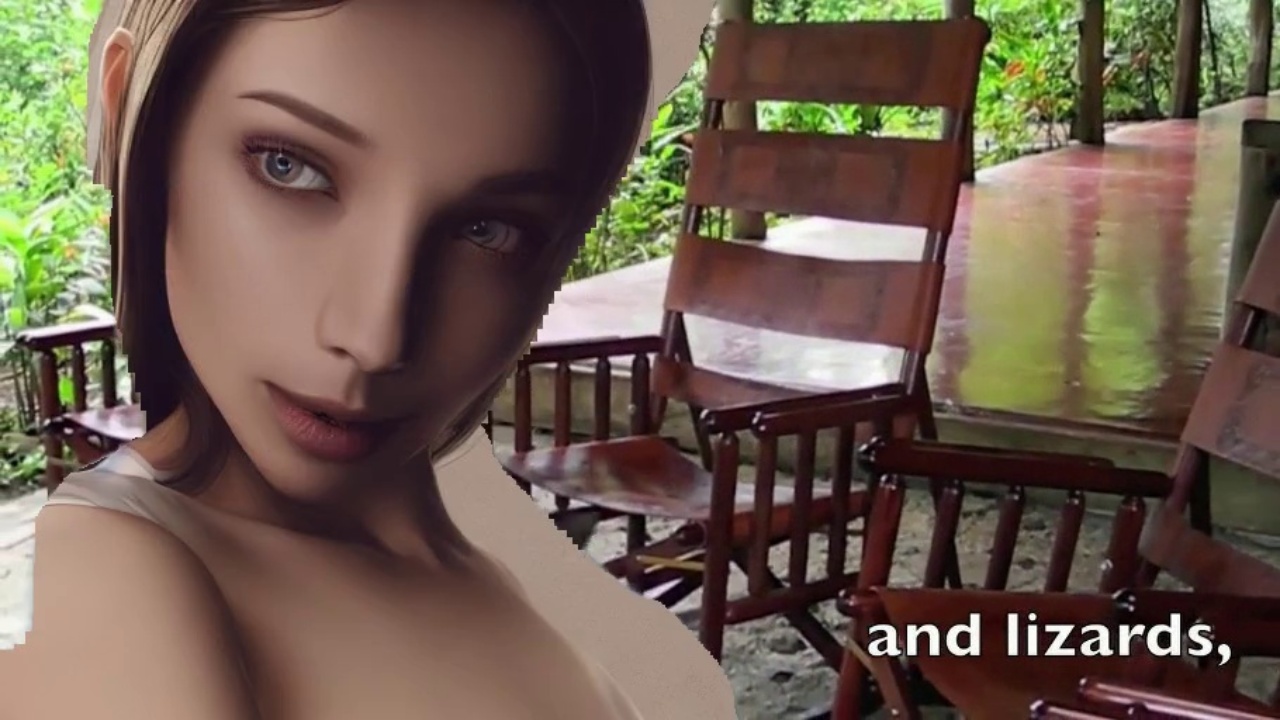} &
\includegraphics[width=0.14\linewidth]{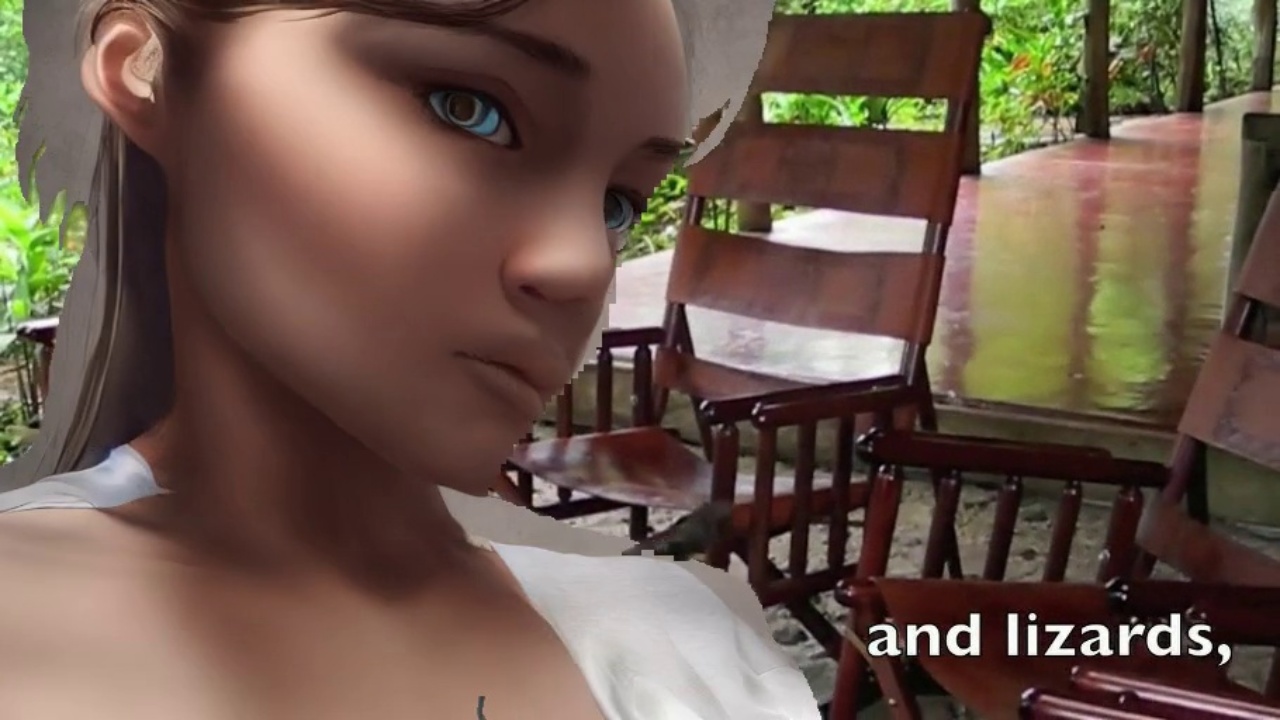} \\[2pt]

\rotatebox{90}{\hspace{1.5em}\small DP2} &
\includegraphics[width=0.14\linewidth]{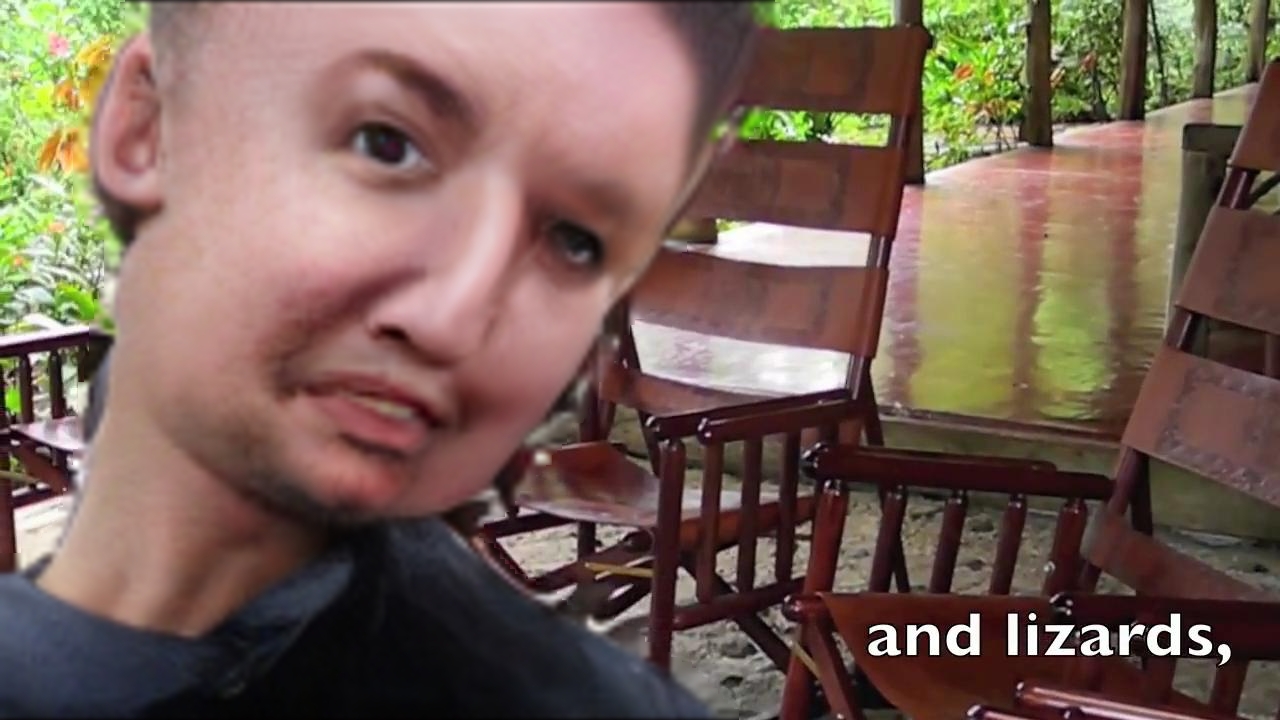} &
\includegraphics[width=0.14\linewidth]{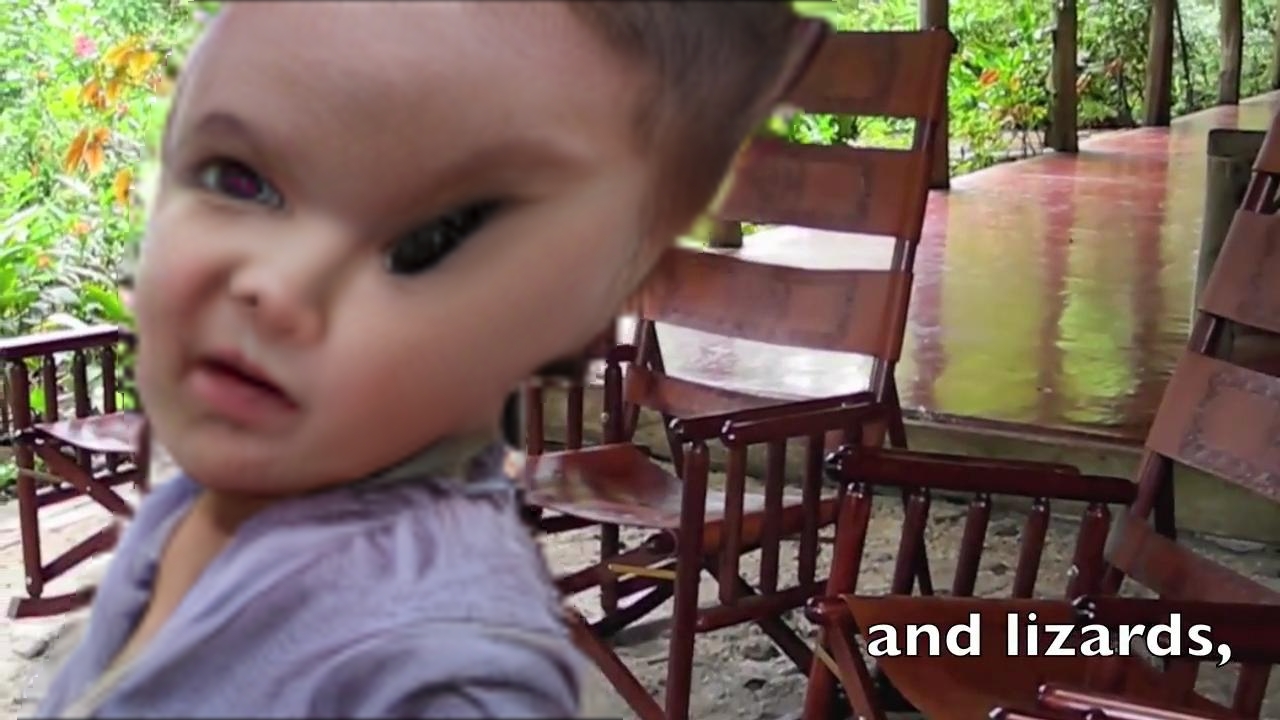} &
\includegraphics[width=0.14\linewidth]{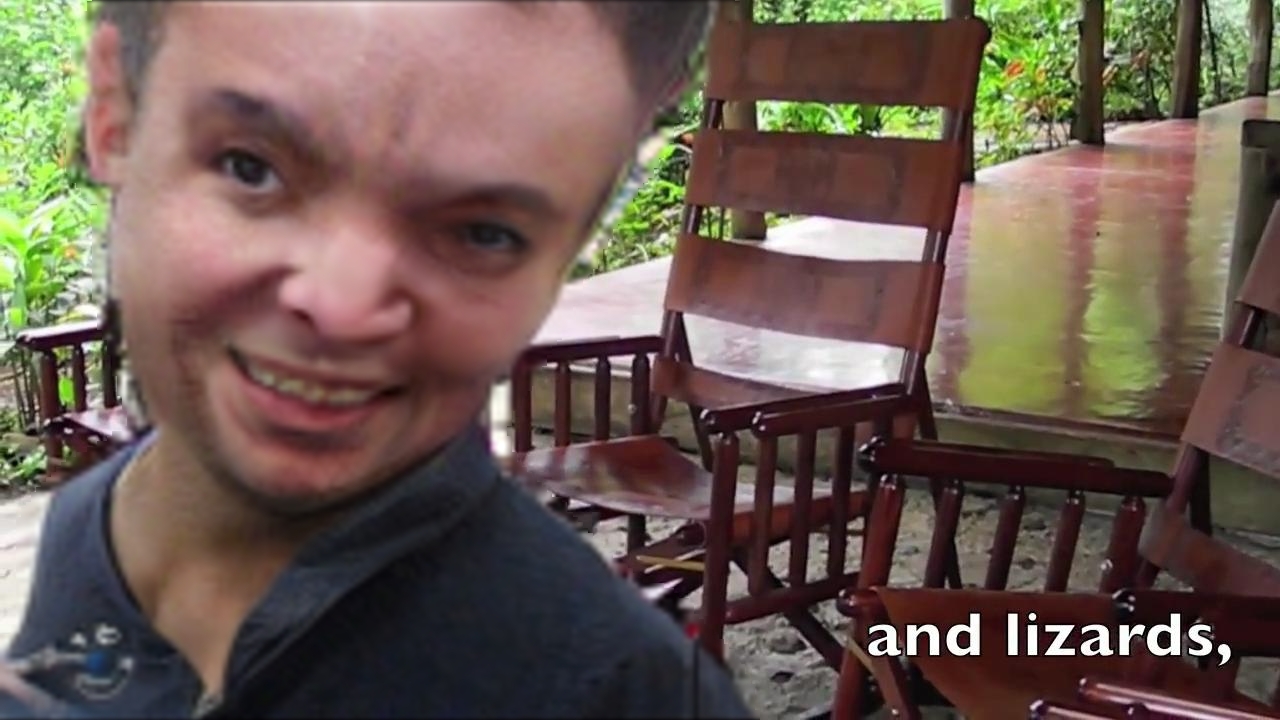} &
\includegraphics[width=0.14\linewidth]{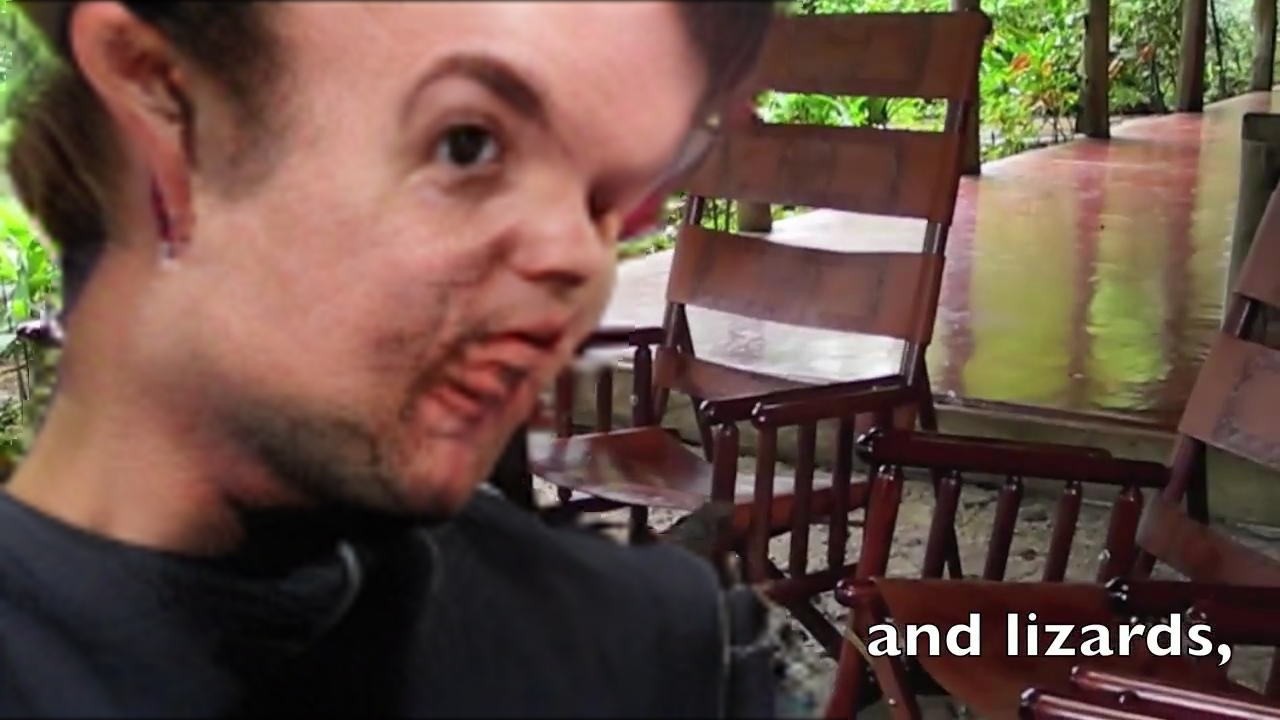} \\[2pt]

\rotatebox{90}{\hspace{0.5em}\small \textbf{\textcolor{blue}{Ours}}} &
\fcolorbox{blue}{white}{\includegraphics[width=0.135\linewidth]{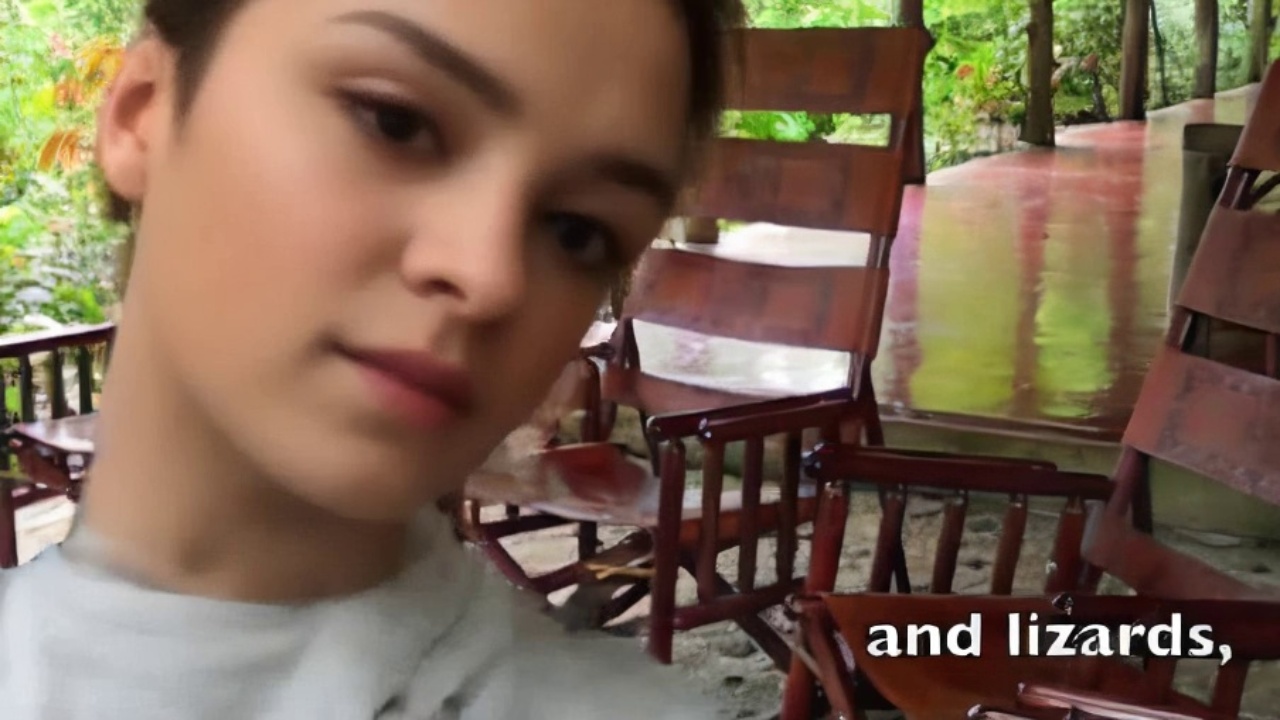}} &
\fcolorbox{blue}{white}{\includegraphics[width=0.135\linewidth]{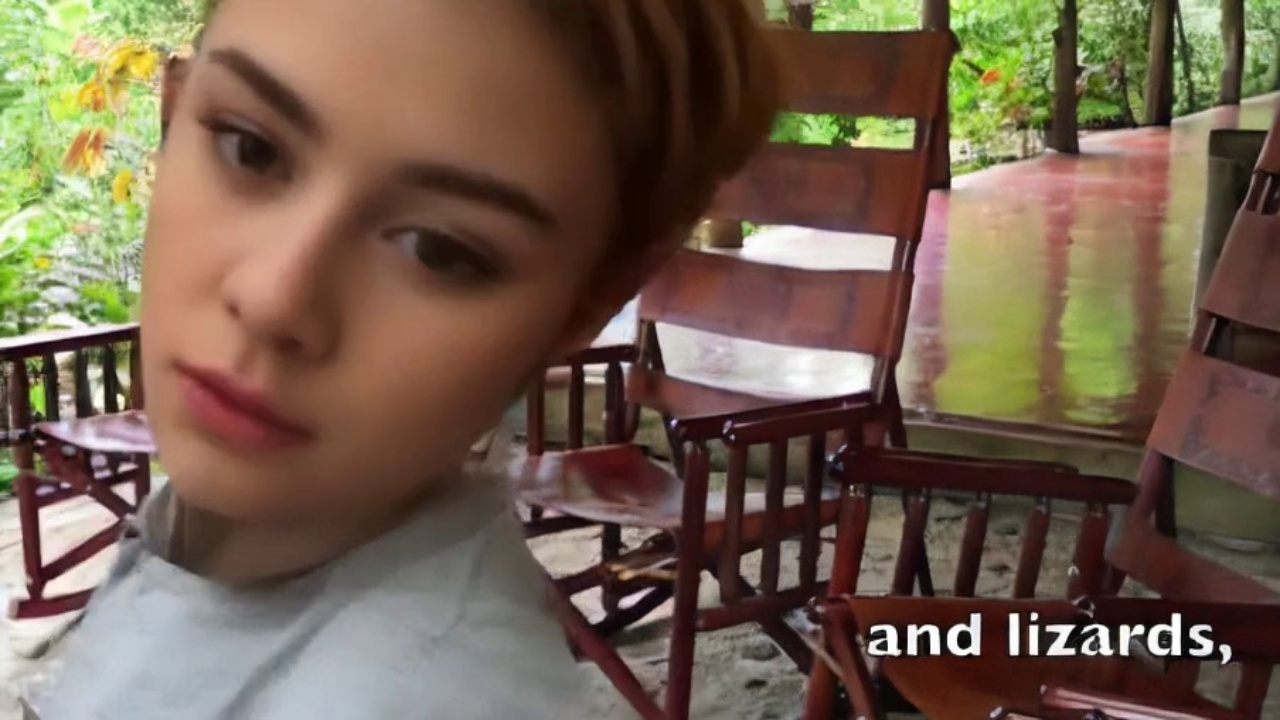}} &
\fcolorbox{blue}{white}{\includegraphics[width=0.135\linewidth]{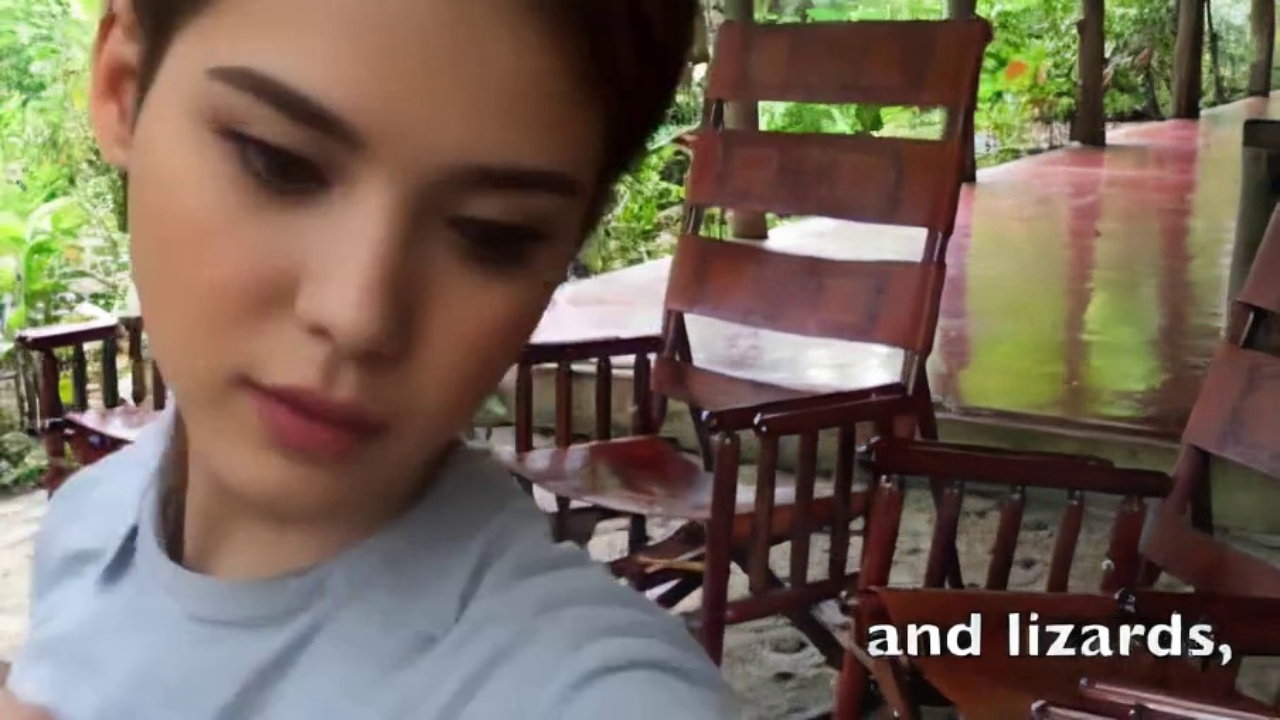}} &
\fcolorbox{blue}{white}{\includegraphics[width=0.135\linewidth]{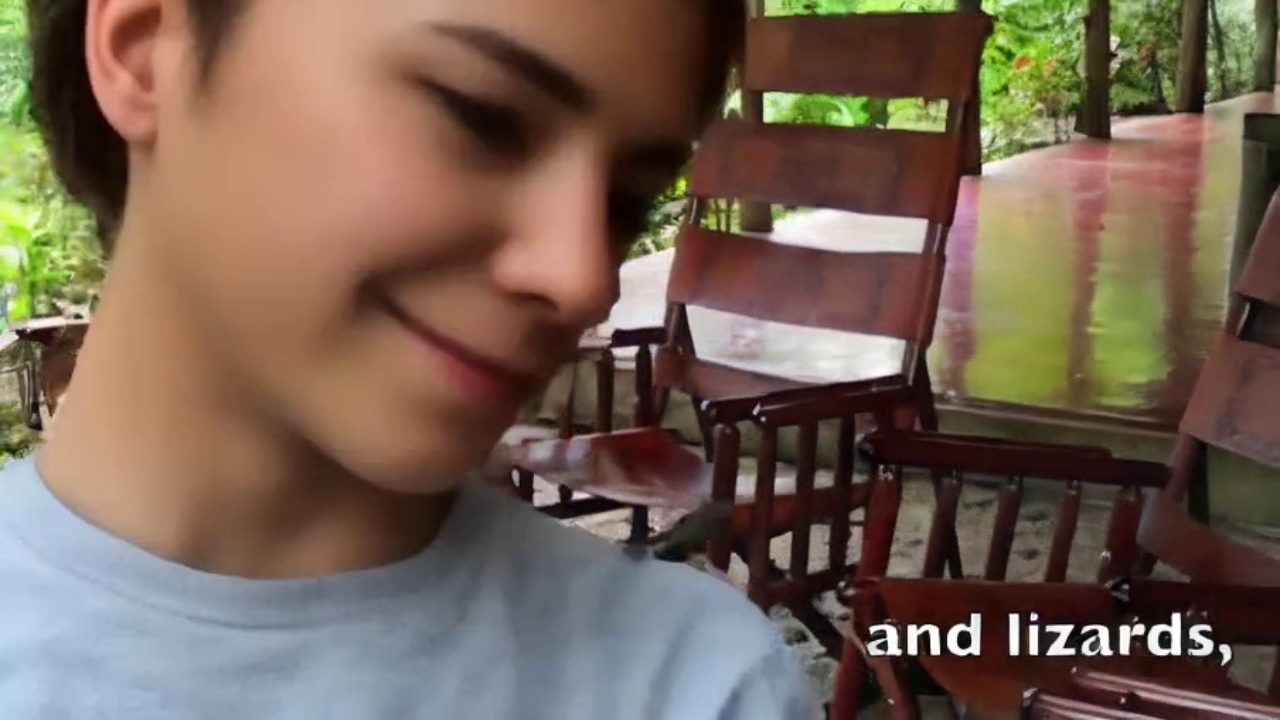}} \\

\end{tabular}

\caption{\textbf{Qualitative comparison on YTVIS19.} We compare our anonymization against original data, SD baseline, and DP2. SD Baseline suffers from unnatural foreground-background separation due to pixel inpainting, while DP2 distorts the structure of the child's face in the video. In contrast, our method produces temporally and structurally consistent results compared to existing baselines. Best viewed zoomed in.}
\label{fig:qualitative_comparison}
\end{figure*}

\paragraph{Latent-Space Inpainting.}
Our objective is to anonymize the foreground subject while preserving the original background, as the scene context is often important for downstream tasks. Existing anonymization methods typically perform inpainting in pixel space by generating anonymized humans separately and then compositing them onto the original background~\cite{hukkelaas2023deepprivacy2, malm2023rad, zwick2024context}. Such pixel-space compositing can introduce boundary artifacts and unrealistic blending between foreground and background, as illustrated in the SD baseline example in \cref{fig:qualitative_comparison}. Moreover, when multiple overlapping people are present, pixel-space methods often require recursive compositing with careful ordering to resolve occlusions~\cite{hukkelaas2023deepprivacy2}.

To avoid these issues, we perform inpainting directly in the latent space throughout the diffusion denoising process~\cite{avrahami2023blended}. At each timestep $t$, the latent of the generated frame $\hat{x}_{t-1}$ is blended with the latent representation of the original frame using the binary human segmentation mask $M_i$:
\[
x_{t-1} = M_i \odot \hat{x}_{t-1} + (1 - M_i) \odot x_{t-1}.
\]
Here, $x_{t-1}$ denotes the latent representation of the original image at noise level $t-1$, and $\hat{x}_{t-1}$ is the generated latent at the same timestep. Because both latents share the same noise level during the denoising process, the foreground and background remain consistent throughout generation, resulting in natural and seamless integration. The original background is retained exactly wherever $M_i$ is accurate.

\section{Experiments \& Results}
\label{sec:exp-results}

\subsection{Experimental Details}
\label{sec:exp-details}
\noindent \textbf{Baselines:} We compare our anonymization method with original data, DP2~\cite{hukkelaas2023deepprivacy2}, Stable Animator~\cite{tu2025stableanimator}, StreamV2V~\cite{liang2024looking} and a stable diffusion (SD) baseline. For the SD Baseline, we implement a naive pixel inpainting with ControlNet conditioning. For Stable Animator baseline, which animates a reference image according to an input pose sequence, we first anonymize the first video frame using SD baseline and use this anonymized frame as the reference image. We use the same mask and pose sequence across all applicable methods.

\noindent \textbf{Tasks/Datasets:} We evaluate three downstream tasks:
(1) \emph{Video instance segmentation} on YTVIS19~\cite{yang2019video},
(2) \emph{Video action recognition} on UCF101~\cite{soomro2012ucf101}, and
(3) \emph{Person re-identification} on MSMT17~\cite{wei2018person} and MARS~\cite{zheng2016mars}.
These tasks were chosen because they are sensitive to temporal inconsistencies and therefore penalize methods such as DP2, which do not generate smooth, coherent videos. We use C3D~\cite{tran2015learning} for UCF101, MaskTrackRCNN~\cite{yang2019video} for YTVIS19, TransReID~\cite{he2021transreid} for MSMT17, and VID-Trans-ReID~\cite{alsehaim2022vid} for MARS. For YTVIS19, we anonymize only the frames containing humans (based on ground-truth masks) for all baselines, leaving non-human frames unchanged. This ensures a fair comparison with DP2, which anonymizes frames only when it detects human masks. Across all tasks we anonymize only the training split and evaluate the resulting model on the original test split (see Fig.~10 of the supplementary).

\noindent \textbf{Metric:} For YTVIS19, we use mAP as the metric, where IoU calculation considers mask video tracklets~\cite{yang2019video}. For UCF101 we report top-1 accuracy~\cite{soomro2012ucf101} in (\%). For Re-ID, we report mAP and rank-1 accuracy~\cite{he2021transreid, alsehaim2022vid}. We state our threat model, the adversary's capabilities, and the scope of our privacy claim in Sec.~8 of the supplementary.%

\subsection{Video Instance Segmentation}
\label{sec:vis19}
\begin{table}[t]
  \centering
  \footnotesize
  \begin{tabular}{lcc}
    \toprule
    Method & mAP (overall) & person AP \\
    \midrule
    \multicolumn{3}{l}{Default Tracker} \\
    \midrule
    Original & 28.20 & 44.18 \\
    DP2      & 27.60 & 34.38 \\
    Stable Animator & 19.20 & 20.26 \\
    StreamV2V & 19.60 & 6.27 \\
    SD Baseline & 29.90 & 32.00 \\
    \rowcolor{gray!20}\textbf{Ours} & 27.30 & \textbf{34.78} \\
    \midrule
    \multicolumn{3}{l}{Only Appearance Cues} \\
    \midrule
    Original & 17.30 & 36.78 \\
    DP2      & 17.10 & 26.45 \\
    \rowcolor{gray!20}\textbf{Ours} & \textbf{17.70} & \textbf{27.50} \\
    \bottomrule
  \end{tabular}

  \vspace{2pt}
  \caption{Video instance segmentation utility on YTVIS19 val using MaskTrackRCNN trained on original or anonymized training videos. We report overall mAP and AP for the \texttt{person} class under the default tracker, which combines appearance, detection score, bbox IoU, and class-label cues, and under an appearance-only matching setting.}
  \label{tab:main_results}
\end{table}

\begin{table}[t]
  \centering
  \footnotesize
  \begin{tabular}{lccc}
    \toprule
    \textbf{Method} & \textbf{Full Dataset} & \textbf{FG Only} & \textbf{Subset (FG)} \\
    \midrule
    Original        & 81.31 & 64.13 & 40.49 \\
    \midrule
    \rowcolor{gray!20}\textbf{Ours} & 77.50 & \textbf{57.97} & \textbf{35.74} \\
    DP2             & 77.90 & 55.67 & 33.10 \\
    SD Baseline     & ---   & ---   & 32.75 \\
    StreamV2V       & ---   & ---   & 22.89 \\
    \bottomrule
  \end{tabular}

  \vspace{2pt}
  \caption{Action recognition performance (Top-1 Accuracy \%) on UCF101 Split 1 using C3D models trained on anonymized data. Models are evaluated on the original validation set. \textbf{Full Dataset:} Results on complete UCF101 (13k videos). \textbf{Foreground Only:} Background removed using detection masks, isolating foreground content. \textbf{Subset:} Representative 15\% subset (2,052 videos) with foreground only, used for computational efficiency when comparing against all baselines.}
  \label{tab:ucf101_results}
\end{table}

\begin{figure}[t]
  \centering
  \includegraphics[width=0.82\linewidth]{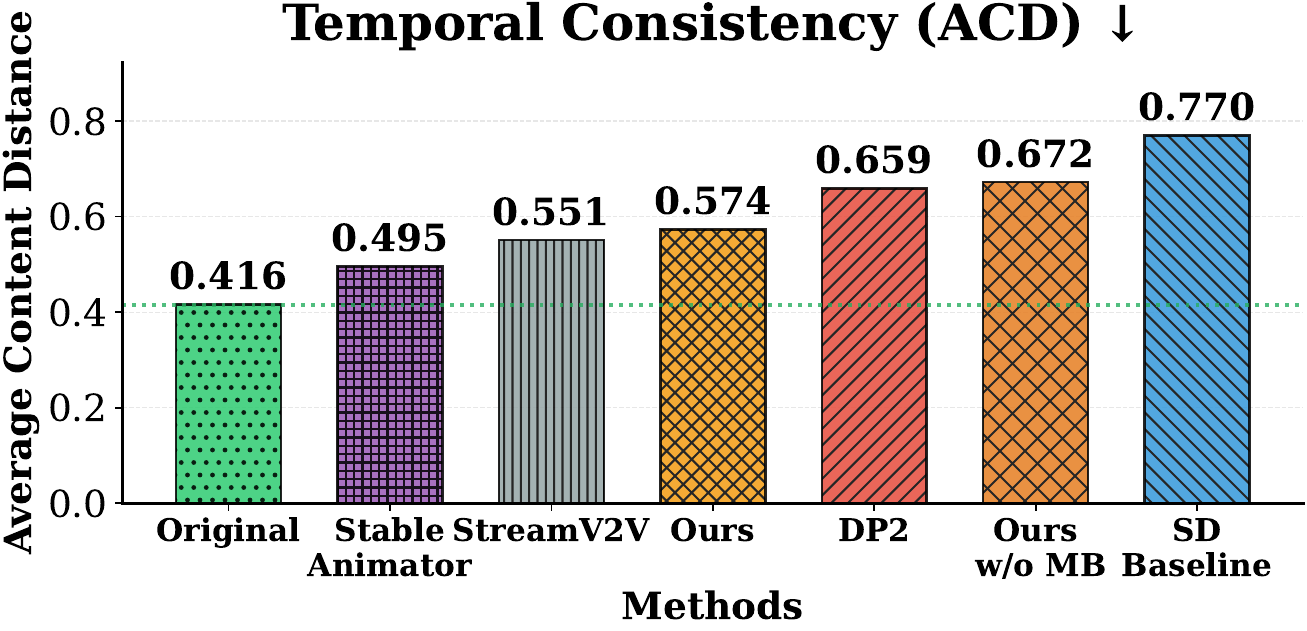}
  \caption{Temporal consistency measured using ACD~($\downarrow$)~\cite{wang2025leo} on YTVIS19 using a pre-trained Re-ID model~\cite{he2021transreid}.}
  \label{fig:acd_metric}
\end{figure}

\cref{tab:main_results} reports YTVIS19 val results using MaskTrackRCNN trained on different anonymized versions of the train set. The default tracker combines appearance similarity with detection score, bbox IoU, and predicted class label~\cite{yang2019video}. Unsurprisingly the model trained on original data performs best, yet DP2 performs remarkably well despite strong temporal inconsistencies, as does our SD baseline, which generates each frame independently. Restricting the tracker to appearance cues only, our anonymization (17.70) outperforms both DP2 (17.10) and the original data (17.30) in overall mAP. DP2's strong showing here is likely explained by only positive proposals contributing to tracker training~\cite{yang2019video}, so the instance embedding space is never trained for repulsion between distinct instances; pre-trained detection weights further bias the task toward per-frame detection over temporal consistency. Consistently, the SD baseline reaches 29.90 overall mAP despite very poor temporal consistency (\cref{fig:acd_metric}); we attribute this to small gains on the non-anonymized non-human classes while \texttt{person} AP decreases. Of the 3,471 training videos, only 647 contain the person class and only 27 contain more than one person.~\cref{fig:qualitative_comparison} shows qualitative comparisons against DP2~\cite{hukkelaas2023deepprivacy2} and the SD Baseline; more are provided in the supplementary material.

\begin{figure}[t!]
    \centering
    \setlength{\tabcolsep}{1pt}
    \renewcommand{\arraystretch}{0.8}
    \resizebox{0.9\linewidth}{!}{%
    \begin{tabular}{@{}c@{\hspace{1pt}}c@{\hspace{1pt}}c@{\hspace{1pt}}c@{\hspace{1pt}}c@{\hspace{8pt}}c@{\hspace{1pt}}c@{\hspace{1pt}}c@{\hspace{1pt}}c@{\hspace{1pt}}c@{}}
        & \scriptsize Frame 1 & \scriptsize Frame 2 & \scriptsize Frame 3 & \scriptsize Frame 4 &
        & \scriptsize Frame 1 & \scriptsize Frame 2 & \scriptsize Frame 3 & \scriptsize Frame 4 \\
        \cmidrule(lr){2-5} \cmidrule(lr){7-10}

        \scriptsize\textbf{(a)} &
        \includegraphics[width=0.11\linewidth]{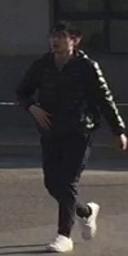} &
        \includegraphics[width=0.11\linewidth]{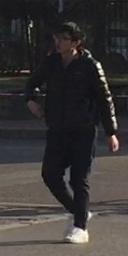} &
        \includegraphics[width=0.11\linewidth]{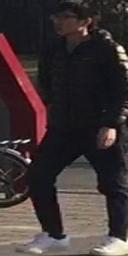} &
        \includegraphics[width=0.11\linewidth]{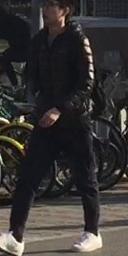} &

        \scriptsize\textbf{(b)} &
        \includegraphics[width=0.11\linewidth]{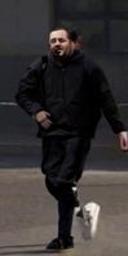} &
        \includegraphics[width=0.11\linewidth]{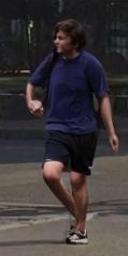} &
        \includegraphics[width=0.11\linewidth]{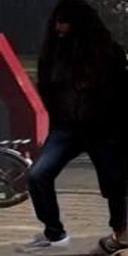} &
        \includegraphics[width=0.11\linewidth]{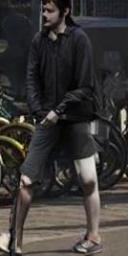} \\[2pt]

        \scriptsize\textbf{(c)} &
        \includegraphics[width=0.11\linewidth]{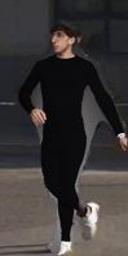} &
        \includegraphics[width=0.11\linewidth]{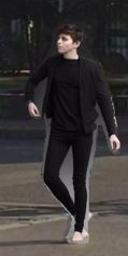} &
        \includegraphics[width=0.11\linewidth]{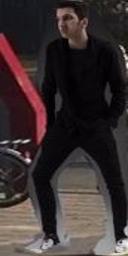} &
        \includegraphics[width=0.11\linewidth]{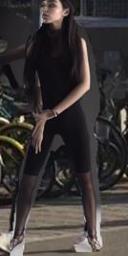} &

        \scriptsize\textbf{(d)} &
        \includegraphics[width=0.11\linewidth]{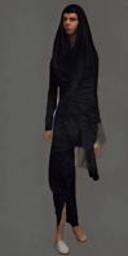} &
        \includegraphics[width=0.11\linewidth]{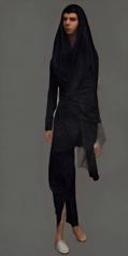} &
        \includegraphics[width=0.11\linewidth]{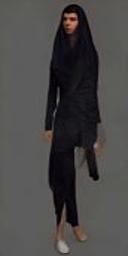} &
        \includegraphics[width=0.11\linewidth]{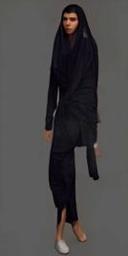} \\[2pt]

        \scriptsize\textbf{(e)} &
        \includegraphics[width=0.11\linewidth]{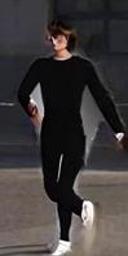} &
        \includegraphics[width=0.11\linewidth]{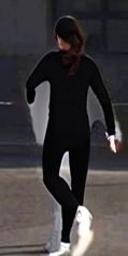} &
        \includegraphics[width=0.11\linewidth]{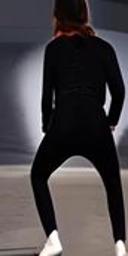} &
        \includegraphics[width=0.11\linewidth]{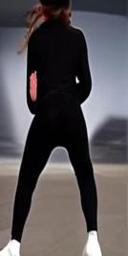} &

        \scriptsize\textbf{(f)} &
        \includegraphics[width=0.11\linewidth]{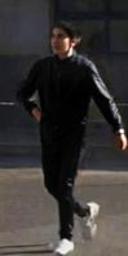} &
        \includegraphics[width=0.11\linewidth]{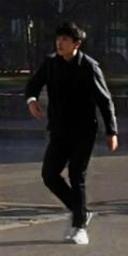} &
        \includegraphics[width=0.11\linewidth]{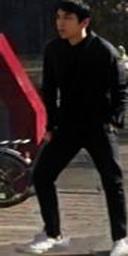} &
        \includegraphics[width=0.11\linewidth]{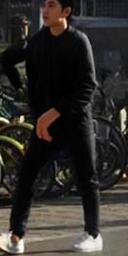} \\
    \end{tabular}%
    }
    \caption{Qualitative comparison of anonymization methods across temporal frames on MSMT17. (a)~Original, (b)~DeepPrivacy2, (c)~SD Baseline, (d)~StreamV2V, (e)~Stable Animator, (f)~\textbf{Ours}. Best viewed zoomed in.}
    \label{fig:qualitative_comparison_msmt17}
\end{figure}

\subsection{Video Action Recognition}
\label{sec:ucf101}

\cref{tab:ucf101_results} compares our method with the existing anonymization baselines in the UCF101 dataset using the C3D network architecture trained on Split 1. In the full UCF101 dataset, our method achieves 77.50\% top-1 acc compared to 81.31\% for the original data, performing comparably to DP2 (77.90\%). We hypothesize that this small gap between methods is due to strong background cues in UCF101 videos, which is preserved similarly across anonymization approaches despite differences in temporal consistency and visual quality of the foreground. For example, the model might simply predict higher class scores for categories like `GolfSwing' and `CricketShot', when it sees the green playground/pitch. Thus, to isolate the impact of foreground anonymization quality, we conduct background removal experiments, using human detection masks to retain only foreground content while replacing backgrounds with black pixels.
In this \textit{foreground-only setting}, our method demonstrates clearer advantages: on the full dataset, we achieve 57.97\% accuracy compared to DP2's 55.67\% (original: 64.13\%). Thus, the temporally consistent and smoother videos from our method enable gains in video action recognition in the foreground-only setting; however, the gap is still not high between DP2 and ours, suggesting that the downstream model might be relying on other cues to predict action class. Due to computational constraints, we evaluate additional baselines in a representative 15\% subset (2,052 videos) of UCF101 with foreground-only data. Table~\ref{tab:ucf101_results} shows that our method achieves 35.74\% accuracy, outperforming DP2 (33.10\%), SD Baseline (32.75\%), and StreamV2V (22.89\%).

\subsection{Re-Identification}
\label{sec:re-id}

\subsubsection{MSMT17 Dataset}
\begin{figure}[t]
  \centering
  \begin{minipage}[b]{0.58\linewidth}
    \centering
    \includegraphics[width=\linewidth]{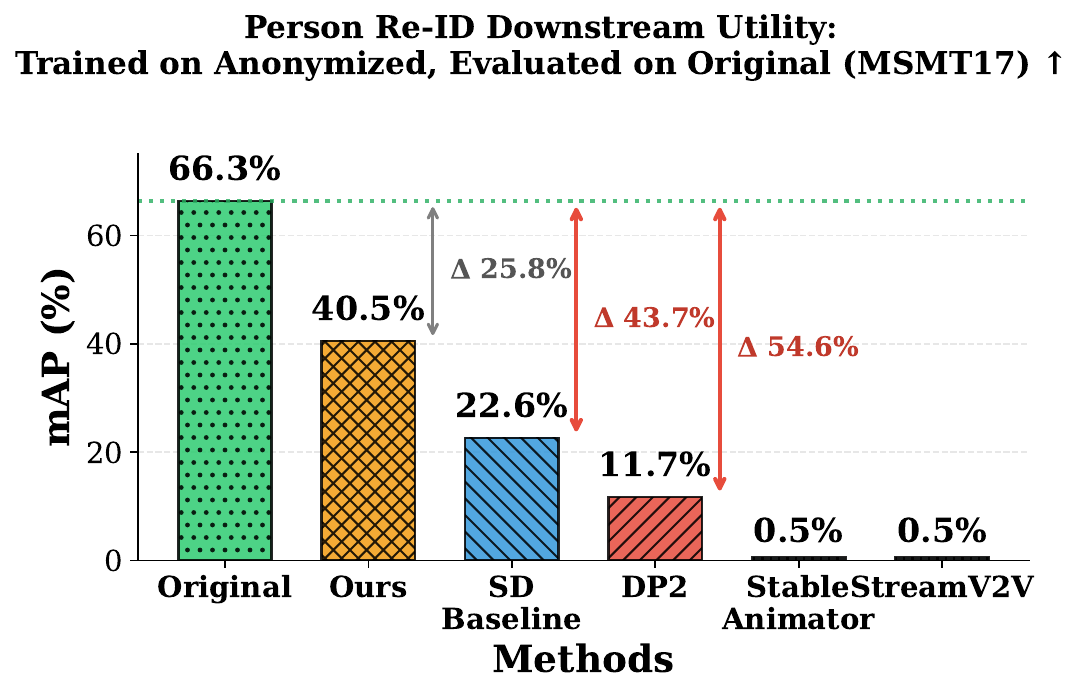}
    \captionof{figure}{Person Re-ID utility on MSMT17~\cite{wei2018person}. We train TransReID~\cite{he2021transreid} on each anonymized training set and evaluate on the original \{query, gallery\} split, reporting the downstream mAP gap to the original-data upper bound.}
    \label{fig:msmt_reid_results}
  \end{minipage}\hfill
  \begin{minipage}[b]{0.39\linewidth}
    \centering
    \resizebox{\linewidth}{!}{%
    \begin{tabular}{@{}ccccc@{}}
          \toprule
          IM & \begin{tabular}[c]{@{}c@{}}Lat.\\Inpnt\end{tabular} & \begin{tabular}[c]{@{}c@{}}Cntrl\\Net\end{tabular} & MB & mAP \\
          \midrule
          \cmark & \cmark & \cmark & \cmark & 40.5 \\
          \xmark & \cmark & \cmark & \cmark & 21.1 \\
          \cmark & \xmark & \cmark & \cmark & 41.6 \\
          \cmark & \cmark & \xmark & \cmark & 30.9 \\
          \cmark & \cmark & \cmark & \xmark & 34.6 \\
          \bottomrule
        \end{tabular}%
    }
    \captionof{table}{Ablation of downstream utility on MSMT17. We report mAP when removing individual components: injection module (IM), latent inpainting (Lat. Inpnt), ControlNet (CntrlNet), and memory bank (MB).}
    \label{tab:ablation-comp}
  \end{minipage}
\end{figure}

\begin{figure}[t]
  \centering
  \includegraphics[width=0.9\linewidth]{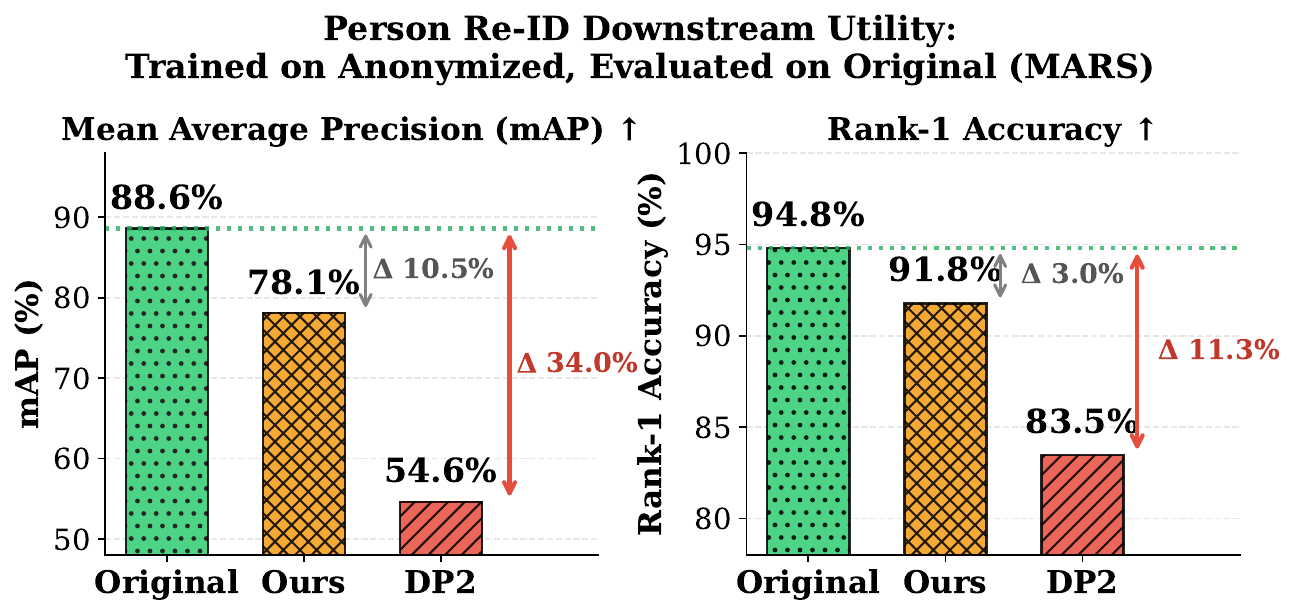}
  \caption{Person Re-ID utility on MARS~\cite{zheng2016mars}. We train VID-Trans-ReID~\cite{alsehaim2022vid} on anonymized training videos and evaluate on the original \{query, gallery\} split, reporting both mAP and rank-1 accuracy.}
  \label{fig:mars_reid_results}
\end{figure}

From \cref{fig:msmt_reid_results}, we can see that our anonymization method has the lowest performance gap compared to the upper baseline of the original data. Further, we observe a significant gap between our method and existing baselines on this task. We attribute this gap to the relatively better temporal consistency, visual quality, and similarity to the input visual domain of our generations compared to the baselines, as seen in~\cref{fig:qualitative_comparison_msmt17} (refer to the supplementary for more examples). From the figure, we can see that both StreamV2V and Stable Animator fail to preserve the original input background and suffer from low visual quality generations, explaining their poor downstream utility on the Re-ID task. In the case of Stable Animator, since we directly use their pre-trained weights, the MSMT data is out of distribution, while for StreamV2V, the lack of our latent inpainting, ControlNet, and injection modules leads to low-quality outputs.

\subsubsection{MARS Dataset}

\begin{figure}[t]
  \centering
  \includegraphics[width=0.95\linewidth]{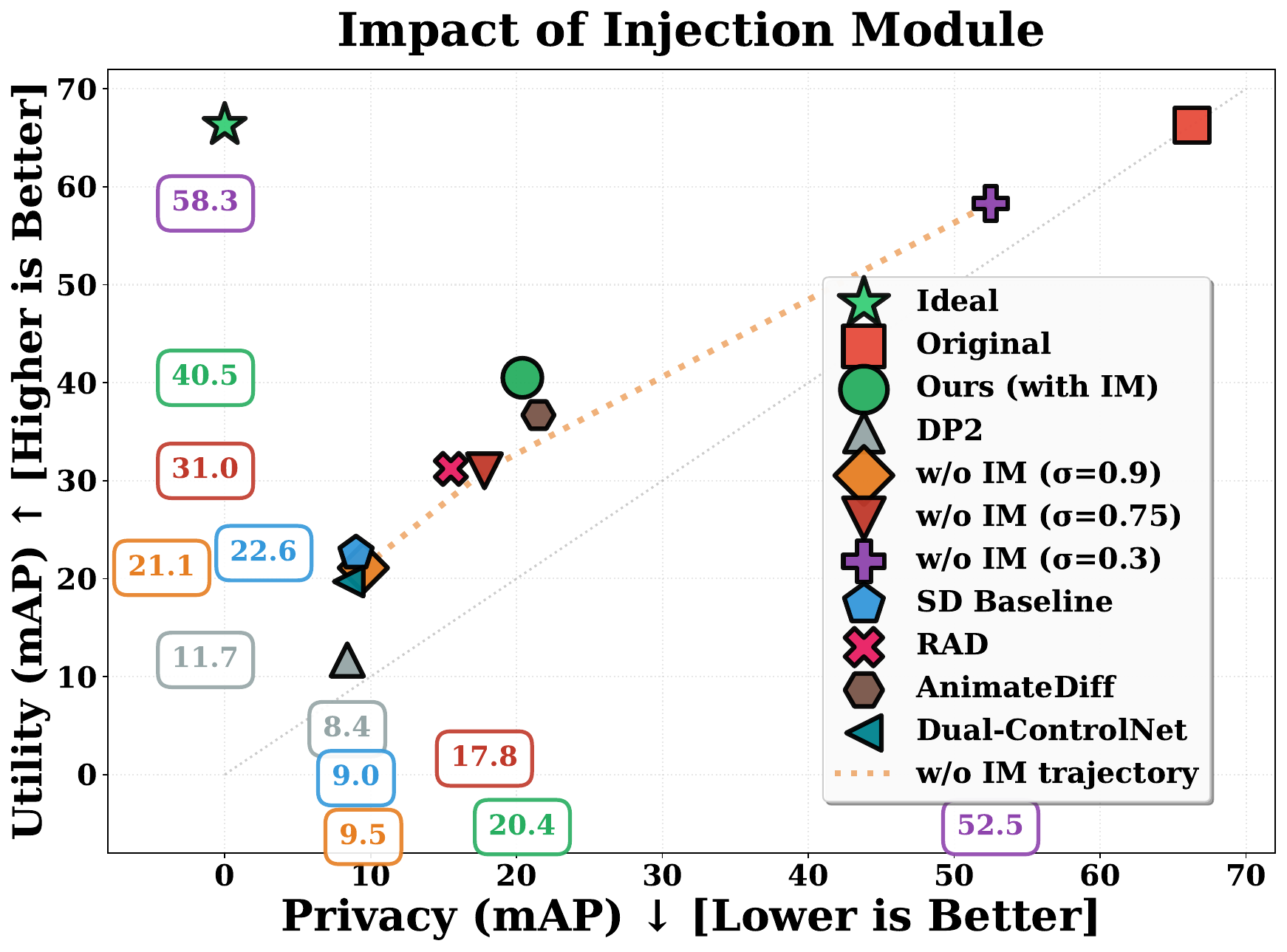}

  \caption{\textbf{Utility-Privacy Trade-off Analysis.} Privacy is measured by evaluating a standard Re-ID model~\cite{he2021transreid} on the anonymized query set (lower mAP indicates stronger privacy). Utility is measured by training the Re-ID model on the anonymized training set and evaluating it on the original \{query, gallery\} set (higher mAP indicates better utility).}
  \label{fig:privacy_utility}
\end{figure}

Our downstream utility results on the MARS dataset are shown in \cref{fig:mars_reid_results}, where we present both mAP and rank-1 accuracy metrics. We observe a notably small performance gap between models trained on original data versus on data anonymized by our method. For rank-1 accuracy, the gap is merely 3\%, demonstrating a strong preservation of utility. However, DP2 achieves surprisingly high performance at 54.6\% mAP and 83.5\% rank-1 accuracy. We hypothesize that this may result from the Re-ID model exploiting consistent generation artifacts as discriminative features. Additionally, this finding may be partially attributed to MARS being an older dataset with lower image quality and less complex lighting variations compared to MSMT17.

\subsection{Ablation Study}
\label{sec:privacy-utility}
We use $\sigma_{g}=0.9$ in the generation branch and $\sigma_{i}=0.3$ in the injection branch. As shown in \cref{fig:privacy_utility}, removing IM and varying the noise ($\sigma=\sigma_{g}$) does not produce a better privacy–utility trade-off than our full method. Further, at a similar privacy level, our method achieves a +9.5 mAP utility gain over the variant without IM, showing that the gains are due to our proposed IM and not simply due to noise tuning. From \cref{fig:privacy_utility}, we also note that StablePrivacy achieves a much better utility-privacy trade-off than DP2 and the SD Baseline. The same figure places three further baselines on this curve, all evaluated under our exact protocol: RAD~\cite{malm2023rad} at $(15.5, 31.2)$ falls close to the \emph{w/o IM} trajectory, AnimateDiff~\cite{guo2024animatediff} at $(21.5, 36.7)$ attains a worse trade-off than ours at $(20.4, 40.5)$, and a dual-ControlNet inpaint+pose variant at $(8.5, 19.7)$ lands near our SD Baseline; all three are
described in Secs.~6.5--6.7 of the supplementary. Notably, higher visual fidelity does not imply higher utility: on a single-person UCF101 subset, WAN2.2-Animate~\cite{cheng2025wananimate}, driven by a FLUX.2~\cite{flux2_2025} reference identity and thus not sharing our SD backbone, produces sharper frames yet obtains a worse person-region FID (57.77 vs.\ 31.05) and lower accuracy (32.43\% vs.\ 35.52\%) than StablePrivacy, indicating that the trade-off is governed by the anonymization design rather than the generator (full comparison, including DP2 and the SD Baseline, in Supplementary Secs.~10.2 and 10.3). Additionally, \cref{tab:ablation-comp} shows the impact of ablating injection module (IM), latent inpainting (Lat. Inpnt), ControlNet (CntrlNet) and the memory bank (MB) of StablePrivacy. Although removing latent inpainting slightly improves performance on MSMT17, it introduces noticeable boundary artifacts (see \cref{fig:qualitative_comparison} and Figs.~12–14, 20, and 21 in the supplementary). Removing the MB costs utility (40.5 to 34.6 mAP) and temporal consistency (ACD 0.574 to 0.672, \cref{fig:acd_metric}).

\section{Conclusion}
\label{sec:conclusion}
We present \textbf{StablePrivacy}, a zero-shot diffusion-based framework for full-body video anonymization that needs no dataset-specific tuning and achieves a stronger privacy--utility trade-off than existing methods, with the clearest gains on person Re-ID and competitive performance on action recognition and video instance segmentation (VIS). Our experiments indicate that temporal consistency has limited impact on downstream performance for action recognition and VIS, but is important for person Re-ID, where our method achieves a substantial improvement over existing baselines. We also find that higher visual fidelity does not by itself imply better utility (Supplementary Secs.~10.2 and 10.3).

\noindent \textbf{Limitations / Future Work:} Our framework offers no explicit control over the generated identity or clothing: what is preserved and what is replaced is governed by the relative noise levels ($\sigma_{g}, \sigma_{i}$) of the two branches. Improving the privacy--utility trade-off remains an open challenge. Further, current privacy evaluations rely on re-ID metrics that emphasize clothing and texture cues. As our method preserves the input pose, gait is a further channel such metrics miss: StablePrivacy halves rank-1 gait retrieval versus the original data and leaks less than DP2, but does not remove it (Supplementary Sec.~10.1). This highlights the need for stronger privacy metrics and standardized benchmarks for full-body video anonymization.

{
    \small
    \bibliographystyle{ieeenat_fullname}
    \bibliography{main}
}
\clearpage
\setcounter{page}{1}
\renewcommand{\thepage}{S\arabic{page}}   
\maketitlesupplementary

\setcounter{figure}{8}   
\setcounter{table}{3}    
\setcounter{section}{5}  

\section{Baseline Methods}
\subsection{DP2}
For the DP2 baseline, we are using the code from their open-source Github release and the config file given by \texttt{configs/anonymizers/FB\_cse.py}.

\subsection{SD Baseline}
As described in the main paper, for the `SD baseline', we are using naive pixel inpainting implementation along with ControlNet model for pose conditioning. We also first encode each incoming video frame, $V_{i}$, using Stable Diffusion's VAE, and then start the noising and denoising process from this encoded frame's latent, similar to the process used in our proposed method. Since we start generation from the encoded frame's latent, it gives us some basic level of spatio-temporal consistency, which can be observed from the various qualitative examples of `SD Baseline'. We use the same prompt given by: ``A high-quality clear full body photo of a person. Detailed eyes, face, body, clothes, hands, and legs.'', across all the Stable Diffusion based baselines like StreamV2V, SD Baseline, and our proposed method. We also start the generation with the same initial noise levels across all the diffusion-based baselines.

\subsection{Stable Animator}
We are using the official open-source code of Stable Animator available on Github with the following hyperparameters: \texttt{num\_optimization\_iter=1}, \linebreak \texttt{start\_refine\_step=5}, \texttt{end\_refine\_step=15}, \texttt{max\_frame\_num=14}, \linebreak \texttt{batch\_frames=14}. We also generate videos with $512 \times 512$ resolution due to GPU VRAM limitations. Furthermore, if the number of video frames is less than \texttt{tile\_size} parameter, we set \texttt{tile\_overlap=0}, and \texttt{tile\_size=num\_frames}. Further, as mentioned in the main paper, we use the SD Baseline to anonymize the first frame of each input video and provide this anonymized frame as the reference image for Stable Animator. We also use the same pose sequence across our StreamV2V, SD Baseline, Stable Animator, and our proposed method. The pose image $P_{i}$ is extracted from frame $V_{i}$ using DWPose~\cite{yang2023effective}, where the detector is constrained to the regions defined by input person masks $M_{i}$. We convert the segmentation masks to bounding boxes and pass these as region proposals to DWPose, ensuring spatial alignment and agreement between the masked regions and detected poses. Thus, this alignment guarantees that the generated content is spatially consistent with both the pose guidance and the input masks.

\subsection{StreamV2V}
We use the same mask, pose, and prompt sequence for the StreamV2V baseline as used in existing baselines and our method. We also keep the generation noise levels the same across the baselines, i.e., set $\sigma=0.9$. Furthermore, we preserve the input video's aspect ratio and resize it such that the shorter side is set to \(512\) when generating videos using StreamV2V or our method. Since stable diffusion models are trained with $512 \times 512$ resolution, we observe that preserving the input aspect ratio and using shorter side set as $512$, helps improve the quality of generation.

\subsection{RAD}
RAD~\cite{malm2023rad} is a per-frame diffusion anonymizer built on an SDXL-class generator
(RealVisXL~V3.0). It segments people with YOLOv8x and SAM, runs full-image img2img at strength $0.8$ under two
ControlNets (Canny on the person silhouette and OpenPose), and stitches the generated foreground back onto the
original background. We evaluate it under our exact MSMT17 protocol: the training split is anonymized to
measure utility and the query split to measure privacy.

\subsection{AnimateDiff}
This baseline tests whether an off-the-shelf temporal module can substitute for our memory bank and injection
module. It is our SD Baseline with a single change: the pretrained AnimateDiff motion
module~\cite{guo2024animatediff}, which inserts temporal attention after each spatial block of the same
SD~1.5 UNet. Backbone, pose ControlNet, VAE, prompt, guidance scale, seed, starting noise level
($\sigma=0.9$), resolution rule and pixel compositing are all held identical to the SD Baseline, so any
difference is attributable to the temporal module alone. Clips longer than the module's native 16-frame
context use FreeNoise (context length 16, stride 4) rather than hard chunking, so the baseline is not
penalised by seams.

\subsection{Dual-ControlNet (Inpaint + Pose)}
This is a stronger per-frame image baseline than our SD Baseline. Every frame is generated from pure Gaussian
noise with SD~1.5 under two simultaneous ControlNets: an inpaint ControlNet conditioned on the frame with the
person pixels masked out, and the same OpenPose ControlNet used elsewhere, both at conditioning scale $1.0$.
It uses the UniPC sampler (20 steps, guidance $7.5$) and the shared prompt. Relative to our method it replaces
latent-space inpainting with the inpaint ControlNet and dual-noise initialisation with a pure-noise start, and
it contains neither the injection module nor the memory bank; it therefore tests whether an off-the-shelf
composition of ControlNets can stand in for those components.

\section{Task Details}
\subsection{Video Instance Segmentation (VIS)}
VIS is the task of image instance segmentation extended to video domain. It tackles simultaneous detection, segmentation, and tracking of instances in a given video. Unlike Video Object Segmentation (VOS), VIS also requires the recognition of object categories in the video. Thus, an instance label needs to be assigned to each proposal. The IoU metric in this task takes into account the instance labels and also the temporal dimension of the videos. We use the MaskTrackRCNN model from the MMDetection~\cite{mmdetection} open-source toolbox for this task. Further, for the binary human instance segmentation masks used in this task, we combine the GT masks of VIS19 dataset with the masks obtained from DP2 using the CSE model, and use the same combined masks across all baseline methods. For other tasks and datasets, we directly use the masks obtained from DP2 model across all our baselines to keep the masks uniform.

\subsection{Video Action Recognition}
Video action recognition involves predicting the action category from a pre-defined list of classes for a given input video. For the UCF-101 dataset, this involves predicting the class label from 101 categories. We use the C3D model from the MMAction2~\cite{2020mmaction2} open-source toolbox for this task. Please follow the pre-processing guide mentioned in MMAction2 before training the model on this dataset.

\subsection{Person Re-ID}
\label{sec:suppl-reid}
This task involves matching a `probe' data point against a large `gallery' set. The goal is to train a model that learns an embedding space where feature vectors of the same identity are pulled closer together, while those belonging to different identities are pushed apart. Typically, training and testing identities are disjoint. Once this embedding function is learned, the model is evaluated on query and gallery sets consisting of feature vectors extracted by the trained model. For each query, nearest neighbors are retrieved from the gallery set within the embedding space. We then employ the mean Average Precision (mAP) metric to evaluate the quality of the retrieved ranked lists. The MSMT17 dataset originates from raw videos; however, the released format consists of cropped bounding box images collected across three time slots: morning, noon, and afternoon. For our evaluation on MSMT17, we concatenate these disjoint static crops into a single video sequence for each identity. Consequently, these constructed sequences may exhibit abrupt changes in pose and background, in contrast to smooth, natural video. Thus, to complement this, we also evaluate the baselines and our proposed approach on the MARS dataset, which comprises spatio-temporally smooth video tracklets that more closely resemble real-world natural video.

\section{Privacy: Threat Model and Evaluation Protocol}
\label{sec:suppl-privacy}

\subsection{Threat Model}
\label{sec:suppl-threatmodel}
We state explicitly what our privacy claim covers, since the term ``anonymization'' is used with different meanings across the literature.

\noindent \textbf{What we mean by privacy.} We target \emph{unlinkability}: given a released anonymized video, an adversary should not be able to match it back to the individual from whom it was derived. This differs from the attribute-suppression view adopted in privacy-preserving action recognition, where privacy is measured by whether sensitive attributes remain detectable. Our method replaces private visual content with synthetic content rather than removing information, so attributes such as clothing type may still be detectable in the output; our claim is that they no longer correspond to the original person (\cref{sec:suppl-attrprotocol}).

\noindent \textbf{Adversary capabilities.} We assume an adversary who obtains the released anonymized data and attempts to re-identify the individuals in it. The adversary holds the original, non-anonymized gallery and a state-of-the-art recognition model pre-trained on external data: TransReID~\cite{he2021transreid} for appearance-based retrieval, and GaitBase~\cite{fan2023opengait} with the GREW~\cite{zhu2021gait} and Gait3D~\cite{zheng2022gait3d} checkpoints for gait-based retrieval. Retrieval accuracy against the original identities is our privacy measure, and lower accuracy indicates stronger privacy.

\noindent \textbf{Assumptions.} The adversary uses off-the-shelf recognition models and does not adapt them to our anonymization; an adaptive adversary that fine-tunes on paired original and anonymized data is outside our scope. We further assume that no standardized threshold exists below which retrieval accuracy can be declared safe. For this reason the main paper reports privacy--utility trade-off curves rather than a single operating point, and we report reductions relative to the original data rather than an absolute guarantee.

\noindent \textbf{Residual channels and failure cases.} Two channels remain open. First, the injection module preserves coarse appearance such as clothing type and colour. These cues identify an individual only when that individual is uniquely distinguished by them within the population, for example the only person wearing a particular colour in a dataset. In the benchmarks we use, which contain many visually similar pedestrians, such cues are typically shared across many identities, so this is a per-individual failure mode rather than an average-case one. Second, because ControlNet conditioning preserves the input pose sequence, gait is retained to a degree that appearance-based Re-ID metrics do not capture; we measure this directly in \cref{sec:suppl-gait}.

\noindent \textbf{Out of scope.} We do not evaluate face verification attacks. Our privacy evaluation is conducted on the person Re-ID benchmarks MSMT17~\cite{wei2018person} and MARS~\cite{zheng2016mars}, whose pedestrian crops place the face region at a few tens of pixels, far below the resolution at which face verification models operate reliably, so such an attack is not meaningful on this data. We also do not evaluate attribute-inference protocols, for the reasons given in \cref{sec:suppl-attrprotocol}. Neither exclusion should be read as a claim that our method resists these attacks, and we do not present StablePrivacy as providing a guarantee of regulatory compliance.

\subsection{Why Attribute-Detectability Protocols Do Not Apply}
\label{sec:suppl-attrprotocol}

Privacy-Preserving Action Recognition (PPAR) methods~\cite{li2023stprivacy, dave2022spact, kumawat2022privacy, hinojosa2022privhar, wu2018towards, ilic2024selective} can be classified under the learning-based transformation paradigm. Specifically, these methods learn an anonymization function through an adversarial training objective that jointly optimizes an action recognition (utility) objective and a privacy objective. As a result, the generated outputs are task-specific, visually unrealistic, and no longer resemble natural images or videos. In contrast, our anonymized outputs maintain visual realism and preserve the overall context of the input scene. Therefore, the privacy evaluation protocol commonly used in PPAR is not meaningful or applicable in our setting.

Existing protocols such as VP-UCF101~\cite{li2023stprivacy} annotate the UCF101 dataset with five human attributes: face, skin color, gender, nudity, and familiar relationship. These annotations are provided as binary labels indicating whether each attribute is present in a given video. Privacy recognition is formulated as a multi-label binary classification problem, with performance evaluated using cMAP and F1 scores. To assess privacy, a video privacy recognizer is trained on the anonymized training split of VP-UCF101 and evaluated on the corresponding anonymized test split. Therefore, lower recognition performance indicates better privacy with respect to the annotated sensitive attributes.

However, our objective fundamentally differs from the privacy-preservation settings in PPAR. Existing protocols evaluate privacy based on whether sensitive attributes (e.g., gender or race) remain detectable after anonymization, implicitly assuming that privacy is achieved by suppressing such information through masking, obfuscation or distortion. In contrast, our method protects privacy by replacing private visual content with synthetic yet realistic content. While sensitive attributes may still be detectable, they no longer correspond to the original individual.
Therefore, attribute-detectability–based evaluation protocols are not meaningful for assessing privacy in our setting.

\section{Miscellaneous Details}
\begin{figure}[t]
  \centering
  \includegraphics[width=\linewidth]{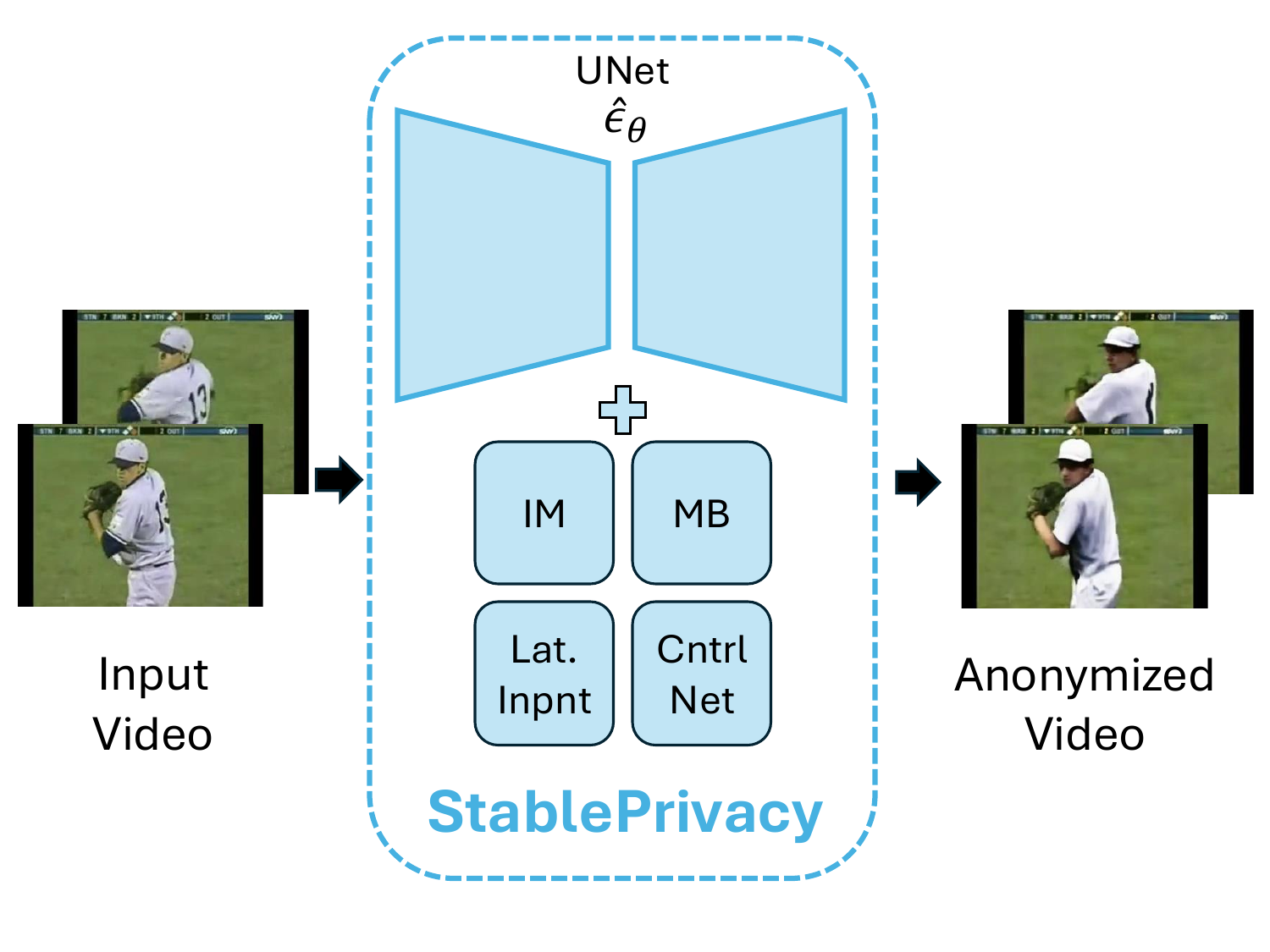}
  \caption{High-level overview of our approach, combining an Injection Module (IM), Memory Bank (MB), Latent Inpainting, and ControlNet conditioning within a pretrained UNet denoiser.}
  \label{fig:method_overview}
\end{figure}

A high-level overview of our anonymization pipeline is shown in~\cref{fig:method_overview}. Our method follows StreamV2V~\cite{liang2024looking} for all other specific hyperparameter / implementation choices. Noise schedules are based on the latent consistency model (LCM) scheduler. Injection is done only in the first 3 denoising steps out of the four. $x_{t_{inj}}$ passes through all encoder blocks up to \texttt{up1} decoder block ($16 \times 16$ res) of UNet where it is injected with $x_t$.

\noindent \textbf{Memory bank.} The memory bank replaces the attention processor in every self-attention layer of the UNet; the caching path is active only for self-attention and leaves cross-attention unchanged. A separate cache is maintained for each of the four LCM denoising steps, so that the features of frame $i$ at step $t$ attend to the features of frame $i-1$ at the same step. The cache stores the keys, values, and attention outputs of the previous frame only, and is updated at every frame. When generating frame $i$, the cached keys and values are concatenated with those of the current frame along the token dimension, so that the current queries attend over the union of both frames, giving an extended self-attention. To bound the memory and compute overhead of this concatenation, cached tokens are compacted by approximately half using random bipartite soft matching (token merging) before being stored. In addition, in the \texttt{mid} block and the first two decoder blocks of the UNet, the attention output of the current frame is blended with its nearest-neighbour features among the cached outputs, using a blending strength of $0.8$ and a cosine similarity threshold of $0.98$.

\begin{figure}[t]
  \centering
  \includegraphics[width=\linewidth]{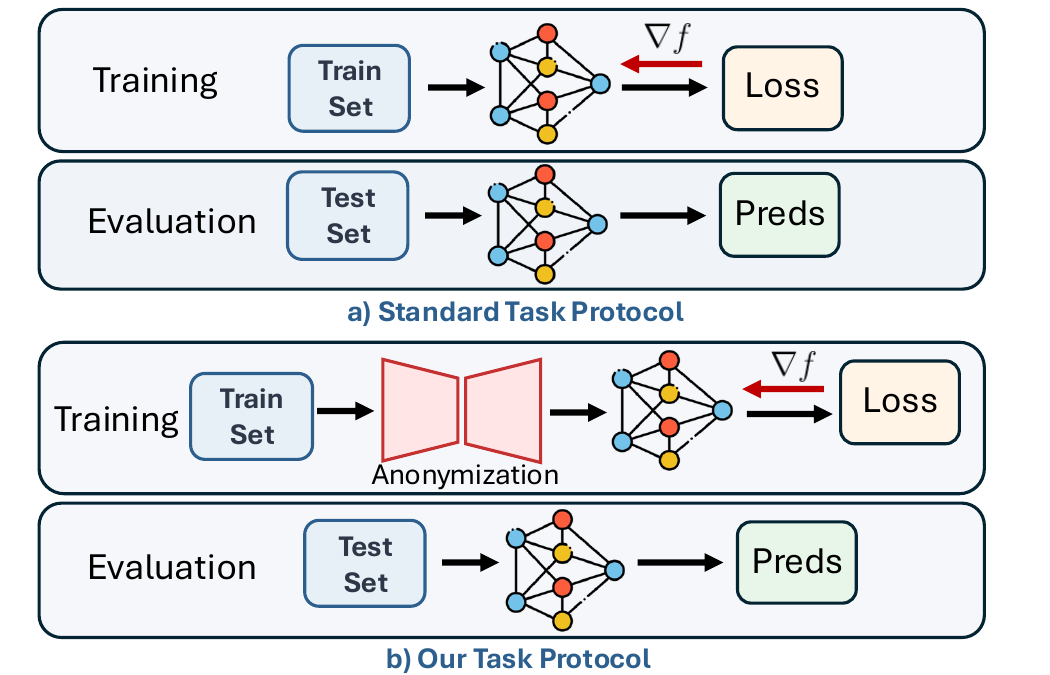}
  \caption{\textbf{Task protocol.} (a) The standard protocol trains and evaluates a downstream model on original data. (b) Ours anonymizes only the training set: the model is trained on anonymized videos and evaluated on the \emph{original}, non-anonymized test set, so utility measures how well anonymized data substitutes for real training data.}
  \label{fig:task_protocol}
\end{figure}

\noindent \textbf{Task protocol.} \cref{fig:task_protocol} contrasts our evaluation protocol with the standard one. We anonymize only the \emph{training} split of each dataset, train the downstream model on it, and evaluate that model on the original, unmodified test split. Utility therefore measures how well anonymized videos can substitute for real data when a training set is released, rather than how well a model copes with anonymized inputs at test time. The one exception is our privacy measurement, where the Re-ID model is applied to anonymized queries in order to test re-identification (\cref{sec:suppl-threatmodel}).

We will release our source code publicly to ensure reproducibility and clarity of our implementation details.

\section{Additional Experiments and Analysis}
\label{sec:suppl-additional}

\subsection{Gait Leakage}
\label{sec:suppl-gait}
Because our method preserves the input pose sequence through ControlNet conditioning, gait is a threat model that person Re-ID metrics do not capture, as Re-ID relies largely on clothing, colour, and texture cues. We evaluate it with GaitBase~\cite{fan2023opengait}, a state-of-the-art silhouette-based gait recogniser, using checkpoints pre-trained on GREW~\cite{zhu2021gait} and Gait3D~\cite{zheng2022gait3d}, and perform nearest-neighbour retrieval over the full MARS test set.

\cref{tab:suppl-gait} shows that anonymisation roughly halves rank-1 gait retrieval accuracy relative to the original data. Across both pre-trained checkpoints, StablePrivacy leaks less gait information than DP2, which is consistent with our Re-ID results. Gait nonetheless remains a residual leakage channel for any pose-preserving anonymisation method. Reducing it further while retaining the pose fidelity that downstream tasks depend on is an important direction for future work.

\begin{table}[t]
  \centering
  \footnotesize
  \begin{tabular}{lcc}
    \toprule
    Probe & GREW~$\downarrow$ & Gait3D~$\downarrow$ \\
    \midrule
    Original & 24.7 & 22.4 \\
    DP2~\cite{hukkelaas2023deepprivacy2} & 15.0 & 13.6 \\
    \rowcolor{gray!20}\textbf{Ours} & \textbf{13.0} & \textbf{12.8} \\
    \bottomrule
  \end{tabular}
  \caption{\textbf{Gait leakage on MARS.} Zero-shot GaitBase~\cite{fan2023opengait} rank-1 retrieval accuracy (\%, lower indicates stronger privacy) using two pre-trained checkpoints, GREW~\cite{zhu2021gait} and Gait3D~\cite{zheng2022gait3d}. Anonymisation roughly halves rank-1 accuracy, and StablePrivacy leaks less than DP2 on both checkpoints.}
  \label{tab:suppl-gait}
\end{table}

\begin{figure*}[t]
  \centering
  \includegraphics[width=\linewidth]{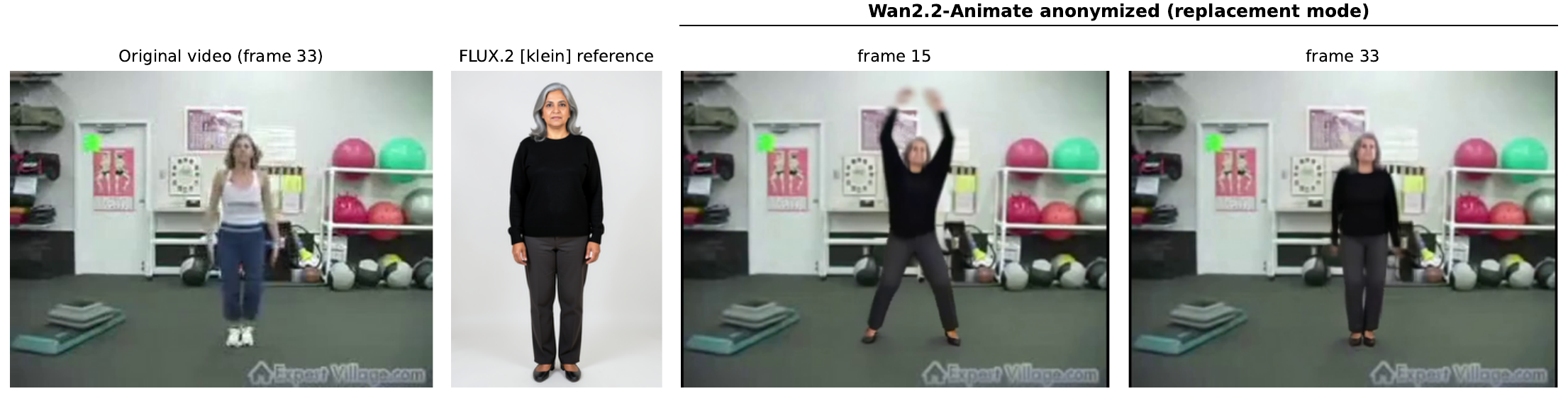}
  \caption{\textbf{High-fidelity generation does not imply better anonymisation utility.} A UCF101 video anonymised with WAN2.2-Animate~\cite{cheng2025wananimate} in replacement mode, driven by a reference identity synthesised with FLUX.2~[klein]~\cite{flux2_2025}. Left to right: an original frame, the FLUX.2 reference identity, and two anonymised output frames. The replacement is sharp and temporally stable, but it substitutes an externally generated person whose appearance is unrelated to the original data distribution; this is reflected in the worst FID in~\cref{tab:suppl-fidelity} and in lower downstream utility than StablePrivacy.}
  \label{fig:wan_qualitative}
\end{figure*}

\subsection{Visual Fidelity Does Not Imply Downstream Utility}
\label{sec:suppl-fidelity}
Our objective is to maximise the privacy--utility trade-off rather than to produce photorealistic full-body replacements, and these two goals are not equivalent. To test this directly, we compare against WAN2.2-Animate~\cite{cheng2025wananimate}, a state-of-the-art high-fidelity video generator, operated in its character-replacement mode. Because WAN2.2-Animate is designed for single-person videos, we evaluate on a 200-video subset of UCF101 restricted to 43 single-person classes. We report FID computed over the person region against the original frames, together with downstream top-1 action recognition accuracy.

As shown in~\cref{tab:suppl-fidelity}, WAN2.2-Animate produces visually sharp frames yet obtains the \emph{worst} FID among all compared methods, because it inserts an externally generated synthetic identity instead of re-synthesising a person consistent with the original data distribution. Its downstream utility is also lower than that of StablePrivacy. Perceptual realism of individual frames is therefore not a reliable proxy for how useful the anonymised data is as training material, which is the property that matters in our setting.

\begin{table}[t]
  \centering
  \footnotesize
  \begin{tabular}{lcc}
    \toprule
    Method & FID~$\downarrow$ & Top-1 Acc.~(\%)~$\uparrow$ \\
    \midrule
    DP2~\cite{hukkelaas2023deepprivacy2} & 33.26 & 30.12 \\
    WAN2.2-Animate~\cite{cheng2025wananimate} & 57.77 & 32.43 \\
    SD Baseline & 48.57 & 34.75 \\
    \rowcolor{gray!20}\textbf{StablePrivacy (Ours)} & \textbf{31.05} & \textbf{35.52} \\
    \midrule
    Original & --- & 37.84 \\
    \bottomrule
  \end{tabular}
  \caption{\textbf{Visual fidelity does not imply downstream utility.} Evaluated on 200 UCF101 videos from 43 single-person classes, as WAN2.2-Animate is designed for single-person video. FID is computed over the person region against the original frames; accuracy is downstream top-1 action recognition. WAN2.2-Animate produces sharp frames but the worst FID, since it inserts an external synthetic identity.}
  \label{tab:suppl-fidelity}
\end{table}

\subsection{Strength of the Underlying Generator}
\label{sec:suppl-generator}
A natural question is whether the observed trade-off is limited by our Stable Diffusion backbone. To test this, the WAN2.2-Animate baseline above is driven by a reference identity synthesised with FLUX.2~[klein]~\cite{flux2_2025}, a recent state-of-the-art image generator; \cref{fig:wan_qualitative} shows an example. This baseline therefore does not share our SD-based generation pipeline and, unlike Stable Animator, is not constrained by a reference frame that we anonymised with the SD Baseline. Even with this stronger generator stack, both FID and downstream utility remain below StablePrivacy (\cref{tab:suppl-fidelity}). We therefore attribute the privacy--utility trade-off primarily to the anonymisation design rather than to the capacity of the underlying generator. The components responsible are the dual-noise initialisation, the injection module, and latent inpainting.


\section{Qualitative Results}
\subsection{MSMT17 Dataset}
\begin{figure*}[t]
    \centering
    \setlength{\tabcolsep}{0.5pt}
    \renewcommand{\arraystretch}{0.5}

    \begin{minipage}[b]{0.32\textwidth}
        \centering
        \resizebox{\linewidth}{!}{%
\begin{tabular}{@{}l@{\hspace{2pt}}c@{\hspace{1pt}}c@{\hspace{1pt}}c@{\hspace{1pt}}c@{}}
            & \scriptsize F1 & \scriptsize F2 & \scriptsize F3 & \scriptsize F4 \\
            \cmidrule(lr){2-5}
            \raisebox{2ex}{\tiny Original} &
            \includegraphics[width=0.22\linewidth]{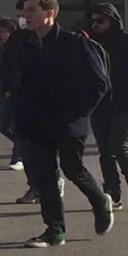} &
            \includegraphics[width=0.22\linewidth]{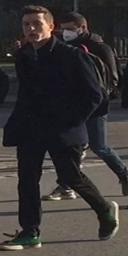} &
            \includegraphics[width=0.22\linewidth]{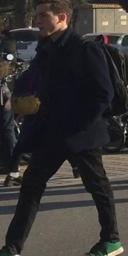} &
            \includegraphics[width=0.22\linewidth]{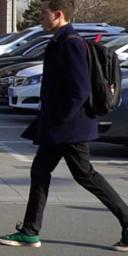} \\
            \raisebox{2ex}{\tiny DP2} &
            \includegraphics[width=0.22\linewidth]{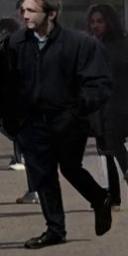} &
            \includegraphics[width=0.22\linewidth]{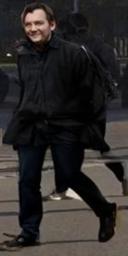} &
            \includegraphics[width=0.22\linewidth]{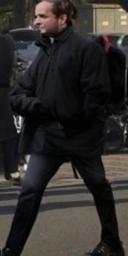} &
            \includegraphics[width=0.22\linewidth]{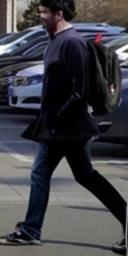} \\
            \raisebox{2ex}{\tiny SD Base} &
            \includegraphics[width=0.22\linewidth]{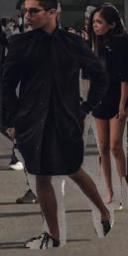} &
            \includegraphics[width=0.22\linewidth]{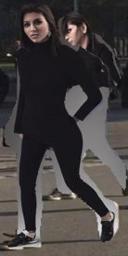} &
            \includegraphics[width=0.22\linewidth]{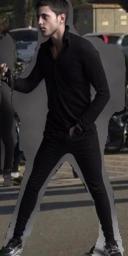} &
            \includegraphics[width=0.22\linewidth]{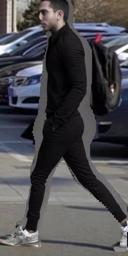} \\
\end{tabular}%
        }
        \vspace{1pt}
        \textbf{\scriptsize Identity 0029}
    \end{minipage}
    \hfill
    \begin{minipage}[b]{0.32\textwidth}
        \centering
        \resizebox{\linewidth}{!}{%
        \begin{tabular}{@{}l@{\hspace{2pt}}c@{\hspace{1pt}}c@{\hspace{1pt}}c@{\hspace{1pt}}c@{}}
            & \scriptsize F1 & \scriptsize F2 & \scriptsize F3 & \scriptsize F4 \\
            \cmidrule(lr){2-5}
            \raisebox{2ex}{\tiny Original} &
            \includegraphics[width=0.22\linewidth]{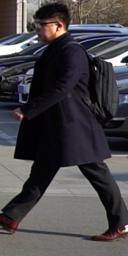} &
            \includegraphics[width=0.22\linewidth]{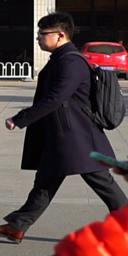} &
            \includegraphics[width=0.22\linewidth]{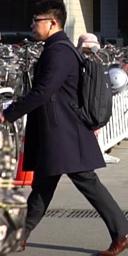} &
            \includegraphics[width=0.22\linewidth]{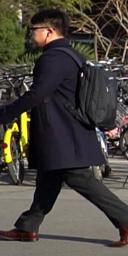} \\
            \raisebox{2ex}{\tiny DP2} &
            \includegraphics[width=0.22\linewidth]{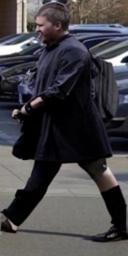} &
            \includegraphics[width=0.22\linewidth]{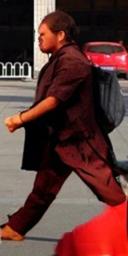} &
            \includegraphics[width=0.22\linewidth]{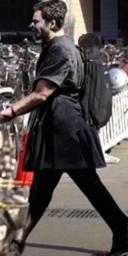} &
            \includegraphics[width=0.22\linewidth]{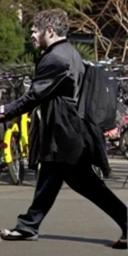} \\
            \raisebox{2ex}{\tiny SD Base} &
            \includegraphics[width=0.22\linewidth]{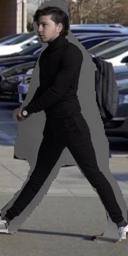} &
            \includegraphics[width=0.22\linewidth]{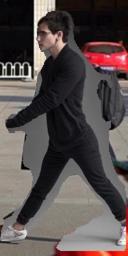} &
            \includegraphics[width=0.22\linewidth]{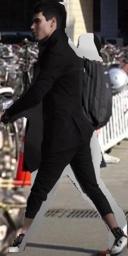} &
            \includegraphics[width=0.22\linewidth]{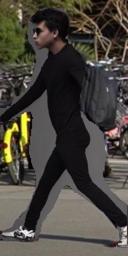} \\
\end{tabular}%
        }
        \vspace{1pt}
        \textbf{\scriptsize Identity 0036}
    \end{minipage}
    \hfill
    \begin{minipage}[b]{0.32\textwidth}
        \centering
        \resizebox{\linewidth}{!}{%
        \begin{tabular}{@{}l@{\hspace{2pt}}c@{\hspace{1pt}}c@{\hspace{1pt}}c@{\hspace{1pt}}c@{}}
            & \scriptsize F1 & \scriptsize F2 & \scriptsize F3 & \scriptsize F4 \\
            \cmidrule(lr){2-5}
            \raisebox{2ex}{\tiny Original} &
            \includegraphics[width=0.22\linewidth]{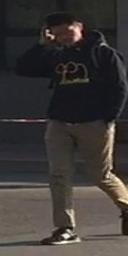} &
            \includegraphics[width=0.22\linewidth]{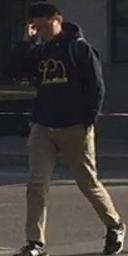} &
            \includegraphics[width=0.22\linewidth]{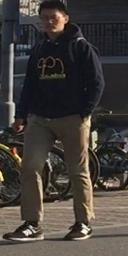} &
            \includegraphics[width=0.22\linewidth]{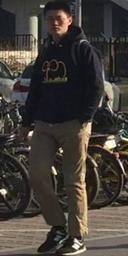} \\
            \raisebox{2ex}{\tiny DP2} &
            \includegraphics[width=0.22\linewidth]{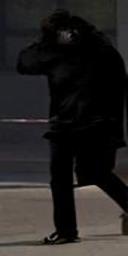} &
            \includegraphics[width=0.22\linewidth]{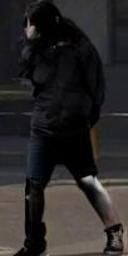} &
            \includegraphics[width=0.22\linewidth]{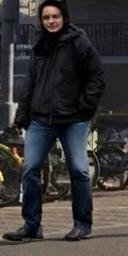} &
            \includegraphics[width=0.22\linewidth]{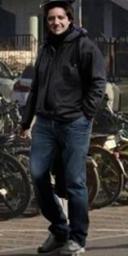} \\
            \raisebox{2ex}{\tiny SD Base} &
            \includegraphics[width=0.22\linewidth]{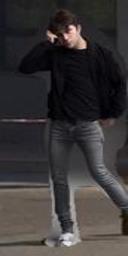} &
            \includegraphics[width=0.22\linewidth]{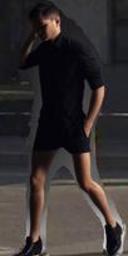} &
            \includegraphics[width=0.22\linewidth]{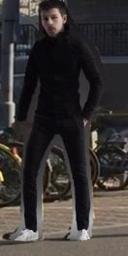} &
            \includegraphics[width=0.22\linewidth]{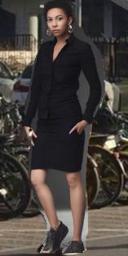} \\
\end{tabular}%
        }
        \vspace{1pt}
        \textbf{\scriptsize Identity 0038}
    \end{minipage}

    \caption{Qualitative comparison of anonymization on MSMT17 dataset. Our method maintains better temporal consistency, generates better quality humans, and preserves non-sensitive features for utility. Best viewed zoomed in.}
    \label{fig:qualitative_part_1_msmt}
\end{figure*}

\begin{figure*}[t]
    \centering
    \setlength{\tabcolsep}{0.5pt}
    \renewcommand{\arraystretch}{0.5}

    \begin{minipage}[b]{0.32\textwidth}
        \centering
        \resizebox{\linewidth}{!}{%
        \begin{tabular}{@{}l@{\hspace{2pt}}c@{\hspace{1pt}}c@{\hspace{1pt}}c@{\hspace{1pt}}c@{}}
            & \scriptsize F1 & \scriptsize F2 & \scriptsize F3 & \scriptsize F4 \\
            \cmidrule(lr){2-5}
            \raisebox{2ex}{\tiny Original} &
            \includegraphics[width=0.22\linewidth]{figures/qualitative/msmt17/train/0034/0034_000_01_0303morning_0499_1.jpg} &
            \includegraphics[width=0.22\linewidth]{figures/qualitative/msmt17/train/0034/0034_001_01_0303morning_0499_1_ex.jpg} &
            \includegraphics[width=0.22\linewidth]{figures/qualitative/msmt17/train/0034/0034_002_01_0303morning_0500_0.jpg} &
            \includegraphics[width=0.22\linewidth]{figures/qualitative/msmt17/train/0034/0034_003_01_0303morning_0500_0_ex.jpg} \\
            \raisebox{2ex}{\tiny DP2} &
            \includegraphics[width=0.22\linewidth]{figures/qualitative/msmt17/dp2/0034/0034_000_01_0303morning_0499_1.jpg} &
            \includegraphics[width=0.22\linewidth]{figures/qualitative/msmt17/dp2/0034/0034_001_01_0303morning_0499_1_ex.jpg} &
            \includegraphics[width=0.22\linewidth]{figures/qualitative/msmt17/dp2/0034/0034_002_01_0303morning_0500_0.jpg} &
            \includegraphics[width=0.22\linewidth]{figures/qualitative/msmt17/dp2/0034/0034_003_01_0303morning_0500_0_ex.jpg} \\
            \raisebox{2ex}{\tiny SD Base} &
            \includegraphics[width=0.22\linewidth]{figures/qualitative/msmt17/SD_baseline/0034/0034_000_01_0303morning_0499_1.jpg} &
            \includegraphics[width=0.22\linewidth]{figures/qualitative/msmt17/SD_baseline/0034/0034_001_01_0303morning_0499_1_ex.jpg} &
            \includegraphics[width=0.22\linewidth]{figures/qualitative/msmt17/SD_baseline/0034/0034_002_01_0303morning_0500_0.jpg} &
            \includegraphics[width=0.22\linewidth]{figures/qualitative/msmt17/SD_baseline/0034/0034_003_01_0303morning_0500_0_ex.jpg} \\
\end{tabular}%
        }
        \vspace{1pt}
        \textbf{\scriptsize Identity 0034}
    \end{minipage}
    \hfill
    \begin{minipage}[b]{0.32\textwidth}
        \centering
        \resizebox{\linewidth}{!}{%
        \begin{tabular}{@{}l@{\hspace{2pt}}c@{\hspace{1pt}}c@{\hspace{1pt}}c@{\hspace{1pt}}c@{}}
            & \scriptsize F1 & \scriptsize F2 & \scriptsize F3 & \scriptsize F4 \\
            \cmidrule(lr){2-5}
            \raisebox{2ex}{\tiny Original} &
            \includegraphics[width=0.22\linewidth]{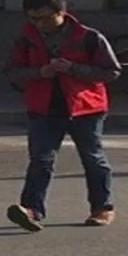} &
            \includegraphics[width=0.22\linewidth]{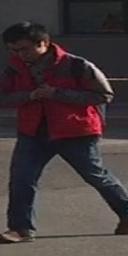} &
            \includegraphics[width=0.22\linewidth]{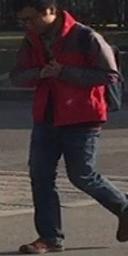} &
            \includegraphics[width=0.22\linewidth]{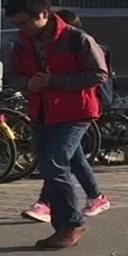} \\
            \raisebox{2ex}{\tiny DP2} &
            \includegraphics[width=0.22\linewidth]{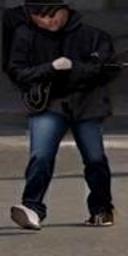} &
            \includegraphics[width=0.22\linewidth]{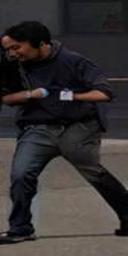} &
            \includegraphics[width=0.22\linewidth]{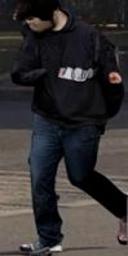} &
            \includegraphics[width=0.22\linewidth]{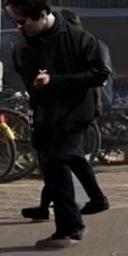} \\
            \raisebox{2ex}{\tiny SD Base} &
            \includegraphics[width=0.22\linewidth]{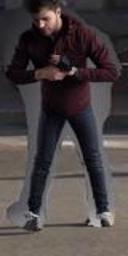} &
            \includegraphics[width=0.22\linewidth]{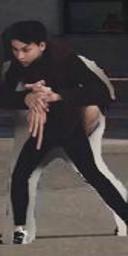} &
            \includegraphics[width=0.22\linewidth]{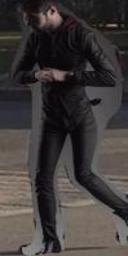} &
            \includegraphics[width=0.22\linewidth]{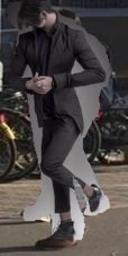} \\
\end{tabular}%
        }
        \vspace{1pt}
        \textbf{\scriptsize Identity 0037}
    \end{minipage}
    \hfill
    \begin{minipage}[b]{0.32\textwidth}
        \centering
        \resizebox{\linewidth}{!}{%
        \begin{tabular}{@{}l@{\hspace{2pt}}c@{\hspace{1pt}}c@{\hspace{1pt}}c@{\hspace{1pt}}c@{}}
            & \scriptsize F1 & \scriptsize F2 & \scriptsize F3 & \scriptsize F4 \\
            \cmidrule(lr){2-5}
            \raisebox{2ex}{\tiny Original} &
            \includegraphics[width=0.22\linewidth]{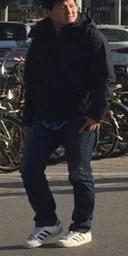} &
            \includegraphics[width=0.22\linewidth]{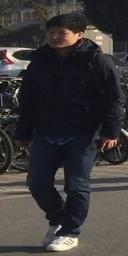} &
            \includegraphics[width=0.22\linewidth]{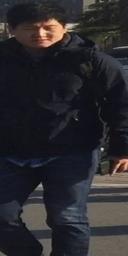} &
            \includegraphics[width=0.22\linewidth]{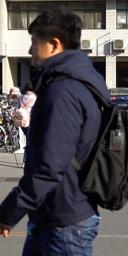} \\
            \raisebox{2ex}{\tiny DP2} &
            \includegraphics[width=0.22\linewidth]{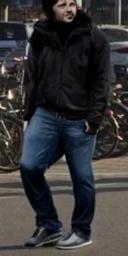} &
            \includegraphics[width=0.22\linewidth]{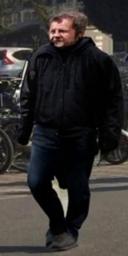} &
            \includegraphics[width=0.22\linewidth]{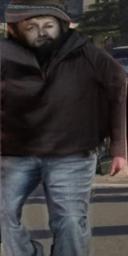} &
            \includegraphics[width=0.22\linewidth]{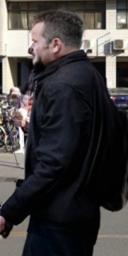} \\
            \raisebox{2ex}{\tiny SD Base} &
            \includegraphics[width=0.22\linewidth]{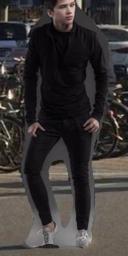} &
            \includegraphics[width=0.22\linewidth]{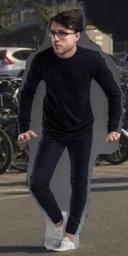} &
            \includegraphics[width=0.22\linewidth]{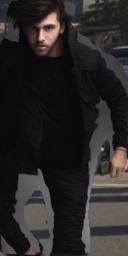} &
            \includegraphics[width=0.22\linewidth]{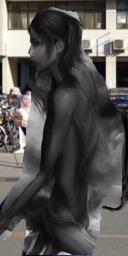} \\
\end{tabular}%
        }
        \vspace{1pt}
        \textbf{\scriptsize Identity 0049}
    \end{minipage}

    \caption{Qualitative comparison of anonymization on MSMT17 dataset (continued). Best viewed zoomed in.}
    \label{fig:qualitative_part_2_msmt}
\end{figure*}

\begin{figure*}[t]
    \centering
    \setlength{\tabcolsep}{0.5pt}
    \renewcommand{\arraystretch}{0.5}

    \begin{minipage}[b]{0.32\textwidth}
        \centering
        \resizebox{\linewidth}{!}{%
        \begin{tabular}{@{}l@{\hspace{2pt}}c@{\hspace{1pt}}c@{\hspace{1pt}}c@{\hspace{1pt}}c@{}}
            & \scriptsize F1 & \scriptsize F2 & \scriptsize F3 & \scriptsize F4 \\
            \cmidrule(lr){2-5}
            \raisebox{2ex}{\tiny Original} &
            \includegraphics[width=0.22\linewidth]{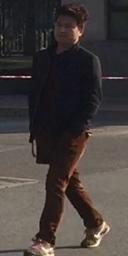} &
            \includegraphics[width=0.22\linewidth]{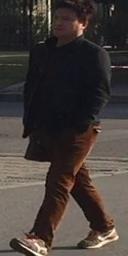} &
            \includegraphics[width=0.22\linewidth]{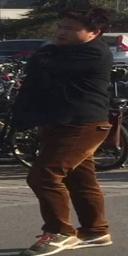} &
            \includegraphics[width=0.22\linewidth]{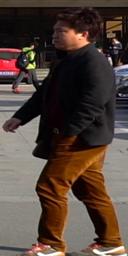} \\
            \raisebox{2ex}{\tiny DP2} &
            \includegraphics[width=0.22\linewidth]{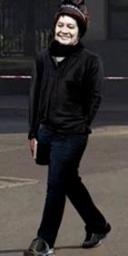} &
            \includegraphics[width=0.22\linewidth]{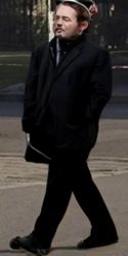} &
            \includegraphics[width=0.22\linewidth]{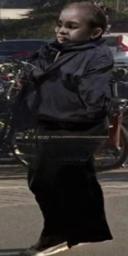} &
            \includegraphics[width=0.22\linewidth]{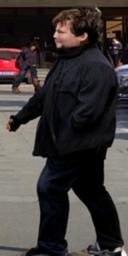} \\
            \raisebox{2ex}{\tiny SD Base} &
            \includegraphics[width=0.22\linewidth]{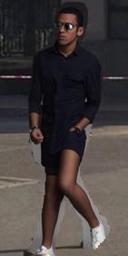} &
            \includegraphics[width=0.22\linewidth]{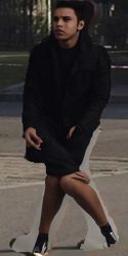} &
            \includegraphics[width=0.22\linewidth]{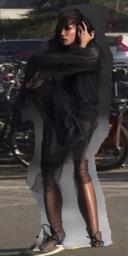} &
            \includegraphics[width=0.22\linewidth]{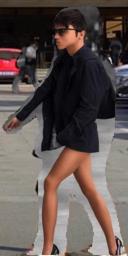} \\
\end{tabular}%
        }
        \vspace{1pt}
        \textbf{\scriptsize Identity 0058}
    \end{minipage}
    \hfill
    \begin{minipage}[b]{0.32\textwidth}
        \centering
        \resizebox{\linewidth}{!}{%
        \begin{tabular}{@{}l@{\hspace{2pt}}c@{\hspace{1pt}}c@{\hspace{1pt}}c@{\hspace{1pt}}c@{}}
            & \scriptsize F1 & \scriptsize F2 & \scriptsize F3 & \scriptsize F4 \\
            \cmidrule(lr){2-5}
            \raisebox{2ex}{\tiny Original} &
            \includegraphics[width=0.22\linewidth]{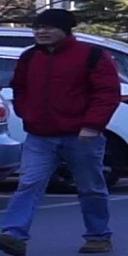} &
            \includegraphics[width=0.22\linewidth]{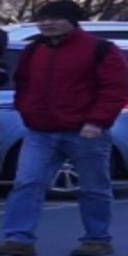} &
            \includegraphics[width=0.22\linewidth]{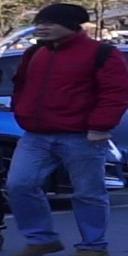} &
            \includegraphics[width=0.22\linewidth]{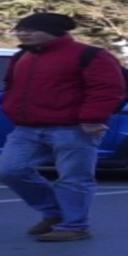} \\
            \raisebox{2ex}{\tiny DP2} &
            \includegraphics[width=0.22\linewidth]{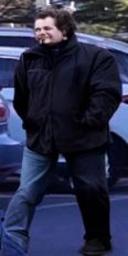} &
            \includegraphics[width=0.22\linewidth]{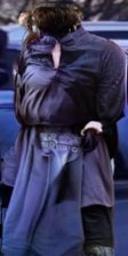} &
            \includegraphics[width=0.22\linewidth]{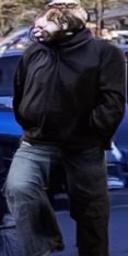} &
            \includegraphics[width=0.22\linewidth]{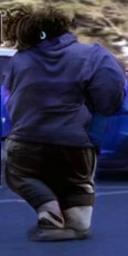} \\
            \raisebox{2ex}{\tiny SD Base} &
            \includegraphics[width=0.22\linewidth]{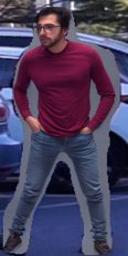} &
            \includegraphics[width=0.22\linewidth]{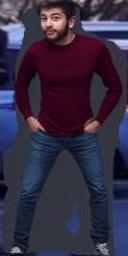} &
            \includegraphics[width=0.22\linewidth]{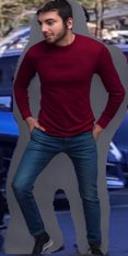} &
            \includegraphics[width=0.22\linewidth]{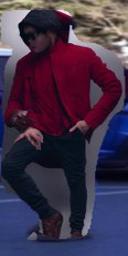} \\
\end{tabular}%
        }
        \vspace{1pt}
        \textbf{\scriptsize Identity 0452}
    \end{minipage}
    \hfill
    \begin{minipage}[b]{0.32\textwidth}
        \centering
        \resizebox{\linewidth}{!}{%
        \begin{tabular}{@{}l@{\hspace{2pt}}c@{\hspace{1pt}}c@{\hspace{1pt}}c@{\hspace{1pt}}c@{}}
            & \scriptsize F1 & \scriptsize F2 & \scriptsize F3 & \scriptsize F4 \\
            \cmidrule(lr){2-5}
            \raisebox{2ex}{\tiny Original} &
            \includegraphics[width=0.22\linewidth]{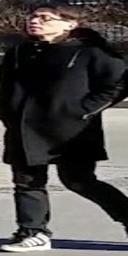} &
            \includegraphics[width=0.22\linewidth]{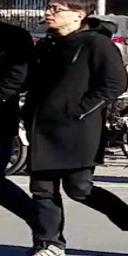} &
            \includegraphics[width=0.22\linewidth]{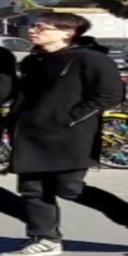} &
            \includegraphics[width=0.22\linewidth]{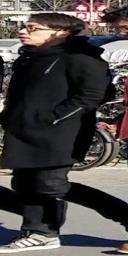} \\
            \raisebox{2ex}{\tiny DP2} &
            \includegraphics[width=0.22\linewidth]{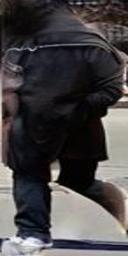} &
            \includegraphics[width=0.22\linewidth]{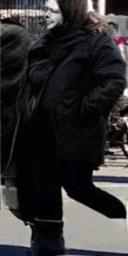} &
            \includegraphics[width=0.22\linewidth]{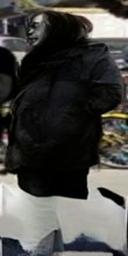} &
            \includegraphics[width=0.22\linewidth]{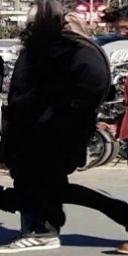} \\
            \raisebox{2ex}{\tiny SD Base} &
            \includegraphics[width=0.22\linewidth]{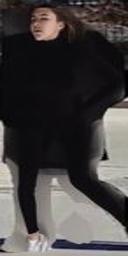} &
            \includegraphics[width=0.22\linewidth]{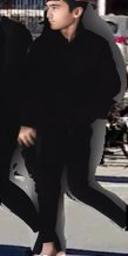} &
            \includegraphics[width=0.22\linewidth]{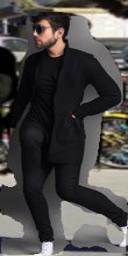} &
            \includegraphics[width=0.22\linewidth]{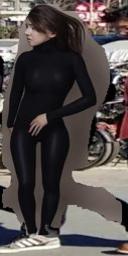} \\
\end{tabular}%
        }
        \vspace{1pt}
        \textbf{\scriptsize Identity 0458}
    \end{minipage}

    \caption{Qualitative comparison of anonymization on MSMT17 dataset (continued). Best viewed zoomed in.}
    \label{fig:qualitative_part_3_msmt}
\end{figure*}

\cref{fig:qualitative_part_1_msmt},~\cref{fig:qualitative_part_2_msmt},~\cref{fig:qualitative_part_3_msmt} shows additional qualitative results of our anonymization compared to the existing baselines on MSMT17. We can see that our method consistently outperforms the baselines in visual quality, consistency, and human structure. In each of the figures, we have shown results corresponding to three separate identities in the MSMT17 train set, in the corresponding subfigures (a), (b), (c). The columns show the four consecutive frames from the dataset, denoted F1, F2, F3, F4. `SD Base' denotes the `SD Baseline' method used in the main paper. `S-Anim' denotes the `Stable Animator' method used in the main paper. `Stream' denotes the `StreamV2V' method used in the main paper.

\subsection{MARS Dataset}
\begin{figure*}[t]
    \centering
    \setlength{\tabcolsep}{0.5pt}
    \renewcommand{\arraystretch}{0.5}

    \begin{minipage}[b]{\textwidth}
        \centering
        \resizebox{0.8\linewidth}{!}{%
        \begin{tabular}{@{}l@{\hspace{2pt}}c@{\hspace{1pt}}c@{\hspace{1pt}}c@{\hspace{1pt}}c@{\hspace{1pt}}c@{\hspace{1pt}}c@{\hspace{1pt}}c@{\hspace{1pt}}c@{}}
             & \scriptsize F1 & \scriptsize F2 & \scriptsize F3 & \scriptsize F4 & \scriptsize F5 & \scriptsize F6 & \scriptsize F7 & \scriptsize F8 \\
            \cmidrule(lr){2-9}
            \raisebox{1.5ex}{\scriptsize Original} &
            \includegraphics[width=0.11\linewidth]{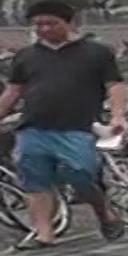} &
            \includegraphics[width=0.11\linewidth]{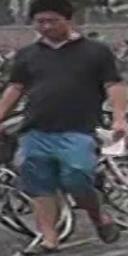} &
            \includegraphics[width=0.11\linewidth]{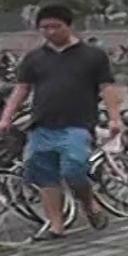} &
            \includegraphics[width=0.11\linewidth]{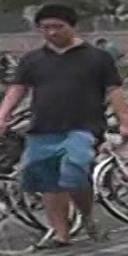} &
            \includegraphics[width=0.11\linewidth]{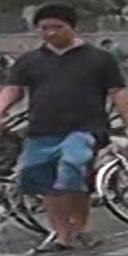} &
            \includegraphics[width=0.11\linewidth]{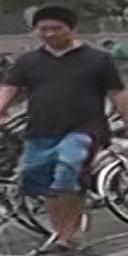} &
            \includegraphics[width=0.11\linewidth]{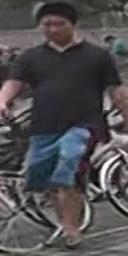} &
            \includegraphics[width=0.11\linewidth]{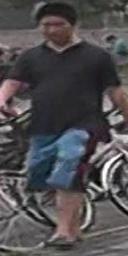} \\
\end{tabular}%
        }

        \vspace{1pt}
        \textbf{\scriptsize Identity 0005 (MARS)}
    \end{minipage}

    \vspace{4pt}

    \begin{minipage}[b]{\textwidth}
        \centering
        \resizebox{0.8\linewidth}{!}{%
        \begin{tabular}{@{}l@{\hspace{2pt}}c@{\hspace{1pt}}c@{\hspace{1pt}}c@{\hspace{1pt}}c@{\hspace{1pt}}c@{\hspace{1pt}}c@{\hspace{1pt}}c@{\hspace{1pt}}c@{}}
            \raisebox{1.5ex}{\scriptsize Original} &
            \includegraphics[width=0.11\linewidth]{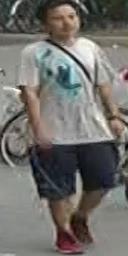} &
            \includegraphics[width=0.11\linewidth]{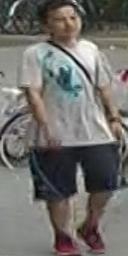} &
            \includegraphics[width=0.11\linewidth]{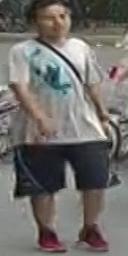} &
            \includegraphics[width=0.11\linewidth]{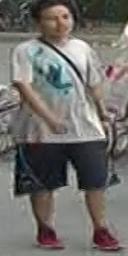} &
            \includegraphics[width=0.11\linewidth]{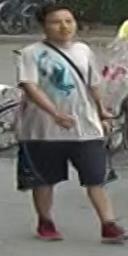} &
            \includegraphics[width=0.11\linewidth]{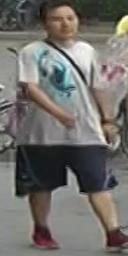} &
            \includegraphics[width=0.11\linewidth]{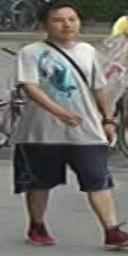} &
            \includegraphics[width=0.11\linewidth]{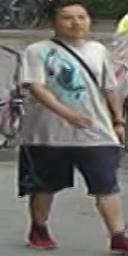} \\

            \raisebox{1.5ex}{\scriptsize DP2} &
            \includegraphics[width=0.11\linewidth]{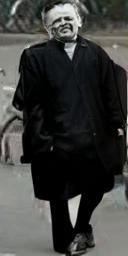} &
            \includegraphics[width=0.11\linewidth]{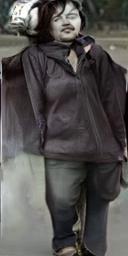} &
            \includegraphics[width=0.11\linewidth]{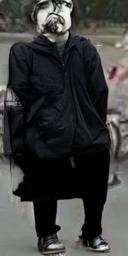} &
            \includegraphics[width=0.11\linewidth]{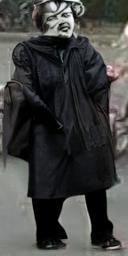} &
            \includegraphics[width=0.11\linewidth]{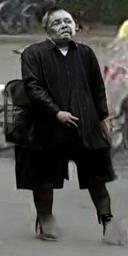} &
            \includegraphics[width=0.11\linewidth]{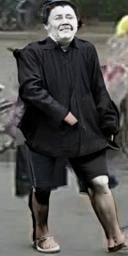} &
            \includegraphics[width=0.11\linewidth]{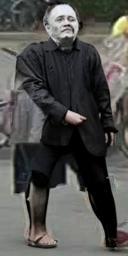} &
            \includegraphics[width=0.11\linewidth]{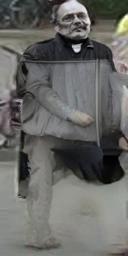} \\
\end{tabular}%
        }

        \vspace{1pt}
        \textbf{\scriptsize Identity 0007 (MARS)}
    \end{minipage}

    \caption{Qualitative comparison of anonymization on MARS dataset. Best viewed zoomed in.}
    \label{fig:mars_qualitative_part1}
\end{figure*}

\begin{figure*}[t]
    \centering
    \setlength{\tabcolsep}{0.5pt}
    \renewcommand{\arraystretch}{0.5}

    \begin{minipage}[b]{\textwidth}
        \centering
        \resizebox{0.8\linewidth}{!}{%
        \begin{tabular}{@{}l@{\hspace{2pt}}c@{\hspace{1pt}}c@{\hspace{1pt}}c@{\hspace{1pt}}c@{\hspace{1pt}}c@{\hspace{1pt}}c@{\hspace{1pt}}c@{\hspace{1pt}}c@{}}
            \raisebox{1.5ex}{\scriptsize Original} &
            \includegraphics[width=0.11\linewidth]{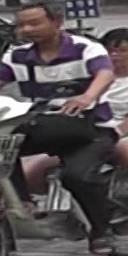} &
            \includegraphics[width=0.11\linewidth]{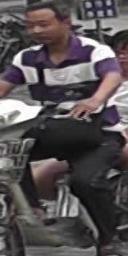} &
            \includegraphics[width=0.11\linewidth]{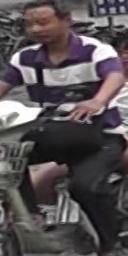} &
            \includegraphics[width=0.11\linewidth]{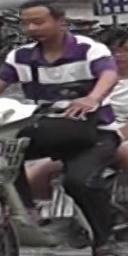} &
            \includegraphics[width=0.11\linewidth]{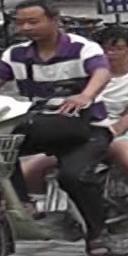} &
            \includegraphics[width=0.11\linewidth]{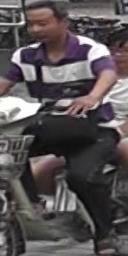} &
            \includegraphics[width=0.11\linewidth]{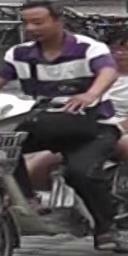} &
            \includegraphics[width=0.11\linewidth]{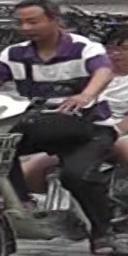} \\

            \raisebox{1.5ex}{\scriptsize DP2} &
            \includegraphics[width=0.11\linewidth]{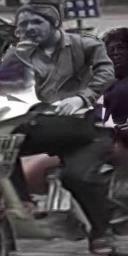} &
            \includegraphics[width=0.11\linewidth]{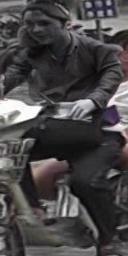} &
            \includegraphics[width=0.11\linewidth]{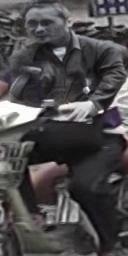} &
            \includegraphics[width=0.11\linewidth]{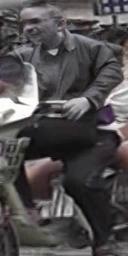} &
            \includegraphics[width=0.11\linewidth]{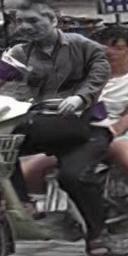} &
            \includegraphics[width=0.11\linewidth]{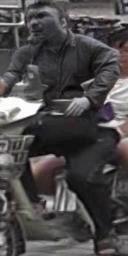} &
            \includegraphics[width=0.11\linewidth]{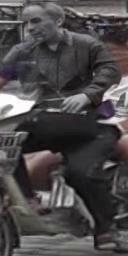} &
            \includegraphics[width=0.11\linewidth]{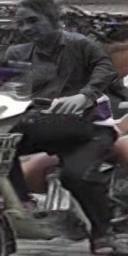} \\
\end{tabular}%
        }

        \vspace{1pt}
        \textbf{\scriptsize Identity 0013 (MARS)}
    \end{minipage}

    \vspace{4pt}

    \begin{minipage}[b]{\textwidth}
        \centering
        \resizebox{0.8\linewidth}{!}{%
        \begin{tabular}{@{}l@{\hspace{2pt}}c@{\hspace{1pt}}c@{\hspace{1pt}}c@{\hspace{1pt}}c@{\hspace{1pt}}c@{\hspace{1pt}}c@{\hspace{1pt}}c@{\hspace{1pt}}c@{}}
            \raisebox{1.5ex}{\scriptsize Original} &
            \includegraphics[width=0.11\linewidth]{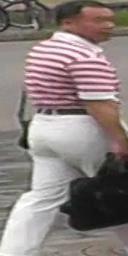} &
            \includegraphics[width=0.11\linewidth]{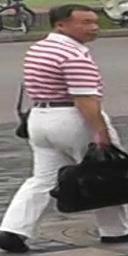} &
            \includegraphics[width=0.11\linewidth]{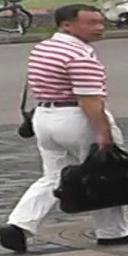} &
            \includegraphics[width=0.11\linewidth]{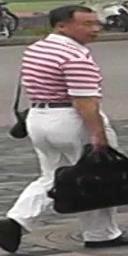} &
            \includegraphics[width=0.11\linewidth]{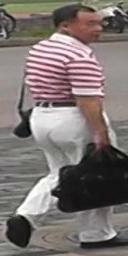} &
            \includegraphics[width=0.11\linewidth]{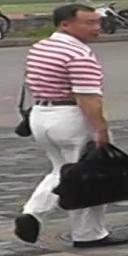} &
            \includegraphics[width=0.11\linewidth]{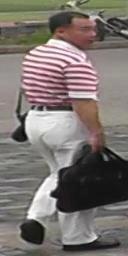} &
            \includegraphics[width=0.11\linewidth]{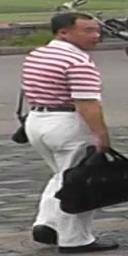} \\

            \raisebox{1.5ex}{\scriptsize DP2} &
            \includegraphics[width=0.11\linewidth]{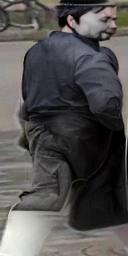} &
            \includegraphics[width=0.11\linewidth]{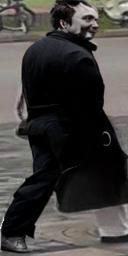} &
            \includegraphics[width=0.11\linewidth]{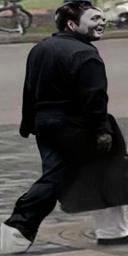} &
            \includegraphics[width=0.11\linewidth]{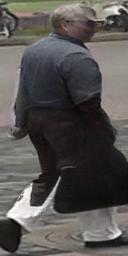} &
            \includegraphics[width=0.11\linewidth]{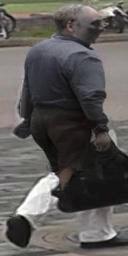} &
            \includegraphics[width=0.11\linewidth]{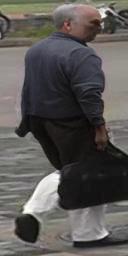} &
            \includegraphics[width=0.11\linewidth]{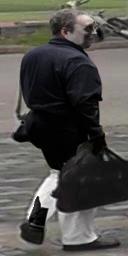} &
            \includegraphics[width=0.11\linewidth]{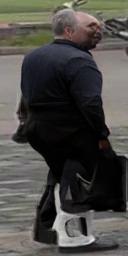} \\
\end{tabular}%
        }

        \vspace{1pt}
        \textbf{\scriptsize Identity 0023 (MARS)}
    \end{minipage}

    \caption{Qualitative comparison of anonymization on MARS dataset (continued). Best viewed zoomed in.}
    \label{fig:mars_qualitative_part2}
\end{figure*}

\begin{figure*}[t]
    \centering
    \setlength{\tabcolsep}{0.5pt}
    \renewcommand{\arraystretch}{0.5}

    \begin{minipage}[b]{\textwidth}
        \centering
        \resizebox{0.8\linewidth}{!}{%
        \begin{tabular}{@{}l@{\hspace{2pt}}c@{\hspace{1pt}}c@{\hspace{1pt}}c@{\hspace{1pt}}c@{\hspace{1pt}}c@{\hspace{1pt}}c@{\hspace{1pt}}c@{\hspace{1pt}}c@{}}
             & \scriptsize F1 & \scriptsize F2 & \scriptsize F3 & \scriptsize F4 & \scriptsize F5 & \scriptsize F6 & \scriptsize F7 & \scriptsize F8 \\
            \cmidrule(lr){2-9}
            \raisebox{1.5ex}{\scriptsize Original} &
            \includegraphics[width=0.11\linewidth]{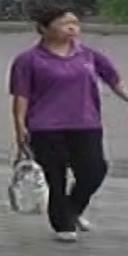} &
            \includegraphics[width=0.11\linewidth]{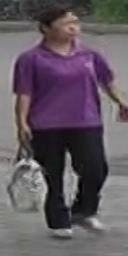} &
            \includegraphics[width=0.11\linewidth]{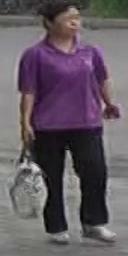} &
            \includegraphics[width=0.11\linewidth]{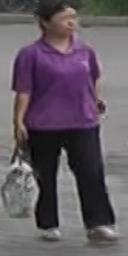} &
            \includegraphics[width=0.11\linewidth]{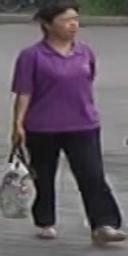} &
            \includegraphics[width=0.11\linewidth]{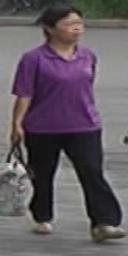} &
            \includegraphics[width=0.11\linewidth]{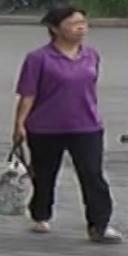} &
            \includegraphics[width=0.11\linewidth]{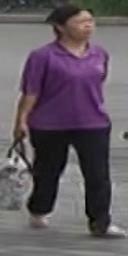} \\
\end{tabular}%
        }

        \vspace{1pt}
        \textbf{\scriptsize Identity 0093 (MARS)}
    \end{minipage}

    \vspace{4pt}

    \begin{minipage}[b]{\textwidth}
        \centering
        \resizebox{0.8\linewidth}{!}{%
        \begin{tabular}{@{}l@{\hspace{2pt}}c@{\hspace{1pt}}c@{\hspace{1pt}}c@{\hspace{1pt}}c@{\hspace{1pt}}c@{\hspace{1pt}}c@{\hspace{1pt}}c@{\hspace{1pt}}c@{}}
            \raisebox{1.5ex}{\scriptsize Original} &
            \includegraphics[width=0.11\linewidth]{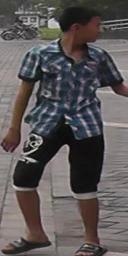} &
            \includegraphics[width=0.11\linewidth]{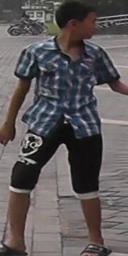} &
            \includegraphics[width=0.11\linewidth]{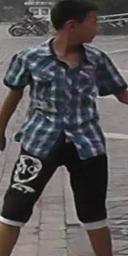} &
            \includegraphics[width=0.11\linewidth]{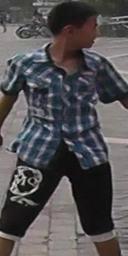} &
            \includegraphics[width=0.11\linewidth]{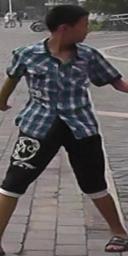} &
            \includegraphics[width=0.11\linewidth]{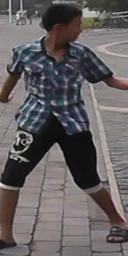} &
            \includegraphics[width=0.11\linewidth]{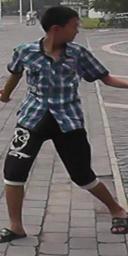} &
            \includegraphics[width=0.11\linewidth]{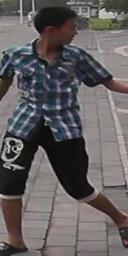} \\

            \raisebox{1.5ex}{\scriptsize DP2} &
            \includegraphics[width=0.11\linewidth]{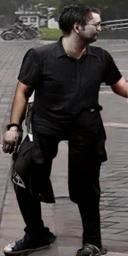} &
            \includegraphics[width=0.11\linewidth]{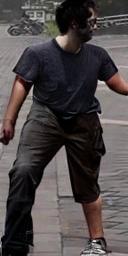} &
            \includegraphics[width=0.11\linewidth]{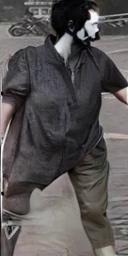} &
            \includegraphics[width=0.11\linewidth]{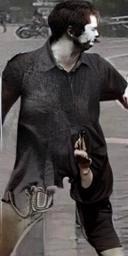} &
            \includegraphics[width=0.11\linewidth]{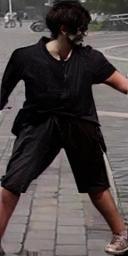} &
            \includegraphics[width=0.11\linewidth]{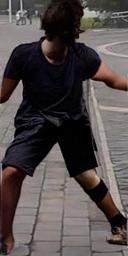} &
            \includegraphics[width=0.11\linewidth]{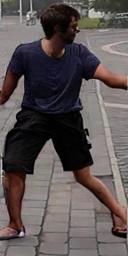} &
            \includegraphics[width=0.11\linewidth]{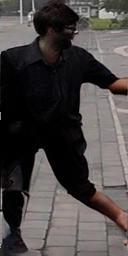} \\
\end{tabular}%
        }

        \vspace{1pt}
        \textbf{\scriptsize Identity 0121 (MARS)}
    \end{minipage}

    \caption{Qualitative comparison of anonymization on MARS dataset (continued). Best viewed zoomed in.}
    \label{fig:mars_qualitative_part3}
\end{figure*}

\cref{fig:mars_qualitative_part1},~\cref{fig:mars_qualitative_part2},~\cref{fig:mars_qualitative_part3} shows qualitative results of our anonymization on MARS dataset. We again notice that our method consistently outperforms the DP2 baseline, which is the existing SOTA full-body anonymization method. Our method generates higher quality human structures while being spatio-temporally consistent. We show results corresponding to two separate identities in each of the subfigures (a), (b). The columns show the eight consecutive video frames from the dataset, denoted F1, F2, F3, F4, F5, F6, F7, F8.

\subsection{UCF101 Dataset}
\begin{figure*}[t]
    \centering
    \setlength{\tabcolsep}{0.5pt}
    \renewcommand{\arraystretch}{0.5}

    \resizebox{\textwidth}{!}{%
    \begin{tabular}{@{}l@{\hspace{3pt}}c@{\hspace{1pt}}c@{\hspace{1pt}}c@{\hspace{1pt}}c@{\hspace{1pt}}c@{\hspace{1pt}}c@{\hspace{1pt}}c@{\hspace{1pt}}c@{}}
        & \scriptsize F1 & \scriptsize F2 & \scriptsize F3 & \scriptsize F4 & \scriptsize F5 & \scriptsize F6 & \scriptsize F7 & \scriptsize F8 \\
        \cmidrule(lr){2-9}

        \raisebox{1.5ex}{\tiny Original} &
        \includegraphics[width=0.115\linewidth]{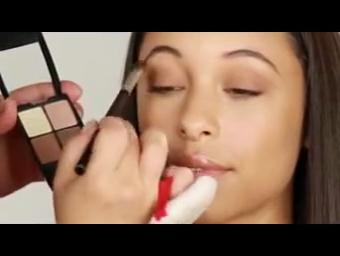} &
        \includegraphics[width=0.115\linewidth]{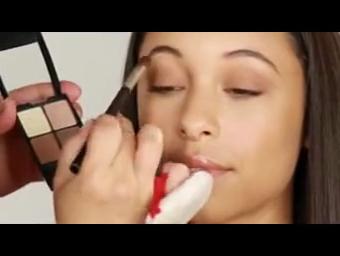} &
        \includegraphics[width=0.115\linewidth]{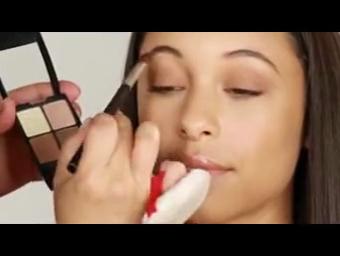} &
        \includegraphics[width=0.115\linewidth]{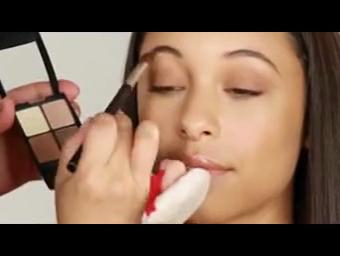} &
        \includegraphics[width=0.115\linewidth]{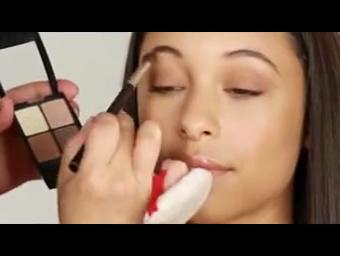} &
        \includegraphics[width=0.115\linewidth]{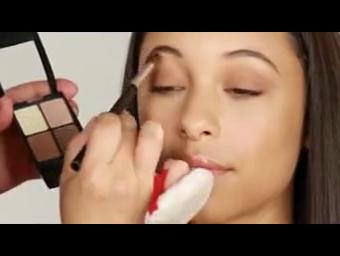} &
        \includegraphics[width=0.115\linewidth]{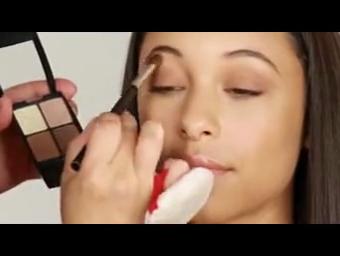} &
        \includegraphics[width=0.115\linewidth]{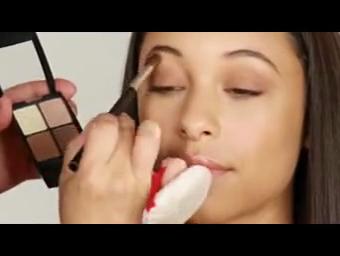} \\
\end{tabular}%
    }
    \caption{Qualitative comparison on UCF101 (ApplyEyeMakeup). We compare against DP2, SD Baseline, Stable Animator, and StreamV2V. Best viewed zoomed in.}
    \label{fig:ucf_makeup}
\end{figure*}

\begin{figure*}[t]
    \centering
    \setlength{\tabcolsep}{0.5pt}
    \renewcommand{\arraystretch}{0.5}

    \resizebox{\textwidth}{!}{%
    \begin{tabular}{@{}l@{\hspace{3pt}}c@{\hspace{1pt}}c@{\hspace{1pt}}c@{\hspace{1pt}}c@{\hspace{1pt}}c@{\hspace{1pt}}c@{\hspace{1pt}}c@{\hspace{1pt}}c@{}}
        & \scriptsize F1 & \scriptsize F2 & \scriptsize F3 & \scriptsize F4 & \scriptsize F5 & \scriptsize F6 & \scriptsize F7 & \scriptsize F8 \\
        \cmidrule(lr){2-9}

        \raisebox{1.5ex}{\tiny Original} &
        \includegraphics[width=0.115\linewidth]{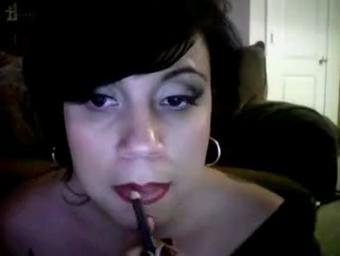} &
        \includegraphics[width=0.115\linewidth]{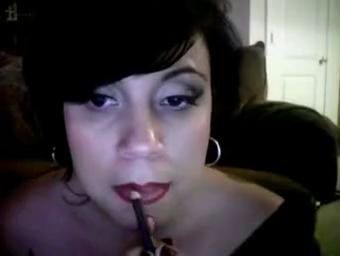} &
        \includegraphics[width=0.115\linewidth]{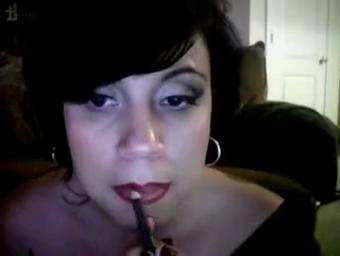} &
        \includegraphics[width=0.115\linewidth]{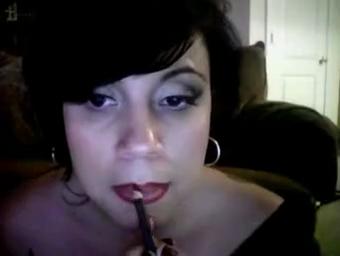} &
        \includegraphics[width=0.115\linewidth]{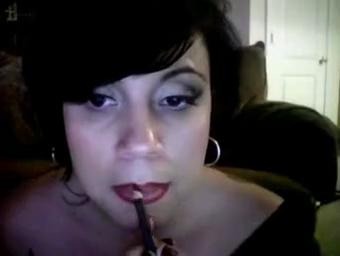} &
        \includegraphics[width=0.115\linewidth]{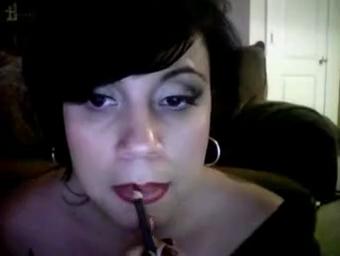} &
        \includegraphics[width=0.115\linewidth]{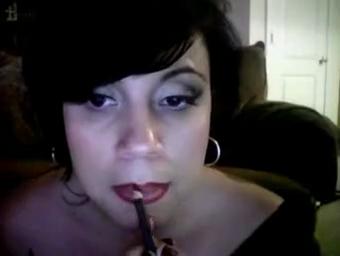} &
        \includegraphics[width=0.115\linewidth]{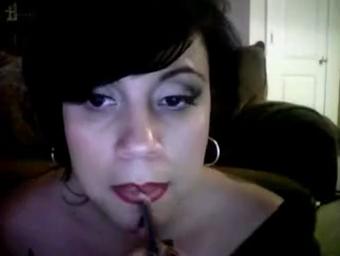} \\
\end{tabular}%
    }
    \caption{Qualitative comparison on UCF101 (ApplyLipstick). Best viewed zoomed in.}
    \label{fig:ucf_lipstick}
\end{figure*}

\begin{figure*}[t]
    \centering
    \setlength{\tabcolsep}{0.5pt}
    \renewcommand{\arraystretch}{0.5}

    \resizebox{\textwidth}{!}{%
    \begin{tabular}{@{}l@{\hspace{3pt}}c@{\hspace{1pt}}c@{\hspace{1pt}}c@{\hspace{1pt}}c@{\hspace{1pt}}c@{\hspace{1pt}}c@{\hspace{1pt}}c@{\hspace{1pt}}c@{}}
        & \scriptsize F1 & \scriptsize F2 & \scriptsize F3 & \scriptsize F4 & \scriptsize F5 & \scriptsize F6 & \scriptsize F7 & \scriptsize F8 \\
        \cmidrule(lr){2-9}

        \raisebox{1.5ex}{\tiny Original} &
        \includegraphics[width=0.115\linewidth]{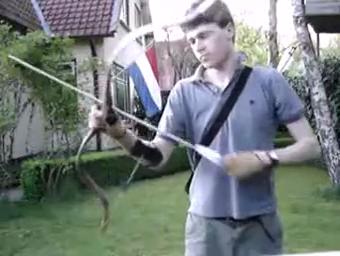} &
        \includegraphics[width=0.115\linewidth]{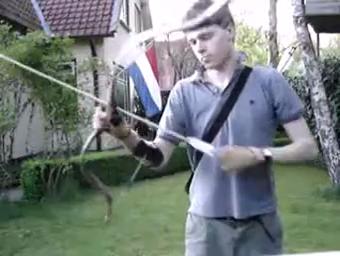} &
        \includegraphics[width=0.115\linewidth]{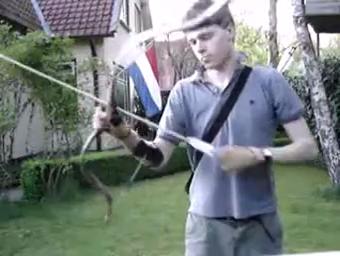} &
        \includegraphics[width=0.115\linewidth]{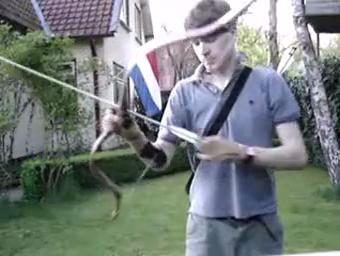} &
        \includegraphics[width=0.115\linewidth]{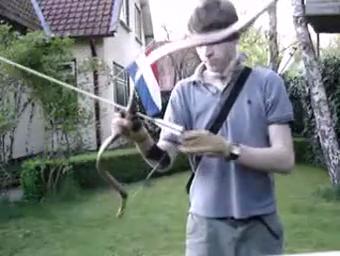} &
        \includegraphics[width=0.115\linewidth]{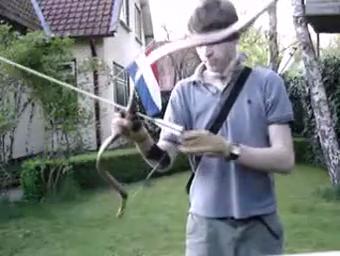} &
        \includegraphics[width=0.115\linewidth]{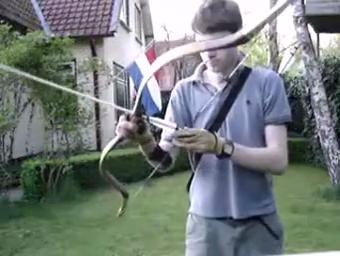} &
        \includegraphics[width=0.115\linewidth]{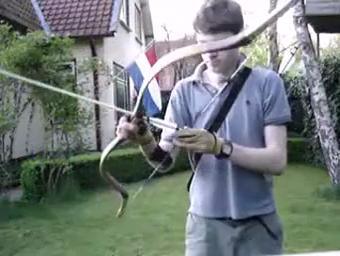} \\
\end{tabular}%
    }
    \caption{Qualitative comparison on UCF101 (Archery). Best viewed zoomed in.}
    \label{fig:ucf_archery}
\end{figure*}

\begin{figure*}[t]
    \centering
    \setlength{\tabcolsep}{0.5pt}
    \renewcommand{\arraystretch}{0.5}

    \resizebox{\textwidth}{!}{%
    \begin{tabular}{@{}l@{\hspace{3pt}}c@{\hspace{1pt}}c@{\hspace{1pt}}c@{\hspace{1pt}}c@{\hspace{1pt}}c@{\hspace{1pt}}c@{\hspace{1pt}}c@{\hspace{1pt}}c@{}}
        & \scriptsize F1 & \scriptsize F2 & \scriptsize F3 & \scriptsize F4 & \scriptsize F5 & \scriptsize F6 & \scriptsize F7 & \scriptsize F8 \\
        \cmidrule(lr){2-9}

        \raisebox{1.5ex}{\tiny Original} &
        \includegraphics[width=0.115\linewidth]{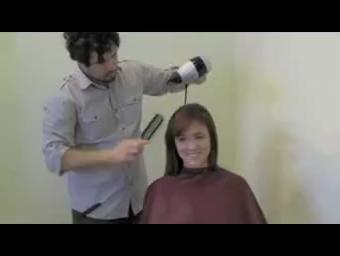} &
        \includegraphics[width=0.115\linewidth]{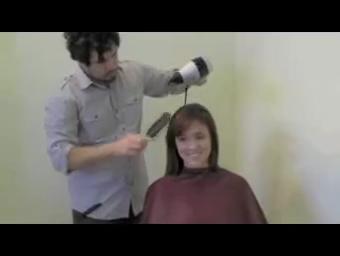} &
        \includegraphics[width=0.115\linewidth]{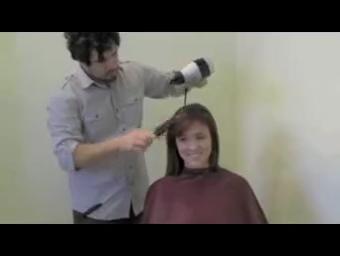} &
        \includegraphics[width=0.115\linewidth]{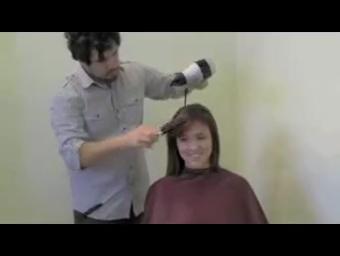} &
        \includegraphics[width=0.115\linewidth]{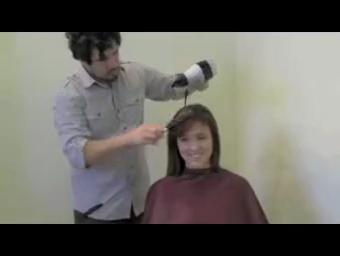} &
        \includegraphics[width=0.115\linewidth]{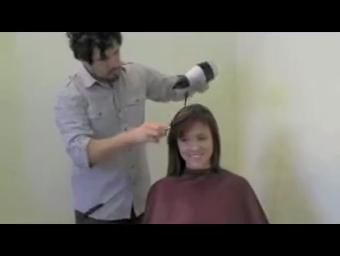} &
        \includegraphics[width=0.115\linewidth]{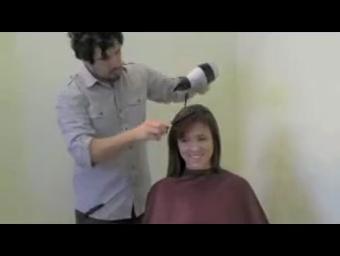} &
        \includegraphics[width=0.115\linewidth]{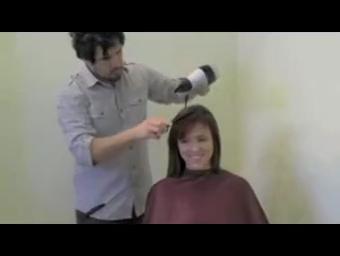} \\
\end{tabular}%
    }
    \caption{Qualitative comparison on UCF101 (BlowDryHair). Best viewed zoomed in.}
    \label{fig:ucf_hair}
\end{figure*}

\begin{figure*}[t]
    \centering
    \setlength{\tabcolsep}{0.5pt}
    \renewcommand{\arraystretch}{0.5}

    \resizebox{\textwidth}{!}{%
    \begin{tabular}{@{}l@{\hspace{3pt}}c@{\hspace{1pt}}c@{\hspace{1pt}}c@{\hspace{1pt}}c@{\hspace{1pt}}c@{\hspace{1pt}}c@{\hspace{1pt}}c@{\hspace{1pt}}c@{}}
        & \scriptsize F1 & \scriptsize F2 & \scriptsize F3 & \scriptsize F4 & \scriptsize F5 & \scriptsize F6 & \scriptsize F7 & \scriptsize F8 \\
        \cmidrule(lr){2-9}

        \raisebox{1.5ex}{\tiny Original} &
        \includegraphics[width=0.115\linewidth]{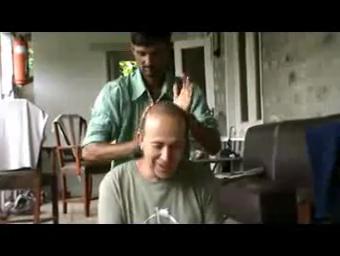} &
        \includegraphics[width=0.115\linewidth]{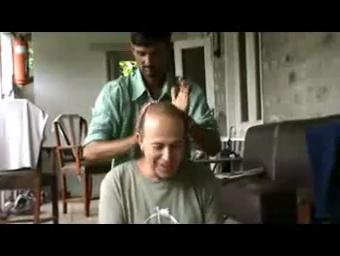} &
        \includegraphics[width=0.115\linewidth]{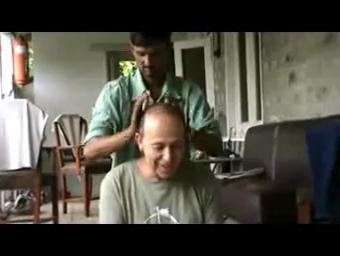} &
        \includegraphics[width=0.115\linewidth]{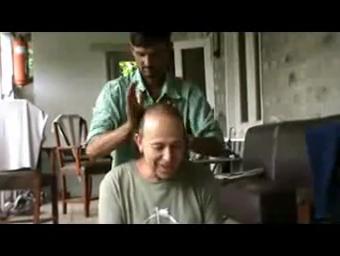} &
        \includegraphics[width=0.115\linewidth]{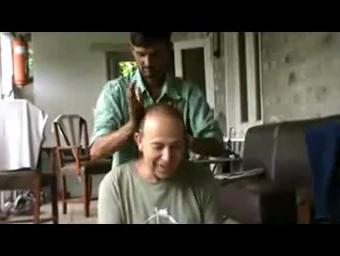} &
        \includegraphics[width=0.115\linewidth]{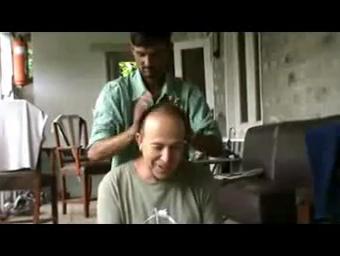} &
        \includegraphics[width=0.115\linewidth]{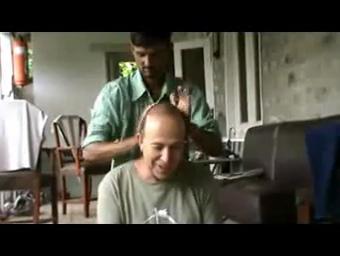} &
        \includegraphics[width=0.115\linewidth]{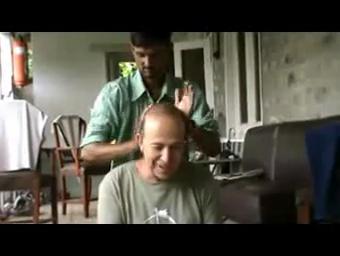} \\
\end{tabular}%
    }
    \caption{Qualitative comparison on UCF101 (HeadMassage). Best viewed zoomed in.}
    \label{fig:ucf_massage}
\end{figure*}

\cref{fig:ucf_makeup},~\cref{fig:ucf_lipstick},~\cref{fig:ucf_archery},~\cref{fig:ucf_hair},~\cref{fig:ucf_massage} show qualitative results of our anonymization on the UCF101 dataset. We observe much better spatio-temporal consistency and anonymization from our method compared to existing baselines. All figures show the anonymization results on eight consecutive frames from UCF101 videos, with categories including `ApplyEyeMakeup', `ApplyLipstick', `Archery', `BlowDryHair', and `HeadMassage'.

\subsection{YTVIS19 Dataset}
\begin{figure*}[p]
    \centering
    \setlength{\tabcolsep}{0.5pt}
    \renewcommand{\arraystretch}{0.5}

    \begin{minipage}[b]{\textwidth}
        \centering
        \resizebox{\textwidth}{!}{%
        \begin{tabular}{@{}l@{\hspace{3pt}}c@{\hspace{1pt}}c@{\hspace{1pt}}c@{\hspace{1pt}}c@{\hspace{1pt}}c@{\hspace{1pt}}c@{\hspace{1pt}}c@{\hspace{1pt}}c@{}}
            & \scriptsize F1 & \scriptsize F2 & \scriptsize F3 & \scriptsize F4 & \scriptsize F5 & \scriptsize F6 & \scriptsize F7 & \scriptsize F8 \\
            \cmidrule(lr){2-9}

            \raisebox{1.5ex}{\tiny Original} &
            \includegraphics[width=0.115\linewidth]{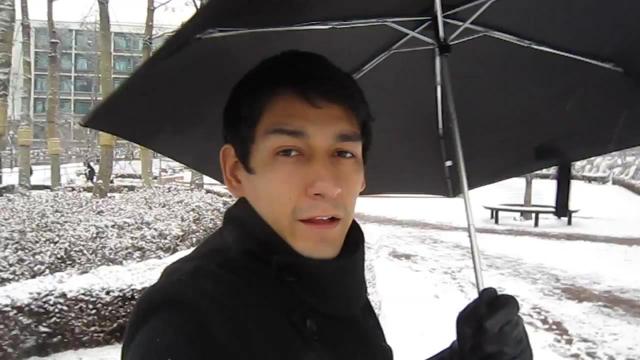} &
            \includegraphics[width=0.115\linewidth]{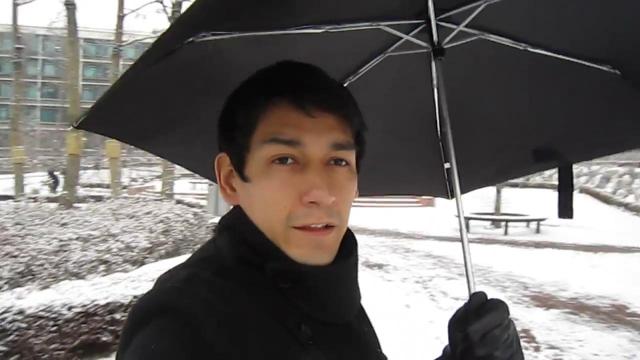} &
            \includegraphics[width=0.115\linewidth]{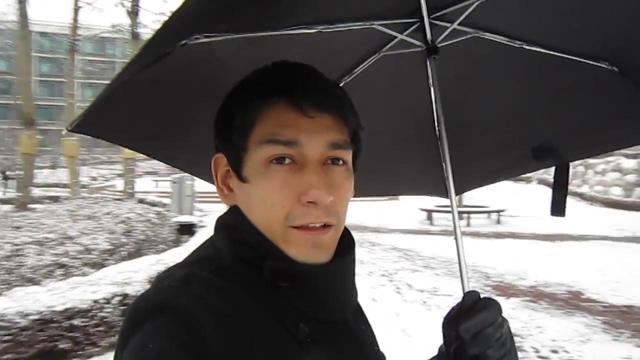} &
            \includegraphics[width=0.115\linewidth]{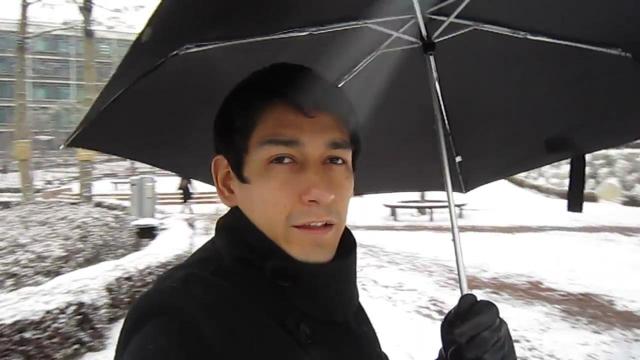} &
            \includegraphics[width=0.115\linewidth]{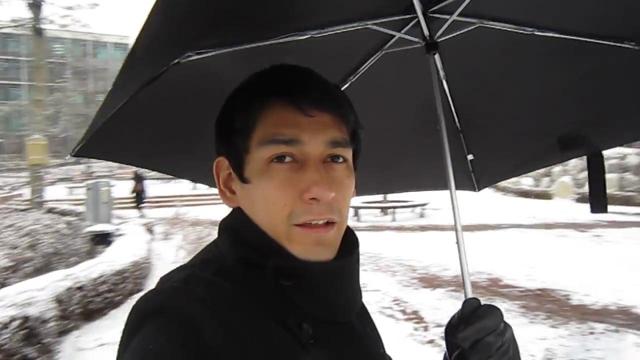} &
            \includegraphics[width=0.115\linewidth]{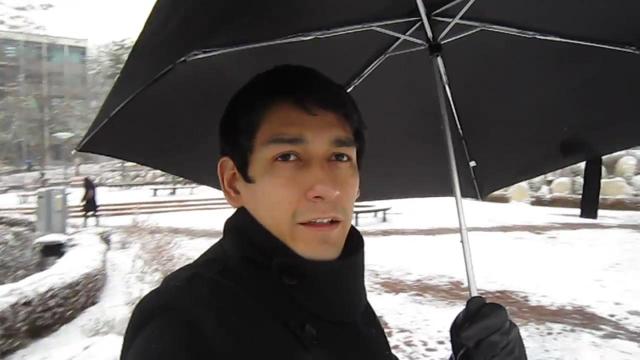} &
            \includegraphics[width=0.115\linewidth]{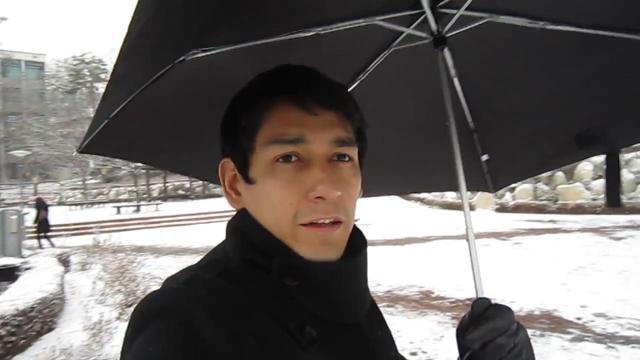} &
            \includegraphics[width=0.115\linewidth]{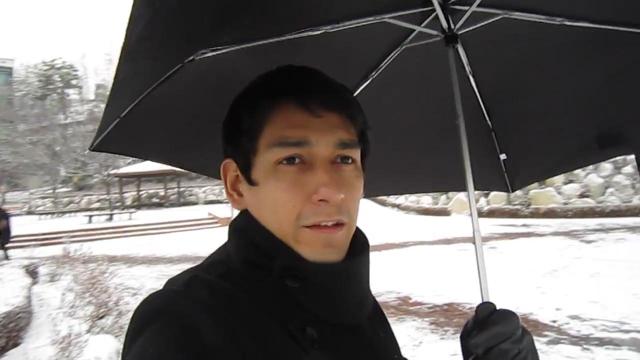} \\
\end{tabular}%
        }

        \vspace{1pt}
        \textbf{\scriptsize Video: 0e9f2785ec}
    \end{minipage}

    \vspace{4pt}

    \begin{minipage}[b]{\textwidth}
        \centering
        \resizebox{\textwidth}{!}{%
        \begin{tabular}{@{}l@{\hspace{3pt}}c@{\hspace{1pt}}c@{\hspace{1pt}}c@{\hspace{1pt}}c@{\hspace{1pt}}c@{\hspace{1pt}}c@{\hspace{1pt}}c@{\hspace{1pt}}c@{}}
            \raisebox{1.5ex}{\tiny Original} &
            \includegraphics[width=0.115\linewidth]{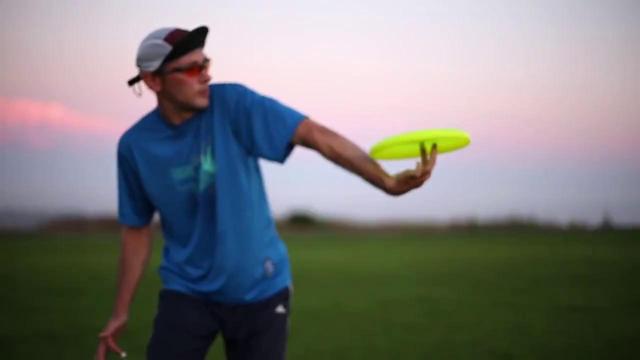} &
            \includegraphics[width=0.115\linewidth]{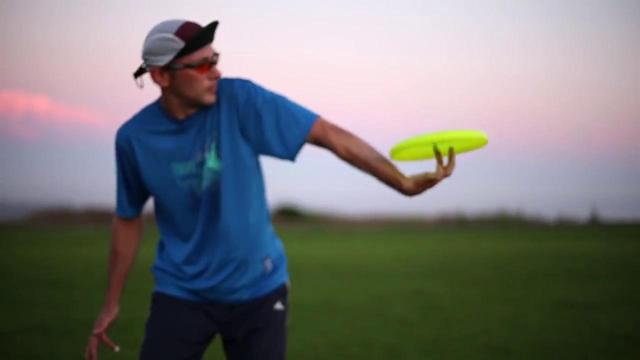} &
            \includegraphics[width=0.115\linewidth]{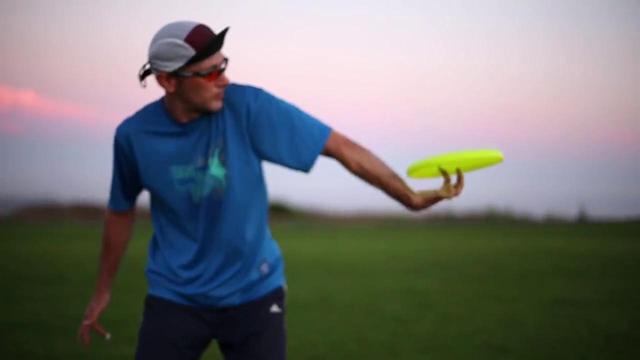} &
            \includegraphics[width=0.115\linewidth]{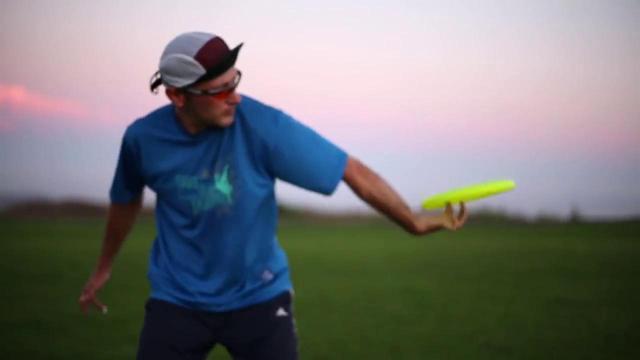} &
            \includegraphics[width=0.115\linewidth]{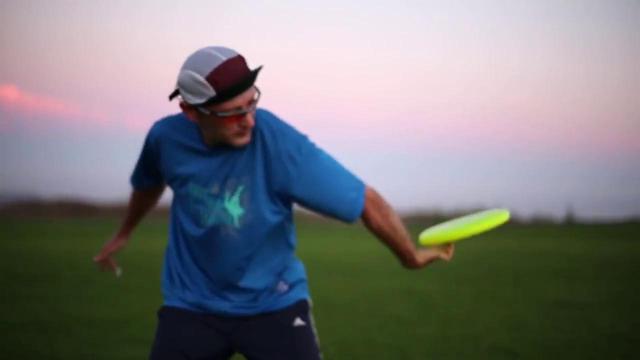} &
            \includegraphics[width=0.115\linewidth]{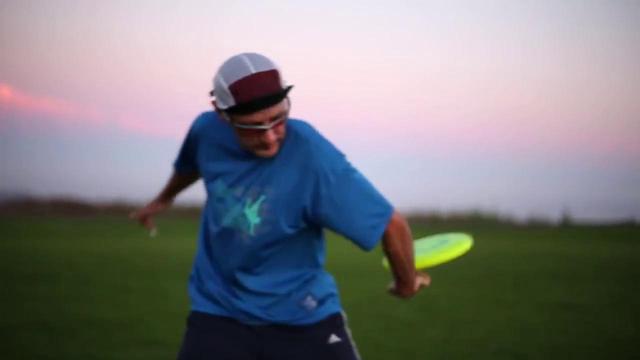} &
            \includegraphics[width=0.115\linewidth]{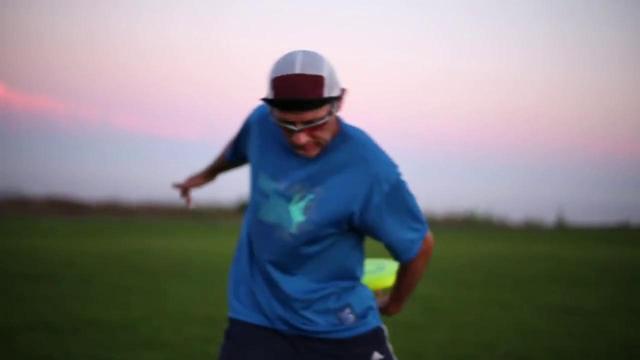} &
            \includegraphics[width=0.115\linewidth]{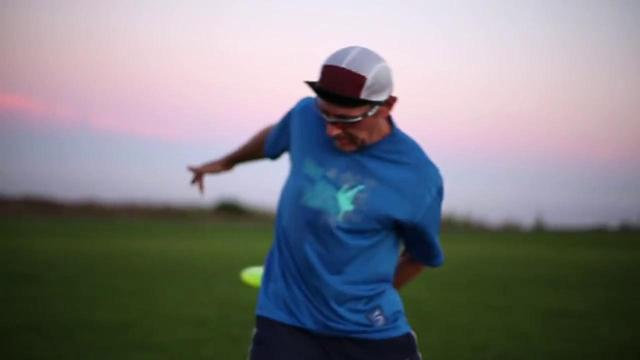} \\

            \raisebox{1.5ex}{\tiny DP2} &
            \includegraphics[width=0.115\linewidth]{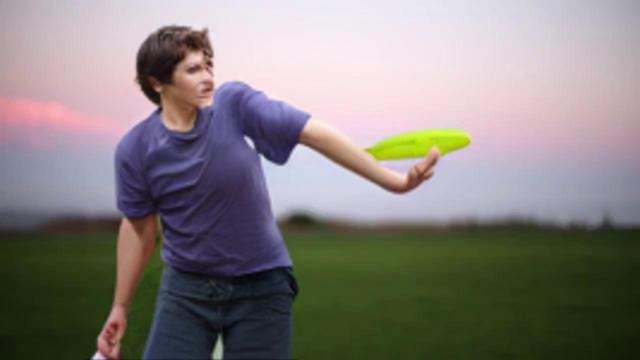} &
            \includegraphics[width=0.115\linewidth]{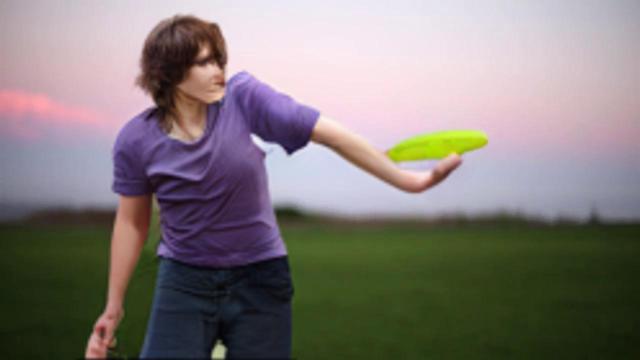} &
            \includegraphics[width=0.115\linewidth]{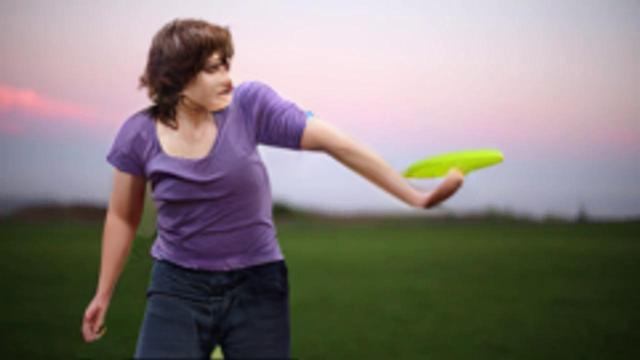} &
            \includegraphics[width=0.115\linewidth]{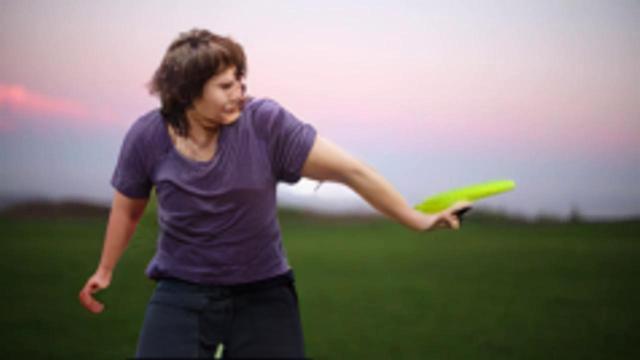} &
            \includegraphics[width=0.115\linewidth]{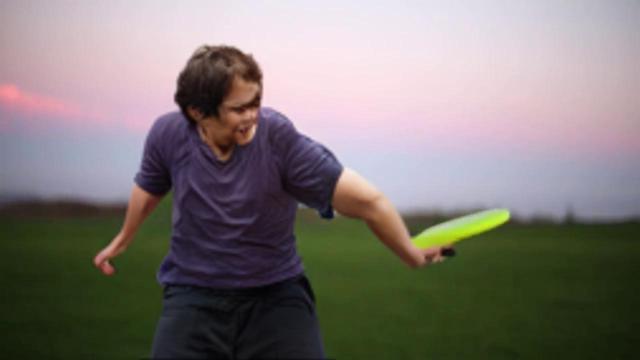} &
            \includegraphics[width=0.115\linewidth]{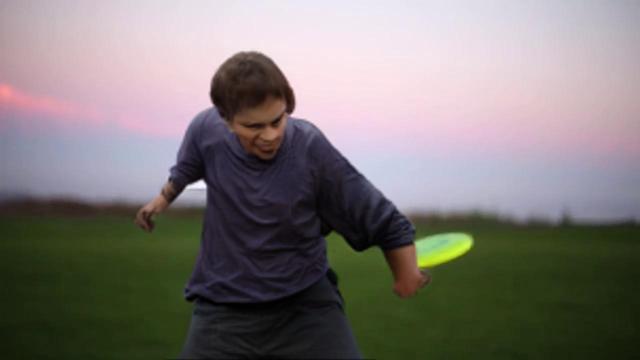} &
            \includegraphics[width=0.115\linewidth]{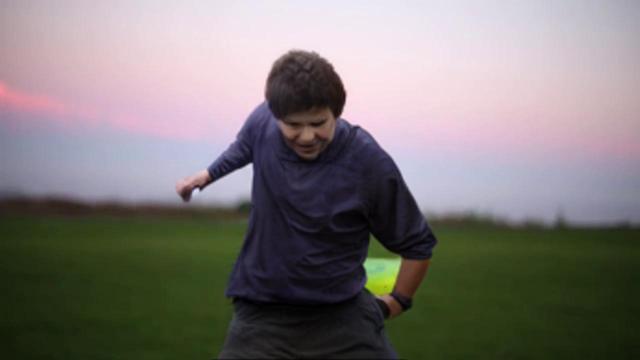} &
            \includegraphics[width=0.115\linewidth]{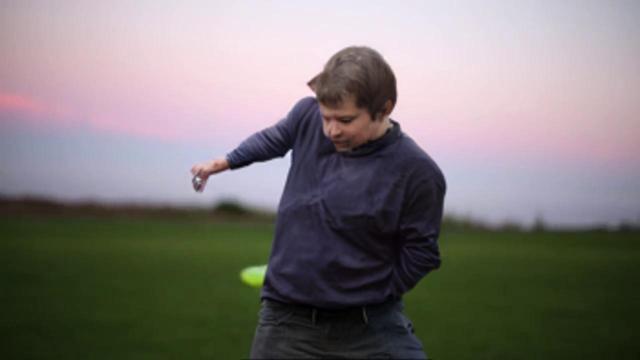} \\
\end{tabular}%
        }

        \vspace{1pt}
        \textbf{\scriptsize Video: 756f76f74d}
    \end{minipage}

    \caption{Qualitative comparison on YTVIS19 dataset. Our method demonstrates superior temporal consistency and feature preservation. Best viewed zoomed in.}
    \label{fig:vis19_qualitative_supplement}
\end{figure*}

\cref{fig:vis19_qualitative_supplement} shows qualitative results of our anonymization in the YTVIS19 data set. We have shown the results for two different videos in the figure and have plotted eight consecutive frames of the video from the dataset. While existing methods fail to preserve the human structure or maintain spatio-temporal consistency, our method shows superior quality of generation and better spatio-temporal consistency in the generated videos.

\end{document}